\documentclass[11pt]{article}

\usepackage[preprint]{acl}

\usepackage{times}
\usepackage{latexsym}

\usepackage[T1]{fontenc}

\usepackage[utf8]{inputenc}

\usepackage{microtype}

\usepackage{inconsolata}

\usepackage{graphicx}

\usepackage{pifont}
\newcommand{\cmark}{\ding{51}} 
\newcommand{\xmark}{\ding{55}} 

\usepackage{array}
\usepackage{amsmath}
\usepackage{enumitem}
\usepackage{subcaption}
\usepackage{multirow}
\usepackage{booktabs}
\usepackage{arydshln}
\usepackage[most]{tcolorbox}
\usepackage{marvosym}

\newtcolorbox{promptbox}[1]{
  breakable,
  colback=yellow!10!white,
  colframe=yellow!80!black,
  fonttitle=\bfseries,
  fontupper=\small,
  title=#1,
  boxrule=0.5mm,
  arc=3mm,
  top=1mm,
  bottom=1mm
}

\newcommand{\GitHubURL}{\url{https://github.com/vladman-25/RoD-TAL}}
\newcommand{\DatasetURL}{\url{https://huggingface.co/datasets/unstpb-nlp/RoD-TAL}}
\newcommand{\EmbeddingURL}{\url{https://huggingface.co/unstpb-nlp/multilingual-e5-small-RoD-TAL}}

\title{RoD-TAL: A Benchmark for Answering Questions in Romanian Driving License Exams}

\author{
  \textbf{Andrei Vlad Man\textsuperscript{1,}\thanks{Equal contribution.}},
  \textbf{Răzvan-Alexandru Smădu\textsuperscript{1,}\footnotemark[1]},
  \textbf{Cristian-George Craciun\textsuperscript{2}},
\\
  \textbf{Dumitru-Clementin Cercel\textsuperscript{1,\Letter}},
  \textbf{Florin Pop\textsuperscript{1,3}},
  \textbf{Mihaela-Claudia Cercel\textsuperscript{4,5}}
\\
\\
  \textsuperscript{1}National University of Science and Technology POLITEHNICA Bucharest, \\
Faculty of Automatic Control and Computers, Bucharest, Romania \\
  \textsuperscript{2}Technical University of Munich, Munich, Germany \\
  \textsuperscript{3}National Institute for Research \& Development in Informatics - ICI Bucharest, \\
Bucharest, Romania \\
  \textsuperscript{4}Paris 1 Panthéon-Sorbonne University, Paris, France \\
  \textsuperscript{5}University of Bucharest, Bucharest, Romania
\\
  \texttt{dumitru.cercel@upb.ro}
}

\begin{document}
\maketitle

\begin{abstract}
The intersection of AI and legal systems presents a growing need for tools that support legal education, particularly in under-resourced languages such as Romanian. In this work, we aim to evaluate the capabilities of Large Language Models (LLMs) and Vision-Language Models (VLMs) in understanding and reasoning about the Romanian driving law through textual and visual question-answering tasks. To facilitate this, we introduce RoD-TAL, a novel multimodal dataset comprising Romanian driving test questions, text-based and image-based, along with annotated legal references and explanations written by human experts. We implement and assess retrieval-augmented generation (RAG) pipelines, dense retrievers, and reasoning-optimized models across tasks, including Information Retrieval (IR), Question Answering (QA), Visual IR, and Visual QA. Our experiments demonstrate that domain-specific fine-tuning significantly enhances retrieval performance. At the same time, chain-of-thought prompting and specialized reasoning models improve QA accuracy, surpassing the minimum passing grades required for driving exams. We highlight the potential and limitations of applying LLMs and VLMs to legal education. 
We release the code and resources through the GitHub repository\footnote{\GitHubURL}.
\end{abstract}

\section{Introduction}

\begin{figure}[!th]
\centering
\begin{subfigure}{\columnwidth}
    \centering
    \includegraphics[width=\linewidth,trim={0.8cm 0.5cm 0.7cm 0.3cm},clip]{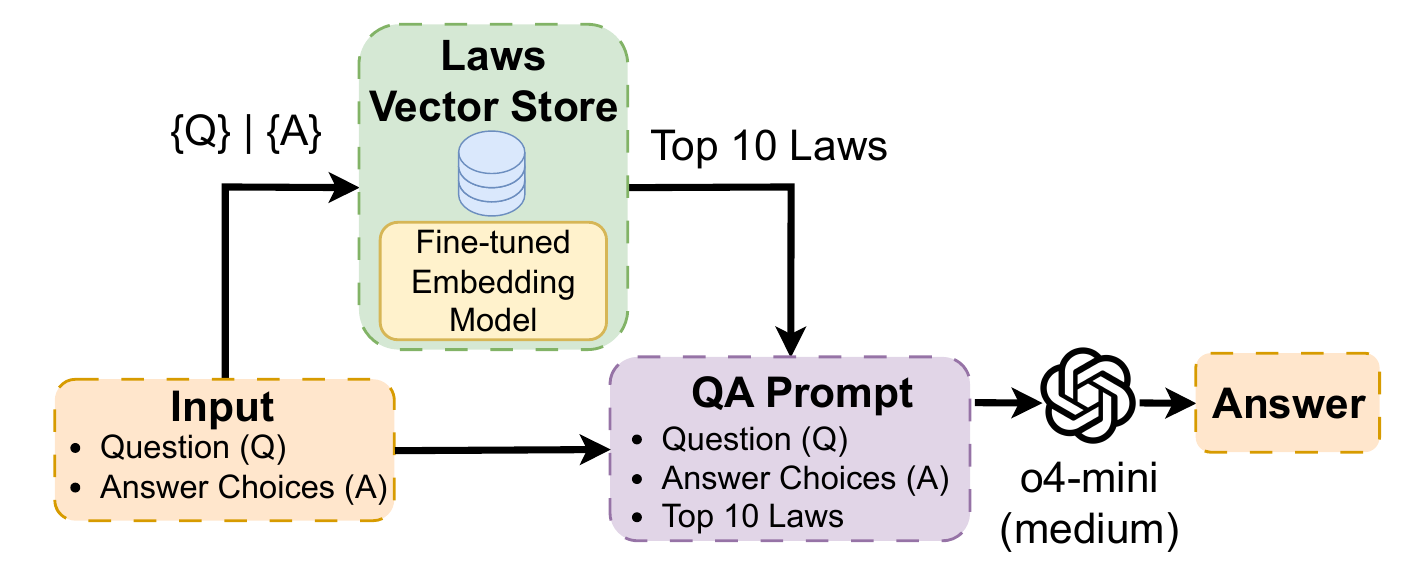}
    \caption{QA pipeline}
    \label{fig:architecture-qa}
\end{subfigure}
\hfill
\begin{subfigure}{\columnwidth}
    \centering
    \includegraphics[width=\linewidth,trim={0.8cm 0.5cm 0.7cm 0.3cm},clip]{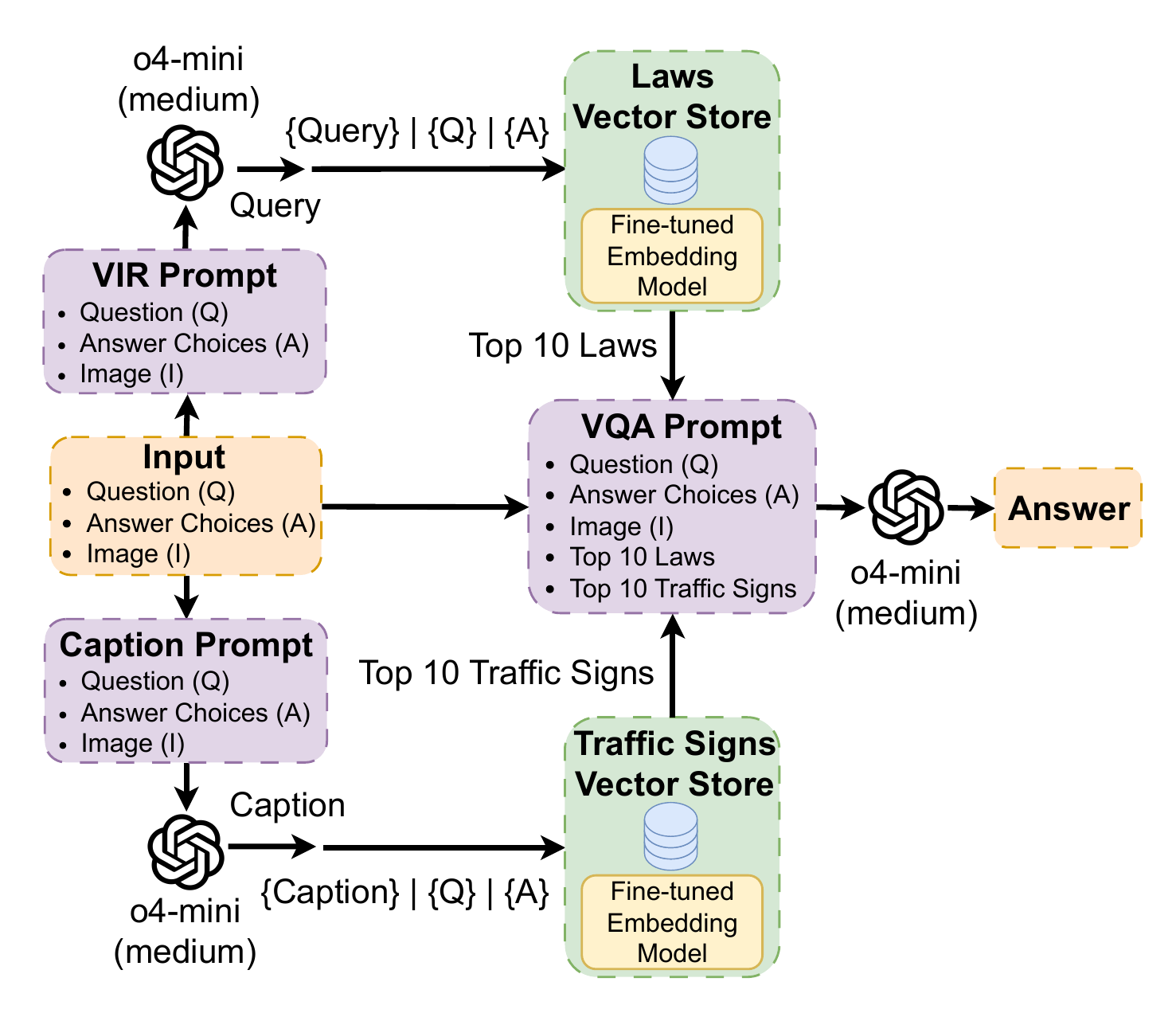}
    \caption{VQA pipeline}
    \label{fig:architecture-vqa}
\end{subfigure}
\caption{Architectures of the QA and VQA pipelines.}
\label{fig:architecture}
\end{figure}

\begin{table*}[ht!]
\centering
\small
\renewcommand{\arraystretch}{1.1}
\begin{tabular}{|l|c|c|c|c|c|c|c|}
\hline
\multicolumn{1}{|c|}{\multirow{2}{*}{\textbf{Method}}} & \multirow{2}{*}{\textbf{Backbone}} & \multirow{2}{*}{\textbf{Modality}} & \multicolumn{5}{c|}{\textbf{Accuracy}} \\
\cline{4-8}
 &  &  &  \textbf{US} & \textbf{JP} & \textbf{CN} & \textbf{SG} & \textbf{RO} \\
\hline
Zero-Shot QA \cite{zhou2024can} & GPT-4 & Text & 92.1\% & 86.5\% & 85.2\% & 88\% & - \\
Zero-Shot VQA \cite{zhou2024evaluating} & GPT-4V & Image & - & 66.7\% & 79.2\%  & - & - \\
\hdashline
IDKB \cite{lu2025can}& GPT-4o & Image &  \multicolumn{5}{c|}{Overall for 15 languages: 53\%} \\
\hdashline
RAG QA (Ours - Fig. \ref{fig:architecture-qa}) & o4-mini & Text & - & - & - & - & 86.4\% \\
RAG VQA (Ours - Fig. \ref{fig:architecture-vqa}) & o4-mini & Image & - & - & - & - & 78.3\% \\
\hline
\end{tabular}
\caption{Comparison of model accuracy across modalities and regions. Results are reported for text and images covering multiple languages and countries (US: United States; JP: Japan; CN: China; SG: Singapore; RO: Romania). Our method is evaluated on Romanian for both modalities.}

\label{tab:related-work-results}
\end{table*}

The intersection of artificial intelligence (AI), legal systems, and web technologies offers a powerful avenue to enhance public access to structured legal knowledge. In particular, road traffic law provides a rule-based, codified domain that is well-suited for computational reasoning and the development of intelligent, web-based educational tools. As web information systems evolve to integrate data-driven AI models, legal education, especially in underserved linguistic and regional contexts, remains a vital, yet underexplored, application area \cite{lai2024large}.
Despite significant advances in large language models (LLMs) and retrieval-augmented architectures, their application to legally grounded question answering in low-resource languages remains limited \cite{hijazi2024arablegaleval,das2024exams}. In countries where access to legal interpretation and educational resources is often inconsistent, such as Romania, there is a growing need for inclusive, intelligent systems that can support legal literacy and public understanding of the law \cite{guha2023legalbench,hoppe2021towards}.

Consequently, we focus this research on three objectives. First, evaluate LLMs' reasoning capabilities on question-answering (QA) and visual QA (VQA) tasks in legal and driving settings. Second, assess models' performance, biases, and limitations, and third, investigate how reliably these models can be integrated to support education and law-related tasks.
We curated and created a novel dataset, called RoD-TAL, composed of \textbf{Ro}manian \textbf{D}riving \textbf{T}ests \textbf{a}nd \textbf{L}aws, annotated and referenced in the QA/VQA pairs, to help us assess our objectives.

The main contributions of this work are:
\begin{itemize}[noitemsep,topsep=0pt]
    \item Introduce a novel QA/VQA dataset called RoD-TAL.
    \item Evaluate LLMs in the context of a low-resource language, Romanian, and a legal domain setting, exposing their biases and limitations.
    \item Propose a foundation for a legal Romanian-based dense retriever that does not focus solely on semantic similarity but on question and legal document alignments.
    \item Propose solutions (see Figure~\ref{fig:architecture} and Table~\ref{tab:related-work-results}) for all our identified research areas: information retrieval (IR), question answering (QA), visual IR (VIR), and visual QA (VQA), with strong results across all tasks.
    \item Present a thorough analysis of hallucinations in legal QA settings presented in Appendix \ref{sec:hallucinations-analysis}, ranging from citations, instruction following, and failure patterns.
\end{itemize}

\section{Related Work}

\begin{table*}[ht!]
\centering
\setlength{\tabcolsep}{2pt}
\small
\renewcommand{\arraystretch}{1.1}
\resizebox{\textwidth}{!}{
\begin{tabular}{|l|c|c|c|c|c|c|c|>{\arraybackslash}p{0.55cm}|>{\arraybackslash}p{0.55cm}|>{\arraybackslash}p{0.55cm}|>{\arraybackslash}p{0.55cm}|c|c|}
\hline
\multicolumn{1}{|c|}{\multirow{2}{*}{\textbf{Dataset}}} & \multicolumn{2}{c|}{\textbf{Data Type}} & \multicolumn{2}{c|}{\textbf{Data Source}} & \multicolumn{3}{c|}{\textbf{Data Domain}} & \multicolumn{4}{c|}{\textbf{Knowledge Domain}} & \multirow{2}{*}{\textbf{Size}} & \textbf{Law} \\
\cline{2-12}
 & \textbf{QA} & \textbf{MCQ} & \textbf{Real} & \textbf{Synth.} & \textbf{Country} & \textbf{Lang.} & \textbf{Type} & \textbf{LR} & \textbf{SS} & \textbf{DT} & \textbf{DD} & & \textbf{Ref.} \\
\hline

BDD-X \cite{kim2018textual} & \xmark & \xmark & \cmark & \xmark & US & EN & Car & \xmark & \cmark & \cmark & \xmark & 26K & \xmark \\
Talk2Car \cite{deruyttere2019talk2car} & \xmark & \xmark & \cmark & \xmark & US, SG & EN & Car & \xmark & \xmark & \xmark & \xmark & 12K & \xmark \\
CODA-LM \cite{li2022coda} & \cmark & \xmark & \cmark & \xmark & DE, CN, SG & EN & Car & \xmark & \cmark & \cmark & \xmark & 10K & \xmark \\
DRAMA \cite{malla2023drama} & \xmark & \xmark & \cmark & \xmark & JP & EN & Car & \xmark & \cmark & \cmark & \xmark & 102K & \xmark \\
nuScenes-QA \cite{qian2024nuscenes} & \cmark & \xmark & \cmark & \xmark & US, SG & EN & Car & \xmark & \xmark & \xmark & \xmark & 460K & \xmark \\
DriveLM \cite{sima2024drivelm} & \cmark & \xmark & \cmark & \cmark & US, SG & EN & Car & \xmark & \cmark & \cmark & \xmark & 2M & \xmark \\
LangAuto CARLA \cite{shao2024lmdrive} & \xmark & \xmark & \xmark & \cmark & US & EN & Car & \xmark & \cmark & \cmark & \xmark & 64K & \xmark \\
SUP-AD \cite{tian2024drivevlm} & \xmark & \xmark & \cmark & \xmark & CN & EN & Car & \xmark & \cmark & \cmark & \xmark & - & \xmark \\
VLAAD \cite{park2024vlaad} & \cmark & \xmark & \cmark & \xmark & US & EN & Car & \xmark & \cmark & \cmark & \xmark & 64K & \xmark \\
IDKB \cite{lu2025can} & \cmark & \cmark & \cmark & \cmark & 15 & 9 & 4 & \cmark & \cmark & \cmark & \cmark & 1M & \xmark \\

\hline
RoD-TAL (Ours) & \xmark & \cmark & \cmark & \xmark & RO & RO & Car & \cmark & \cmark & \cmark & \cmark & \begin{tabular}{@{}c@{}} 1.2K$^{*}$ \end{tabular} & \cmark\\
\hline
\end{tabular}}
\caption{Comparison of datasets, domains, and knowledge coverage, inspired by \citet{lu2025can}. The data source is categorized as real or synthetic (synth.). Knowledge Domain spans: \textit{laws and regulations} (LR), \textit{signs and signals} (SS), \textit{driving techniques} (DT), and \textit{defensive driving} (DD). Countries: China (CN), Japan (JP), Romania (RO), Singapore (SG), and the United States (US). $^{*}$RoD-TAL spans 1.2K questions, of which 400 contain images.}
\label{tab:related-work-datasets}
\end{table*}

\textbf{Driving QA and VQA.} \citet{zhou2024can} evaluated ChatGPT with GPT-4 \cite{openai2024gpt4technicalreport} on 814 written driving-license questions from California, Tokyo, Beijing, and Singapore, assessing performance across dimensions such as legal reasoning, situational understanding, and safety bias. The accuracy ranged from 85.2\% to 92.1\%, with lower scores on region-specific legal questions (e.g., 63.2\% in China). Although the model performed well overall, it showed limitations in handling local regulations and context-sensitive reasoning.

In follow-up work, \citet{zhou2024evaluating} evaluated vision-language models (VLMs) such as ChatGPT and Bard on visual driving-license questions from Tokyo and Beijing. Although the models performed moderately on traffic sign recognition (70\% accuracy) and better on scenario-based (80\% accuracy) and combined visual tasks (80\% accuracy), the study highlighted ongoing challenges in applying VLMs to real-world autonomous driving contexts.

However, these studies evaluated out-of-the-box LLM or VLM performance; they did not aim to optimize model accuracy or incorporate retrieval augmentation techniques. Although GPT models have likely encountered many laws and question pairs during pretraining, reframing this task as a retrieval-augmented generation (RAG) \cite{DBLP:conf/nips/LewisPPPKGKLYR020} problem could offer a more principled and scalable approach to improving legal and regulation-specific reasoning.

\textbf{Vision-Language Driving Datasets.} Multiple works \cite{kim2018textual,deruyttere2019talk2car,li2022coda,malla2023drama,qian2024nuscenes,sima2024drivelm,shao2024lmdrive,tian2024drivevlm,park2024vlaad,lu2025can} have introduced and evaluated vision-language driving datasets for autonomous driving, some based on open QA, or multiple-choice question answering (MCQA), and covering multiple languages, topics, and vehicle categories. Table~\ref{tab:related-work-datasets} compares some of these resources. Additionally, their limitation is a lack of references to legal corpora and reliance solely on the internal knowledge of vision-language models during QA tasks.

\textbf{Multiple-Choice QA with LLMs Across Domains.} \citet{zhong2304agieval} assessed LLMs, including GPT-4, on Chinese-based human exams such as the Law School Admission Test (LSAT) and Lawyer Qualification Test (LQT), using formats based on multiple-choice and fill-in-the-blank. Among various prompting strategies, few-shot with chain-of-thought (CoT) \cite{DBLP:conf/nips/Wei0SBIXCLZ22} performed the best. GPT-4 scored 34–40\% on the LQT and 31–87\% on the LSAT. The study found stronger reasoning in high-resource languages such as English and in domains such as history and logic, while underperforming in law, math, and physics. Challenges included concept disambiguation, strict logical reasoning, and multi-hop inference, highlighting areas for further improvement. In Romanian, a proposed MCQA dataset for the legal domain \cite{cruaciun2024graf} was curated from legal examinations of different levels and specializations, and an MCQA dataset for the medical domain \citep{dima2024roqllama} was built from university entrance tests.

\textbf{Multiple-Choice VQA.} \citet{das2024exams} proposed EXAMS-V, a multimodal, multilingual benchmark for evaluating VLMs on multiple-choice questions across diverse domains excluding law. GPT-4V \cite{openai2023gpt4v} scored an average of 42.5\%, revealing limitations in multimodal reasoning and in integrating visual and textual information, despite early potential.

\textbf{Information Retrieval and Retriever Fine-Tuning.} \citet{moreira2024nv} investigated fine-tuning dense retrievers for RAG tasks, emphasizing hard-negative mining strategies. Their results showed that starting from positive-aware setups and gradually introducing harder negatives significantly improved retrieval accuracy and response quality, underscoring the importance of training data curation in RAG pipelines.

\textbf{Visual Information Retrieval.} \citet{dong2024modality} proposed a modality-aware retrieval approach that leverages visual LLMs to generate dense image captions for downstream querying, thus integrating visual content more effectively and enhancing retrieval performance.

\section{RoD-TAL: Romanian Driving Tests And Laws}

The novel resource RoD-TAL comprises a law corpus from Romanian legislation, RoD-Law, and an MCQA dataset, RoD-QA, including text- and image-based questions.

\subsection{RoD-Law Corpus}

A central component of the RoD-TAL framework is its law-grounded foundation, the RoD-Law corpus, which ensures that every answer can be explicitly traced back to Romanian legislation. The legal corpus, used as the retrieval base for all downstream tasks in this work, was compiled from official sources, with abrogated sections removed to ensure relevance and currency, until March 2025.
The composition of the legal corpus is summarized in Table~\ref{tab:corpus},
including laws from traffic regulations, road code, penal code, and technical
inspection law, as well as civil auto liability insurance law.

\begin{table}[t!]
\centering
\small
\begin{tabular}{|c|c|}
\hline
\textbf{Corpus Source} & \textbf{Articles} \\
\hline
Traffic Regulation Rules & 225 \\
Road Code & 147 \\
Penal Code (traffic-related) & 9 \\
Technical Inspection Law & 15 \\
Civil Auto Liability Insurance Law & 47 \\
\hdashline
Traffic Signs & 140 \\
\hline
\end{tabular}
\caption{Distribution of articles in the RoD-Law corpus by legal source.}
\label{tab:corpus}
\end{table}
\begin{figure}[!t]
\centering
\includegraphics[width=\columnwidth,trim=1.15cm 0cm 0cm 1.2cm,clip]{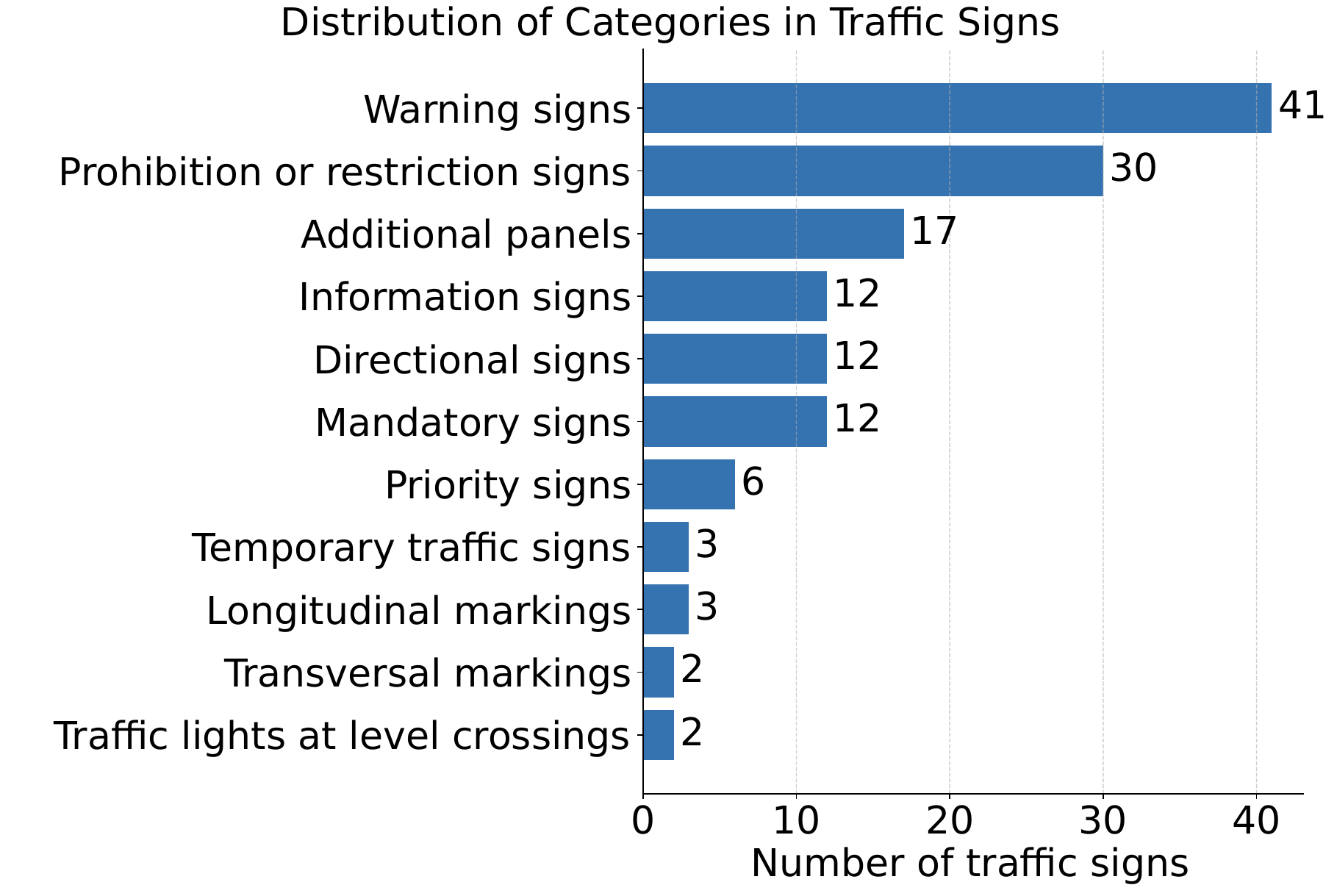}
\caption{Distribution of traffic signs per category.}
\label{fig:distribution_corpus_indicators_category_horizontal}
\end{figure}

The RoD-Law legal corpus is complemented by an annotated collection of 140 distinct traffic signs, extracted from the answer references throughout the dataset and spanning 11 major categories (see Figure~\ref{fig:distribution_corpus_indicators_category_horizontal}). Each sign is provided with its name, category, and a concise explanation, which supports both QA annotation and the VQA tasks. Other statistics are presented in Appendix \ref{app:rod_law_stats}.

\subsection{RoD-QA Dataset}
\label{sec:rod_qa_dataset}

\paragraph{Dataset Structure.}
Based on a curated legal base, the RoD-TAL dataset consists of multiple-choice questions sourced from the written Romanian driving-license tests, henceforth referred to as RoD-QA, available on the public educational platform Scoala Rutiera\footnote{\url{https://www.scoalarutiera.ro/}}. Each question is annotated with relevant legal references from RoD-Law, enabling the evaluation of both standard LLMs and RAG systems grounded in actual law. \textit{Scoala Rutiera} is one of the largest platforms in Romania for learning and preparing for theoretical driving tests, according to internet traffic\footnote{\url{https://www.similarweb.com/website/scoalarutiera.ro/\#overview}} and mobile app downloads on Google Play\footnote{\url{ https://play.google.com/store/apps/details?id=com.app.scoalarutiera}}, as of December 2025. As claimed on their website, the questions are the official ones provided by the authorities during the exams. Therefore, we assume that these annotations are correct.


The data structure for each sample includes the question, a set of candidate answers (with an explicit correct answer or more), an explanation, legal references, and a list of traffic signs where relevant. The visual questions were further categorized and enriched by o4-mini-based \cite{o3-o4mini-system-card} sign annotation, followed by manual verification. For the experimental setup, we evaluated various combinations of text- and image-based questions with or without law annotations. An overview of the dataset splits is provided in Table~\ref{tab:datasets}.
This comprehensive annotation schema supports evaluation of not only the LLM answer accuracy, but also the legal retrieval precision/recall and legal grounding in RAG setups. The presence of visual (VQA) questions further enables research into legally grounded multimodal models, an emerging area in AI and law.
We make the dataset publicly available on HuggingFace\footnote{\DatasetURL}.

\begin{table}[t!]
\centering
\small
  \centering
  \begin{tabular}{|c|c|c|c|}
    \hline
    \textbf{Dataset} & \textbf{Modality} & \textbf{Law Ref.} & \textbf{Size} \\
    \hline
    Split 1 & Text-based & \cmark & 638 \\
    Split 2 & Text-based & \xmark & 131 \\
    Split 3 & Image-based & \cmark & 316 \\
    Split 4 & Image-based & \xmark & 71 \\
    \hline
  \end{tabular}
  \caption{RoD-QA splits by modality and legal reference annotation.}
  \label{tab:datasets}
\end{table}
\begin{figure}[!t]
\centering
\includegraphics[width=\columnwidth,trim=0.2cm 0 0.2cm 1.2cm,clip]{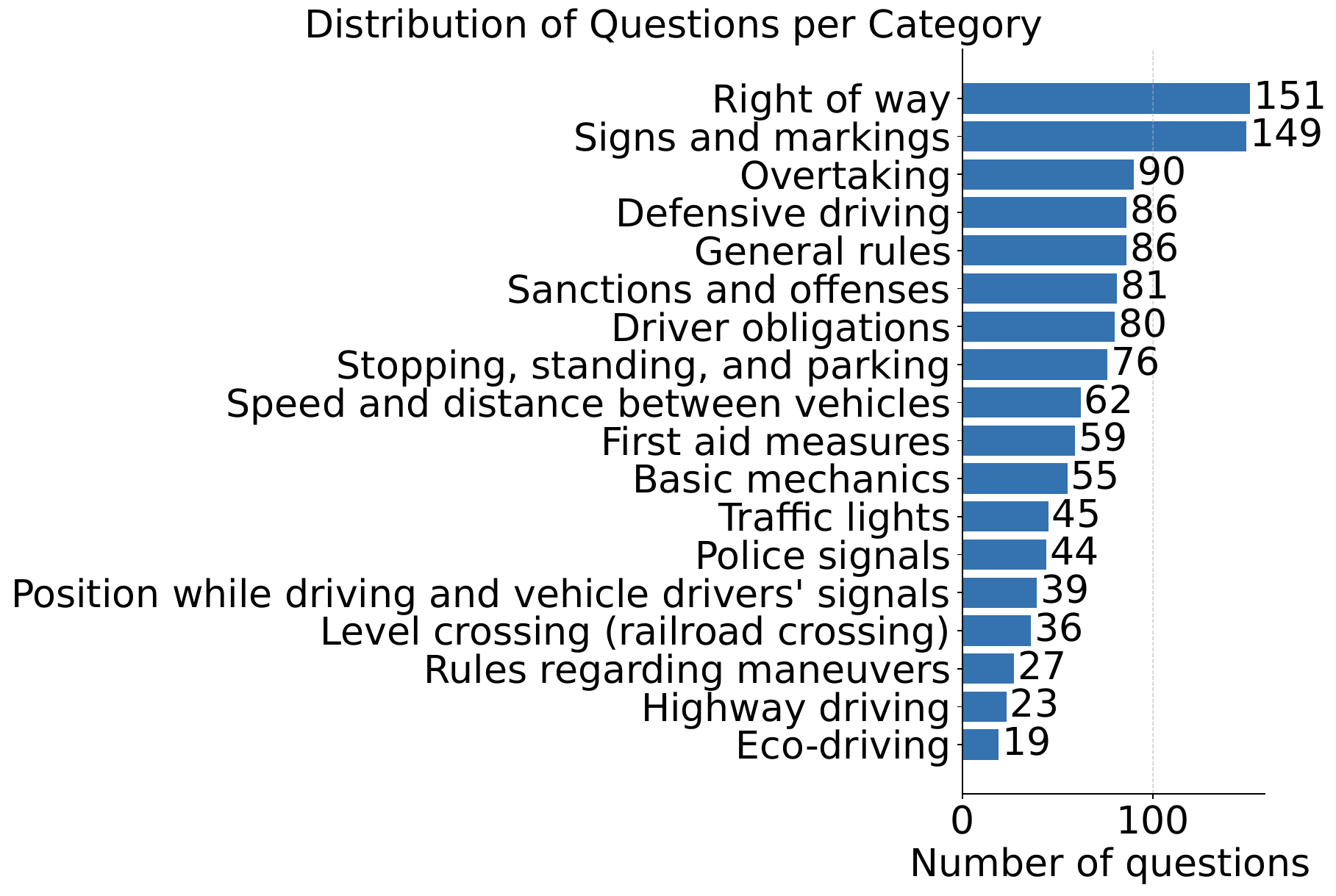}
\caption{Distribution of questions per primary category.}
\label{fig:distribution_questions_per_category}
\end{figure}

\begin{figure}[!t]
\centering
\includegraphics[width=\columnwidth,trim=0 0 0 1.6cm,clip]{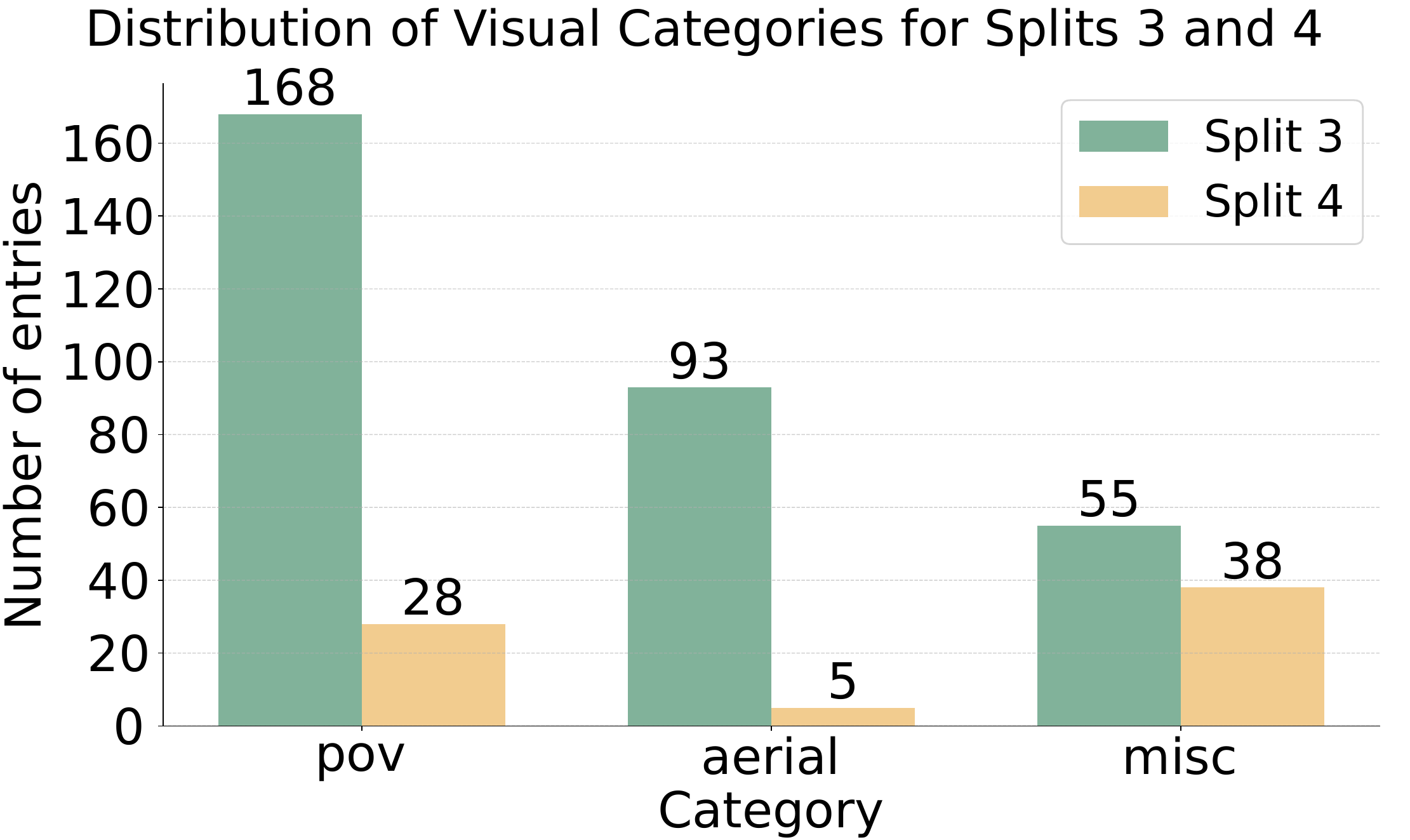}
\caption{Distribution of visual questions per secondary category for splits 3 and 4.}
\label{fig:distribution_category_split3_split4}
\end{figure}

\begin{figure*}[htbp]
\centering
\includegraphics[trim=0.1cm 1.2cm 0.1cm 1cm,clip,width=\textwidth]{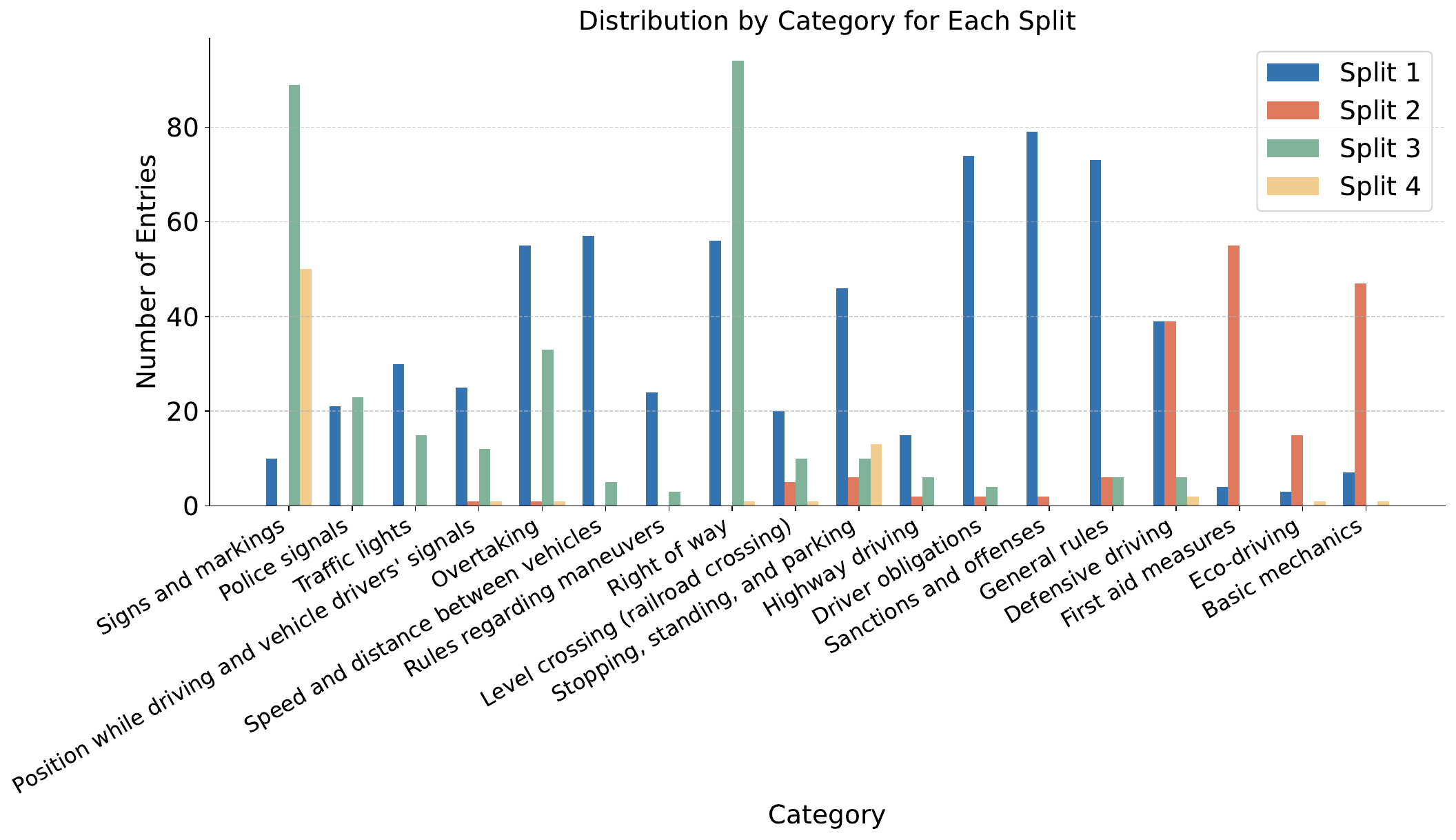}
\caption{Number of questions per primary category and split.}
\label{fig:distribution_by_category_per_split}
\end{figure*}

\paragraph{Dataset Statistics.}
The dataset encompasses 18 primary question categories, with the visual subset further categorized into three secondary categories: point of view (pov), aerial, and miscellaneous (misc). Figure~\ref{fig:distribution_category_split3_split4} details the distribution between visual-question secondary categories.
In Figure~\ref{fig:distribution_questions_per_category}, we present the distribution of questions per category, most of them addressing \textit{right of way} and \textit{signs and markings}.
We also provide a fine-grained breakdown of questions by primary category and dataset split in Figure~\ref{fig:distribution_by_category_per_split}. Some categories are better represented than others depending on the split. For example, split 1 contains more questions related to \textit{driver obligations} and \textit{sanctions and offenses}, while the same categories contain fewer samples in the rest of the splits. For more statistics, see Appendix \ref{app:rod_qa_stats}.

\section{Experiments}

The problem of developing an AI system capable of answering legal questions based on Romanian traffic law can be decomposed into several distinct but interconnected tasks. These tasks span both textual and visual modalities, collectively defining the core components required for building, evaluating, and improving such a system. By segmenting the challenge into modular tasks, we facilitate targeted experimentation, fine-grained performance evaluation, and the possibility of optimizing each sub-component independently.
We can pursue four topics: information retrieval with split 1, question answering with splits 1 and 2,
visual information retrieval with split 3, and visual question answering with splits 3 and 4.

\subsection{Information Retrieval}

An essential component of our system is the ability to retrieve relevant legal text passages that justify the correct answer to a question. Our focus is on maximizing recall within the top-k ranking results ($Recall@k$), as large values of $k$ are impractical for downstream processing and LLM-based generation. Based on the distribution of legal articles that ground the questions (Figure~\ref{fig:distribution_legal_articles} from Appendix~\ref{app:rod_qa_stats}), we consider $k=10$ sufficient for our experiments to provide a relevant context without bloating the LLM input.

In our work, we experiment with the multilingual embedding model mE5\textsubscript{small} \cite{wang2024multilingual}, evaluating different query-building techniques, reranking, and fine-tuning on real or augmented data.
For document representation, we concatenate the article's section title with its content, without chunking. For queries, we concatenate the questions with their answer options. On these inputs, we employ reranking and query rewriting to further improve performance. We use our fine-tuned mE5\textsubscript{small} to embed all the inputs, and we retrieve the top $k=10$ documents using cosine similarity.
The full experimental details can be found in Appendix \ref{sec:ir-details}. We provide the fine-tuned model on HuggingFace\footnote{\EmbeddingURL}.

\subsection{Question Answering}

We utilize a CoT prompting strategy with the GPT-4o mini model, incorporating 10 retrieved documents alongside the question and answer options in the context. This setup reflects a standard RAG pipeline.
To assess the value of retrieval, we experiment with a baseline prompt-based approach that does not employ document retrieval. This allows us to isolate the benefit of retrieval in performance.
Additionally, we evaluated an \textit{ideal RAG} setup in which only the exact relevant documents were included, simulating the ideal recall and precision conditions.

We explore prompt engineering based on observed failure patterns in the model's behavior and propose a better prompt (henceforth denoted BP). The model is therefore tasked with acting as a human expert and providing clear and well-supported responses. The prompt also contains the question, the answer choices, and other relevant information for the task.
We also carried out experiments without the CoT technique to test the role of step-by-step reasoning in answer accuracy. Lastly, we experiment with reasoning-tuned models under the same input specifications and prompts.
We explore reasoning models (o4-mini) \cite{o3-o4mini-system-card} with the same specifications and prompts as before to see if native reasoning helps the problem.
We also assess the performance of open-weight models, such as Mistral Small 3.1 Instruct\footnote{\url{https://mistral.ai/news/mistral-small-3-1}} and Gemma 3 27B Instruct \cite{team2025gemma}, following the same scenarios as with the GPT-4o mini \cite{openai2024gpt4ocard}  model.

\subsection{Visual Information Retrieval}

In the VIR setting, we evaluate the performance of retrieving the correct laws from an image and its associated multiple-choice question. We retrieve and identify the correct traffic signs in the presented image separately.
We adopt several methodologies to enhance text queries with image characteristics. As a baseline, we compare this against a plain QA search.
First, we generate and save 50-100-word captions using the o4-mini with CoT. We propose combining caption (C) and QA (i.e., C + QA). Then, we use the o4-mini model to rewrite our queries and evaluate scenarios that combine images (I), captions, and QA (i.e., I + QA and I + C + QA, respectively). As an extra measure, we also include QA for the latter examples, in case the LLM omits some details.
For traffic sign retrieval testing, a similar methodology is followed. However, we test two embedding scenarios: first, we embed only the traffic sign name and category; second, we also add the explanation.

\subsection{Visual Question Answering}

We assess the performance of CoT prompting on the o4-mini model in three main configurations:

\begin{itemize}[noitemsep,topsep=0pt]
    \item \textbf{Model's prior knowledge}: The model answers questions using just the question, candidate answers, and image, without any external retrieval.
    \item \textbf{Ideal RAG}: The model is provided with exactly the relevant documents or legal references for each question (i.e., perfect retrieval conditions).
    \item \textbf{Best RAG}: The model is provided with the best retrieval methodology from the VIR task for laws and traffic signs.
\end{itemize}

Similar to the VIR task, we propose a methodology for comparing combinations of caption, image, and QA, all leveraging CoT (i.e., C + QA, I + QA, and I + C + QA).

\begin{table*}
    \centering
    \small
    \begin{tabular}{|l|c|c|c|c|c|c|}
        \hline
        \multirow{2}{*}{\textbf{Method}} & \multicolumn{3}{c|}{\textbf{\underline{Retrieval - Train}}} & \multicolumn{3}{c|}{\textbf{\underline{Retrieval - Test}}} \\
         & \textbf{R@10} & \textbf{P@10} & \textbf{nDCG@10} & \textbf{R@10} & \textbf{P@10} & \textbf{nDCG@10} \\
         \hline
         (1) Question based (Q) & 50.15 & 11.45  & 38.75 & 49.29 & 11.01  & 37.48 \\
         (2) Question + Answer Choices (QA) & 60.05 & 14.00  & 51.33 & 59.31 & 13.43  & 51.23 \\
         (3) QA + ReRanker jina & 68.70 & 16.47  & 64.15 & 70.08 & 16.71  & 63.90 \\
         (4) QA + ReRanker bert-msmarco & 63.99 & 15.21  & 56.22 & 65.60 & 15.54  & 56.49 \\
         (5) QA rephrased using GPT 4o-mini & 60.84 & 14.27  & 51.70 & 60.64 & 14.14  & 49.86 \\
         (6) Finetuned Retriever & \textbf{99.89} & \textbf{27.21}  & \textbf{99.71} & \textbf{88.14} & \textbf{23.28}  & \textbf{81.41} \\
         (7) Finetuned Retriever + ReRanker jina & 80.43 & 20.25  & 71.44 & 77.55 & 19.92 & 70.23 \\
         (8) Augmented Finetuned Retriever & 63.80 & 14.80  & 57.10 & 62.76 & 14.53  & 57.43 \\
         \hline
    \end{tabular}
    \caption{Precision@10, Recall@10, and nDCG@10 metric scores for the tried and tested methods on the information retrieval task.}
    \label{tab:full-ir-results}
\end{table*}

\subsection{Evaluation Metrics}
\label{sec:evaluation_metrics}

For information retrieval, we employ Recall@k (R@k), Precision@k (P@k), and normalized discounted cumulative gain at k (nDCG@k)~\cite{jarvelin2002cumulated}. Recall@k measures the proportion of all relevant documents that appear in the top-k retrieved results, while Precision@k evaluates the proportion of retrieved documents in the top-k that are relevant. nDCG@k further incorporates the ranking position, assigning a higher weight to relevant documents that appear earlier in the list. Among these, Recall@k and nDCG@k are the most important, while Precision@k can be misleading when more documents are retrieved than needed.

For QA and VQA tasks, we use Precision, Recall, F1-score, and exact match (EM). Precision and Recall assess prediction accuracy and completeness, while F1 balances both. These metrics also account for partially correct answers. See Appendix~\ref{app:qa_vqa_task_eval} for their definition and interpretation in the context of MCQA. EM is our primary metric, requiring an exact match with the ground truth and not offering partial credit. Given our emphasis on strict answer correctness, EM is the central measure of QA performance. Detailed experimental results for the IR/QA RAG setup can be found in Appendix~\ref{app:additional_experiments}.

\subsection{Experimental Setup}

Details about the experimental setup are presented in Appendix \ref{sec:experimental_setup}.
All the metric comments refer to the exact match score, which is also the performance metric used to evaluate candidates in the MCQA setting.

\section{Results}

\subsection{Information Retrieval}

Our baseline question-based experiment (1) in Table~\ref{tab:full-ir-results} yields only a 50\% Recall@10 performance. Incorporating answer choices increases this performance by almost 10\% in (2), indicating that much of the information needed for the search is also present in the answer choices.
Retrieving 40 documents and re-ranking the top 10 in experiments (3) and (4), we also see a 5\%-10\% boost in performance. Using an LLM to rephrase the query (5) yields results similar to those of QA (2); thus, the problem is semantic, not syntactic.

By fine-tuning our embedding model on the dataset and evaluating only on the test split, we see an 18\% improvement in recall (6). Adding a reranker (7) hampers our performance by 11\% (observed specifically in split 2). This has its roots in the misalignment from which the retriever model also suffered. The augmented fine-tuned model yields only 3\% over its baseline (see (8) compared to (2)), but it underperforms massively compared to the one trained on the data. This suggests a shift in the distribution from the original one: the complexity of the questions is lower, and, as stated earlier, the number of related articles is much smaller.

These findings emphasize the importance of domain-specific representation learning for legal NLP in under-resourced languages and demonstrate that fine-tuning on even modestly sized, targeted datasets can significantly improve performance, more so than generic improvements in model architecture or reranking strategies.

\begin{table}[t!]
\centering
\scriptsize
\setlength{\tabcolsep}{2pt}
\resizebox{\columnwidth}{!}{
\begin{tabular}{|l|c|c|c|}
\hline
\multirow{2}{*}{\centering \textbf{Method}} & \multicolumn{2}{c|}{\textbf{Split 1}} & \textbf{Split 2} \\
\cline{2-4}
                                   & \textbf{Train} & \textbf{Test} & \textbf{Test} \\

\hline
(1) GPT-4o mini + CoT + RAG & 57.8  & 55.5  & 72.9  \\
(2) GPT-4o mini + CoT w/o RAG  & 46.1  & 43.0  & 71.3  \\
(3) GPT-4o mini + CoT + Ideal RAG & 59.0  & 63.3  & 69.6  \\
(4) GPT-4o mini + CoT + RAG + BP & 67.8  & 75.0  & 79.6  \\
(5) GPT-4o mini + RAG + BP w/o CoT & 42.7  & 47.7  & 60.8  \\
(6) o4-mini + CoT + RAG + BP & \textbf{86.3}  & \textbf{91.4}  & \textbf{83.4}  \\
(7) o4-mini + CoT + BP w/o RAG & 64.3  & 63.3  & 82.3  \\
\hdashline
(8) Mistral + CoT + RAG & 42.5  & 42.2  & 51.9  \\
(9) Mistral + CoT w/o RAG  & 46.1  & 39.8  & 68.0  \\
(10) Mistral + CoT + Ideal RAG & 30.4  & 35.2  & 4.4  \\
(11) Mistral + CoT + RAG + BP & 47.6  & 53.9  & 48.1  \\
(12) Mistral + RAG + BP w/o CoT & 57.5  & 53.9  & 75.7  \\
\hdashline
(13) Gemma 3 + CoT + RAG & 60.8  & 53.9  & 75.1  \\
(14) Gemma 3 + CoT w/o RAG   & 48.0  & 38.3  & 65.7  \\
(15) Gemma 3 + CoT + Ideal RAG  & 61.0  & 57.0  & 71.8  \\
(16) Gemma 3 + CoT + RAG + BP  & 67.1  & 53.1  & 80.7  \\
(17) Gemma 3 + RAG + BP w/o CoT  & 55.1  & 46.9  & 72.4  \\
\hline
\end{tabular}}
\caption{Exact match score on IR/QA RAG pipeline. BP indicates that a better prompt was used.}
\label{tab:full-qa-results}
\end{table}
\begin{table*}
    \centering
    \small
    \resizebox{0.75\textwidth}{!}{
    \begin{tabular}{|l|c|c|c|c|c|c|}
        \hline
        \multirow{2}{*}{\textbf{Method}} & \multicolumn{3}{c|}{\textbf{\underline{Retrieval Indicators - Split 3}}} & \multicolumn{3}{c|}{\textbf{\underline{Retrieval Indicators - Split 4}}} \\
         & \textbf{R@10} & \textbf{P@10} & \textbf{nDCG@10} & \textbf{R@10} & \textbf{P@10} & \textbf{nDCG@10} \\
         \hline
         (1) QA & 47.52 & 7.18  & 33.50 & 61.97 & 7.32 & 49.99 \\
         (2) C + QA & 60.49 & 9.39  & \textbf{46.62} & 73.70 & \textbf{9.29} & 56.53 \\
         (3) R[I + QA] & 60.39 & 9.20 & 42.93 & 70.18 & 8.59 & 49.84 \\
         (4) R[I + C + QA] & 57.80 & 8.82 & 43.66 & 67.37 & 8.45 & 50.87 \\
         (5) R[I + QA] + QA & 60.23 & 9.24 & 46.14 & 73.47 & 9.01 & 56.29\\
         (6) R[I + C + QA] + QA & 58.25 & 8.92 & 45.15 & \textbf{74.41} & 9.15 & \textbf{60.38} \\
         (1*) QA & 41.98 & 6.13 & 28.72 & 64.08 & 7.60 & 48.38 \\
         (2*) C + QA & \textbf{63.13} & \textbf{9.65} & 44.96 & 69.95 & 8.59 & 53.13 \\
         (3*) R[I + QA] & 58.80 & 9.05 & 41.67 & 63.14 & 7.46 & 50.11 \\
         (4*) R[I + C + QA] & 57.25 & 8.95 & 39.95 & 68.54 & 8.16 & 50.63 \\
         (5*) R[I + QA] + QA & 58.93 & 8.98 & 41.93 & 70.89 & 8.59 & 54.89\\
         (6*) R[I + C + QA] + QA & 59.34 & 9.11 & 42.13 & 72.06 & 8.59 & 56.11 \\
         \hline
    \end{tabular}}
    \caption{Precision@10, Recall@10, and nDCG@10 metric scores for the visual information retrieval of indicators task. Notation: QA -- question and answer choices, I -- image, C -- caption, \texttt{R[...]} -- rephrasing using o4-mini, * -- corpus has different content embedded into vectors.}
    \label{tab:full-vir-results-indicators}
\end{table*} 
\begin{table}
    \centering
    \scriptsize
    \setlength{\tabcolsep}{2pt}
    \resizebox{0.93\columnwidth}{!}{
    \begin{tabular}{|l|c|c|c|}
        \hline
        \multirow{2}{*}{\textbf{Method}} & \multicolumn{3}{c|}{\textbf{\underline{Retrieval Laws - Split 3}}} \\
         & \textbf{R@10} & \textbf{P@10} & \textbf{nDCG@10} \\
         \hline
         (1) QA & 60.45 & 12.72  & 56.18 \\
         (2) C + QA & 70.30 & 15.06  & 56.41 \\
         (3) R[I + QA] & 73.51 & 16.13  & 58.38 \\
         (4) R[I + C + QA] & 67.70 & 14.87  & 52.85 \\
         (5) R[I + QA] + QA & \textbf{77.09} & \textbf{16.83}  & \textbf{62.67} \\
         (6) R[I + C + QA] + QA & 75.17 & 16.32  & 59.67 \\
         \hline
    \end{tabular}}
    \caption{Precision@10, Recall@10, and nDCG@10 scores for the visual information retrieval of laws task. Notation: QA -- question and answer choices, C -- caption,  I -- image, \texttt{R[...]} -- rephrasing using o4-mini.}
    \label{tab:full-vir-results-laws}
\end{table}

\subsection{Question Answering}
\label{sec:qa_results}

Our first three experiments (1), (2), and (3) in Table~\ref{tab:full-qa-results} aim to compare the retrieval component. In the first split, we observe that the RAG component improves performance by 11\%, and the difference between the best setup and an ideal RAG is marginal. This shows that even when not all documents are retrieved, or some are over-retrieved and irrelevant, the LLM can complete missing information and respond similarly, with only a 2\% observed variation.

The error analysis revealed three main failure modes (see also Appendix \ref{sec:appendix-qa} for a more in-depth analysis):
\begin{itemize}[noitemsep,topsep=0pt]
\item \textbf{Difficult Questions}: Difficulty handling nuanced or misleading options.
\item \textbf{Safety Bias}: Prioritizing safe answers over legally correct ones.
\item \textbf{Overthinking}: Overextending reasoning beyond the immediate question scope.
\end{itemize}

With this refinement in mind, using a better prompt to mitigate these findings, we achieve another 9\% improvement over the previous results in strategy (4). In this step, we also want to ablate the CoT, making the model respond directly. This shows a 25\% loss in performance in experiment (5).

By adding a reasoning model to the previous experiments, we achieve another 19\% increase in performance for the strategy (6). Ablating the retriever also shows that it remains relevant, as we lose 22\% in strategy (7).

Experiments (8)-(12) with Mistral show similar patterns but overall weaker results. One thing to note is that this model sometimes did not follow the instructions, either by answering with the responses first and then arguing, or by entering an infinite loop of generating additional laws instead of answering. Also, for the experiments where we ablate CoT, the model would perform the same or better by 10-30\%, depending on the dataset used.

Gemma achieves in (13)-(17) results comparable to those of OpenAI's non-reasoning models. We discuss the comparison between open- and closed-source models, the potential answer bias introduced by our prompt, and the difficulty of the dataset in Appendix \ref{sec:discussions}.


\subsection{Visual Information Retrieval}

Using the fine-tuned retriever, our text-based baseline yields a 60\% recall in retrieval, which is a strong initial result (strategy (1) in Table~\ref{tab:full-vir-results-laws}). Adding the caption to the search query improves the performance by 10\%.
Furthermore, adding image information via an LLM to rephrase the query (entries denoted with \texttt{R[...]} represent rephrasing of the query using a VLM) improves recall by 3\% in the I + QA scenario and results in a loss of 3\% when also using the caption. Contrary to the belief that more context helps the model to produce a better query, it does not in this scenario.
Concatenating the QA pair with the previous experiments also shows a 7-8\% improvement in both scenarios, suggesting that some details were left out of the rephrasing.
Looking at the retrieval of the traffic signs in Table~\ref{tab:full-vir-results-indicators}, with the same experiments in mind, they follow a similar pattern in performance.
When changing the embedding method to include more context and details for the traffic signs (the entries marked with a *), we see a slight improvement in the C + QA scenario. This combination helps match additional details in the caption with those in the description, yielding a 3\% gain in strategy (2*) over the previous best.

\subsection{Visual Question Answering}

\begin{table}[t!]
\centering
\scriptsize
\setlength{\tabcolsep}{2pt}
\resizebox{\columnwidth}{!}{
\begin{tabular}{|l|c|c|}
\hline
\textbf{Method} & \textbf{Split 3} & \textbf{Split 4} \\
\hline
(1) o4-mini + C + QA + CoT  & 64.2  & 71.8  \\
(2) o4-mini + I + QA + CoT   & 71.5  & 74.6  \\
(3) o4-mini + I  + C + QA + CoT  & 64.9  & 74.6  \\
(4) o4-mini + C + QA + CoT + Ideal RAG & 69.9  & 78.9  \\
(5) o4-mini + I + QA + CoT + Ideal RAG  & \textbf{77.8 } & 78.9  \\
(6) o4-mini + I  + C + QA + Ideal RAG & 71.8  & 78.9  \\
(7) o4-mini + C + QA + CoT + RAG & 67.4  & 76.1  \\
(8) o4-mini + I + QA + CoT + RAG  & 75.6  & \textbf{90.1 } \\
(9) o4-mini + I  + C + QA + RAG & 69.3  & 77.5  \\
\hline
\end{tabular}}
\caption{Exact match scores on the VIR/VQA RAG pipeline using strategy (5) for laws and strategy (2*) for indicators. Notation: QA -- question and answer choices, I -- image, C -- caption.}
\label{tab:full-vqa-results}
\end{table}

We aim to evaluate VQA across three prompting and three retrieval scenarios, as shown in Table~\ref{tab:full-vqa-results}.
First, we want to test the input combinations as before: C + QA, I + QA, or I + C + QA.
Consistently, I + QA has a better performance than C + QA by 7-8\%, while I + C + QA is only 2-3\% above C + QA. This is a similar pattern we spotted in the retrieval task, namely, adding more context (i.e., the caption) does not improve performance, but it does affect it.
Upon checking the retrieval component, we find that the theoretical best and our best setup yield comparable results (1-2\% difference), whereas ablating RAG results in at least a 5\% performance loss.

\subsection{Additional Results Analysis}

We present QA examples in which a better prompting strategy improved the results and mitigated the initial limitations observed in Appendix \ref{sec:appendix-qa}. We also present examples of incorrect visual questions for each of the three secondary categories using the best RAG setup.

We add a more in-depth error analysis in Appendix \ref{sec:appendix-error-analysis}, focusing on performance across all tasks, splits, and categories. We examine the tendency to overselect answers, which directly reduces precision and exact matches, and the number of reasoning steps used by LLMs, which indicate the categories of questions that are more difficult overall.

We discuss hallucinations in Appendix \ref{sec:hallucinations-analysis}. We analyze citation counts and show that the model tends to hallucinate documents when citing more documents per output, thereby degrading performance. We analyze the instruction-following of the models and discover higher failure rates for output formatting in Mistral 3.1, leading to a performance downgrade due to the inability to parse outputs; and higher failure rates for citing in the indicated way on OpenAI o4-mini, while maintaining superior performance. We analyze hallucinations in captions and demonstrate their direct correlation with performance degradation. Finally, using LLM-as-a-Judge \cite{zheng2023judging}, we identify and classify hallucinations, revealing failure patterns in the legal context.

\section{Conclusions}

In this work, we evaluated information retrieval, visual information retrieval, question answering, and visual question answering both independently and in combination on a newly introduced dataset. Our results demonstrated promising performance on tasks and highlighted specific areas for improvement. In the IR component, further fine-tuning was required to prevent the inclusion of actual positives during hard-negative mining. For VIR, future work should explore more advanced methods that incorporate joint image–text embeddings. In QA and VQA, improved prompting strategies or targeted model fine-tuning could help mitigate the limitations and biases inherent to current large language models.

Additionally, we examined the challenges LLMs face in multiple-choice QA settings, particularly the importance of minimizing over-selection to maximize precision. This research makes a novel contribution to the intersection of QA, VQA, IR, and VIR in the Romanian legal domain, specifically focused on traffic law, with potential applicability to a wide range of legal tasks.

\section*{Limitations}


\textbf{Text Truncation.} The choice of not splitting larger documents into sub-chunks and truncating them instead aimed to simplify our experiments and benchmarking of the dataset, and we acknowledge that it might set us back from achieving a better retrieval result.

\textbf{Evaluation of Small LLMs.} We also acknowledge that we did not run the experiments with larger models, which could have improved the scores, due to hardware or budget constraints. For the same reasons, we did not run the experiments on multiple seeds to explore variation.

\section*{Risks and Ethical Considerations}

The only risk we see is in educational settings. LLMs are known to suffer hallucinations \cite{openai2024gpt4ocard}, which we also address in  Appendix \ref{sec:hallucinations-analysis}. If users were to learn for the driving exams using our setup, we cannot guarantee that the information is correct, as reflected in the imperfect results. In addition, our methods cannot be used for legal advice.

Since law may change in the future, and as discussed in Appendix \ref{sec:ir-details}, the performance of the fine-tuned retriever model is not impacted if case law articles are repealed, by simply removing them from the corpus. We provide, under the MIT license, the code for scraping, cleaning, and normalizing articles from official sources, to benefit from the most recent changes in the law while using our experimental setup.

 \section*{Acknowledgments}
 We would like to thank "Școala Rutieră"\footnote{\href{https://www.scoalarutiera.ro/}{Școala Rutieră}} for providing access to the dataset used in this study. Their support made this work possible, and they have kindly allowed it to be used for academic purposes now and in the future. This work was (partially) supported by RExQTCS: Romanian Excellence Quantum Technologies enhancing Cybersecurity, a grant of the Ministry of Education and Research, CCCDI - UEFISCDI, project number PN-IV-P6-6.1-CoEx-2024-0214, within PNCDI IV.

\bibliography{custom}

\clearpage
\appendix

\section{Dataset Statistics}
\label{app:dataset-statistics}

In this section, we present statistics of the RoD-TAL dataset, which comprises two main components: the legal corpus RoD-Law and the multiple-choice question-answering dataset RoD-QA. We analyze the characteristics and distributions of each component to provide insights into the structure and content of the dataset.

\subsection{RoD-Law Statistics}
\label{app:rod_law_stats}

\begin{figure}[!ht]
\centering
\includegraphics[width=\columnwidth]{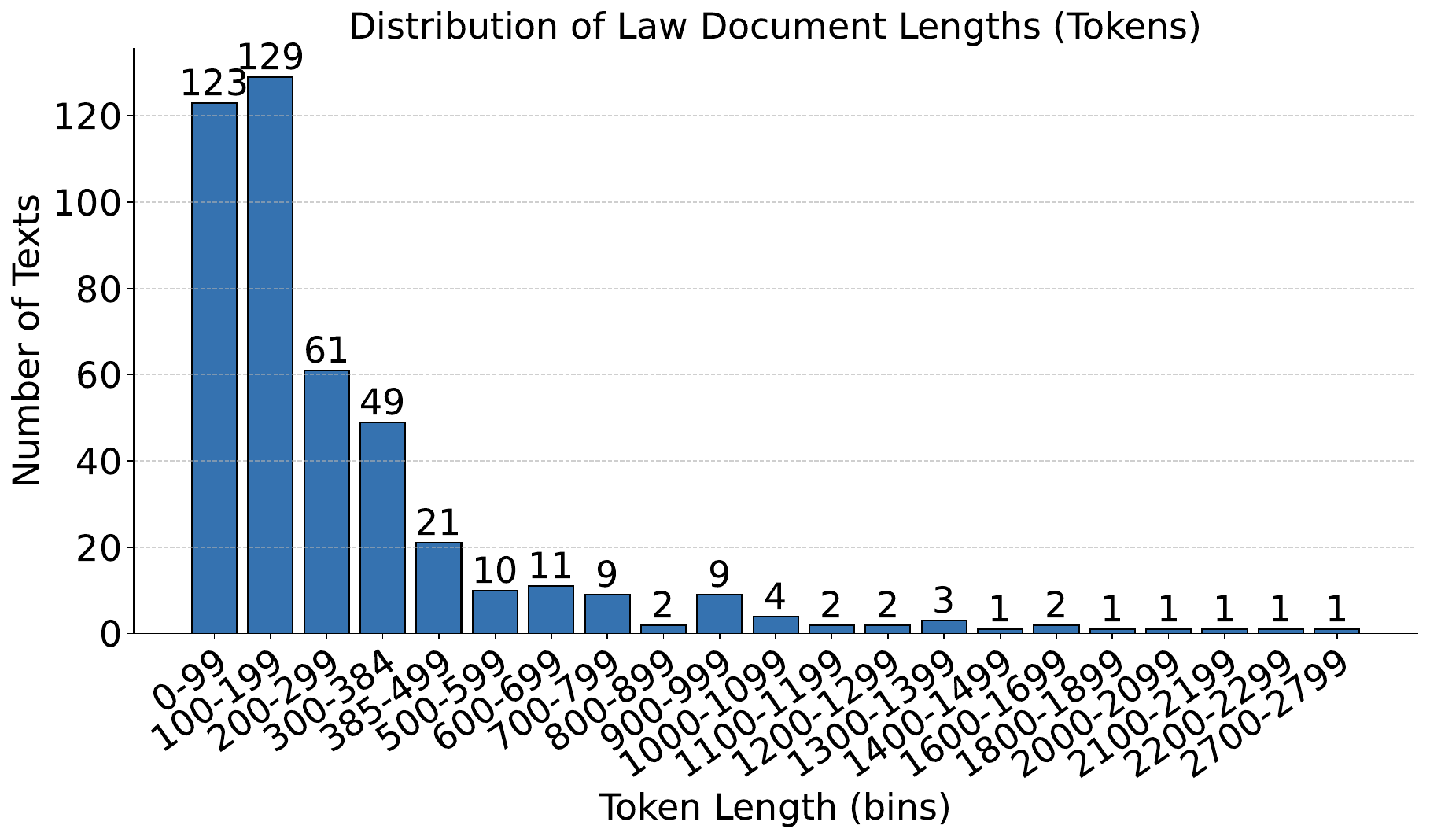}
\caption{Tokens distribution on the entire RoD-Law corpus.}
\label{fig:token_length_distribution}
\end{figure}

The corpus totals 443 legal documents from Romanian law. In Figure \ref{fig:token_length_distribution}, we present the distribution of the number of tokens in the entire RoD-Law corpus. The distribution roughly follows a long-tail power law, with most documents containing fewer tokens; approximately 84\% contain text under 500 tokens.

\begin{figure}[!ht]
\centering
\includegraphics[width=\columnwidth]{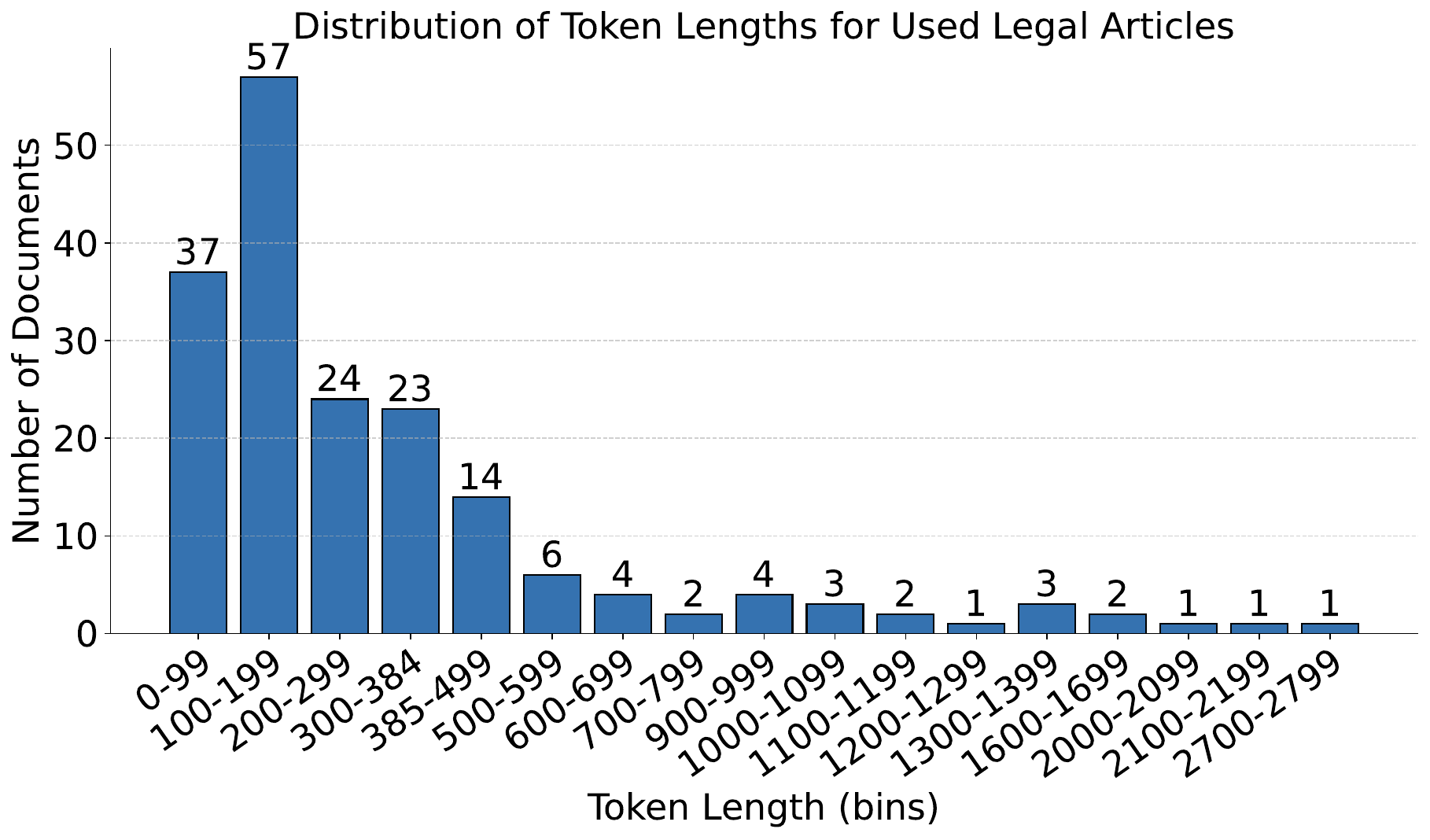}
\caption{Tokens distribution on the documents referenced in RoD-QA.}
\label{fig:token_length_distribution_used_legislation}
\end{figure}

Of the 443 documents, 185 have references in the QA dataset. We present the token distribution in Figure \ref{fig:token_length_distribution_used_legislation}. Similarly, 70\% of the documents could enter a context window of 500 tokens.

\begin{figure}[!t]
\centering
\includegraphics[width=\columnwidth]{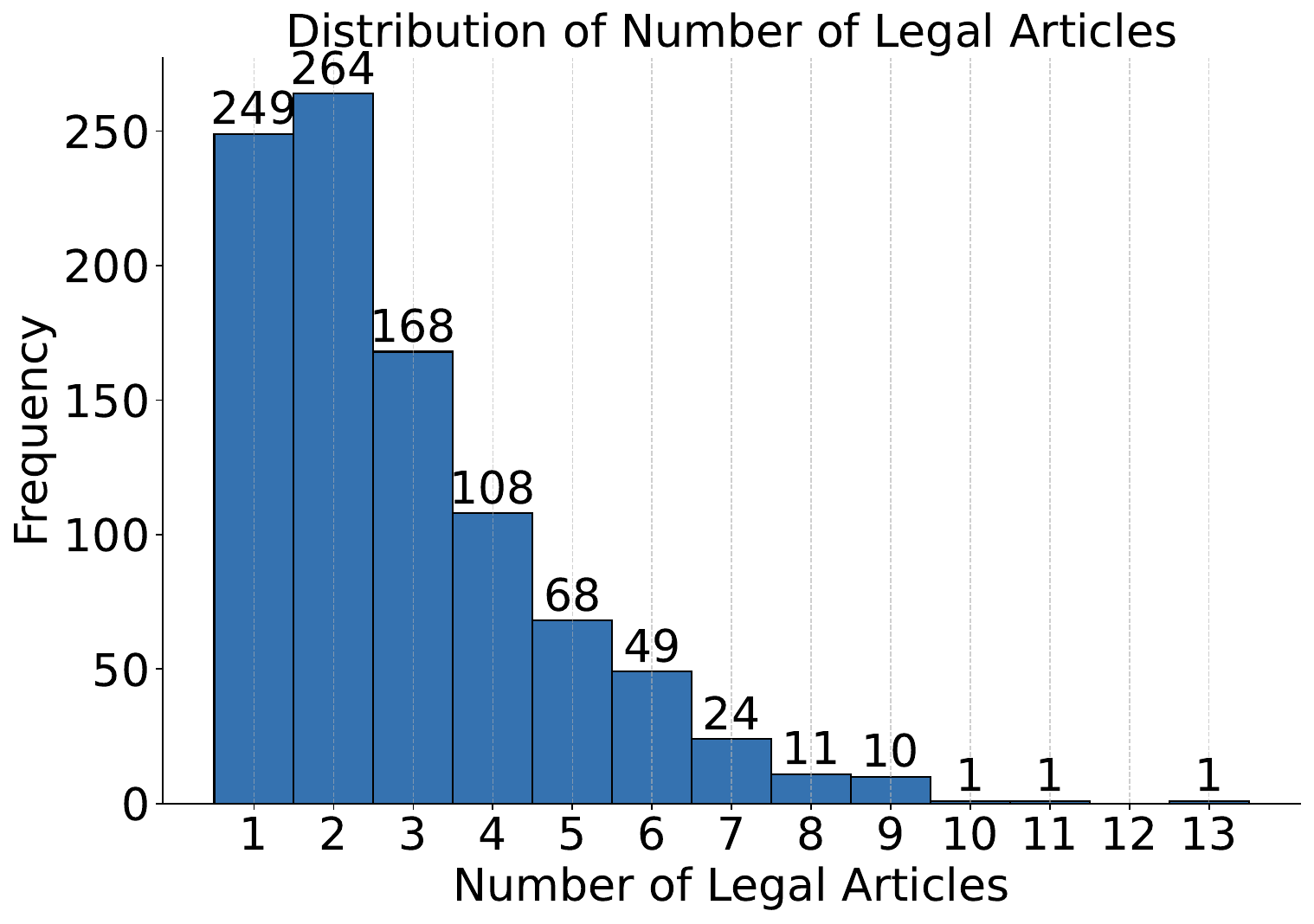}
\caption{Distribution of legal article references per question.}
\label{fig:distribution_legal_articles}
\end{figure}

\begin{figure}[!t]
\centering
\includegraphics[width=\columnwidth]{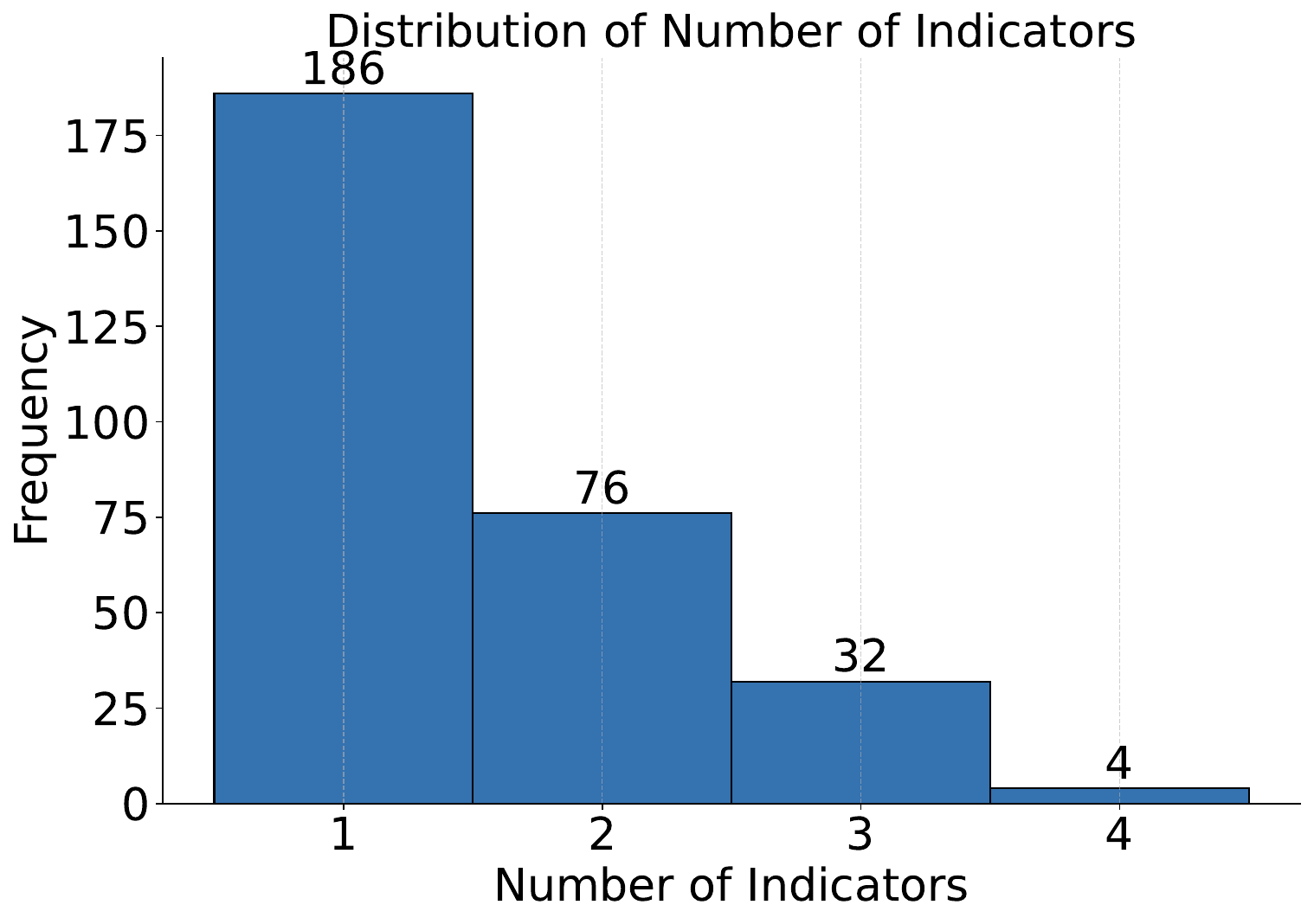}
\caption{Distribution of traffic sign indicators per question.}
\label{fig:distribution_indicators}
\end{figure}

\begin{figure*}[htbp]
\centering
\includegraphics[trim=0cm 1cm 0cm 1cm,clip,width=\textwidth]{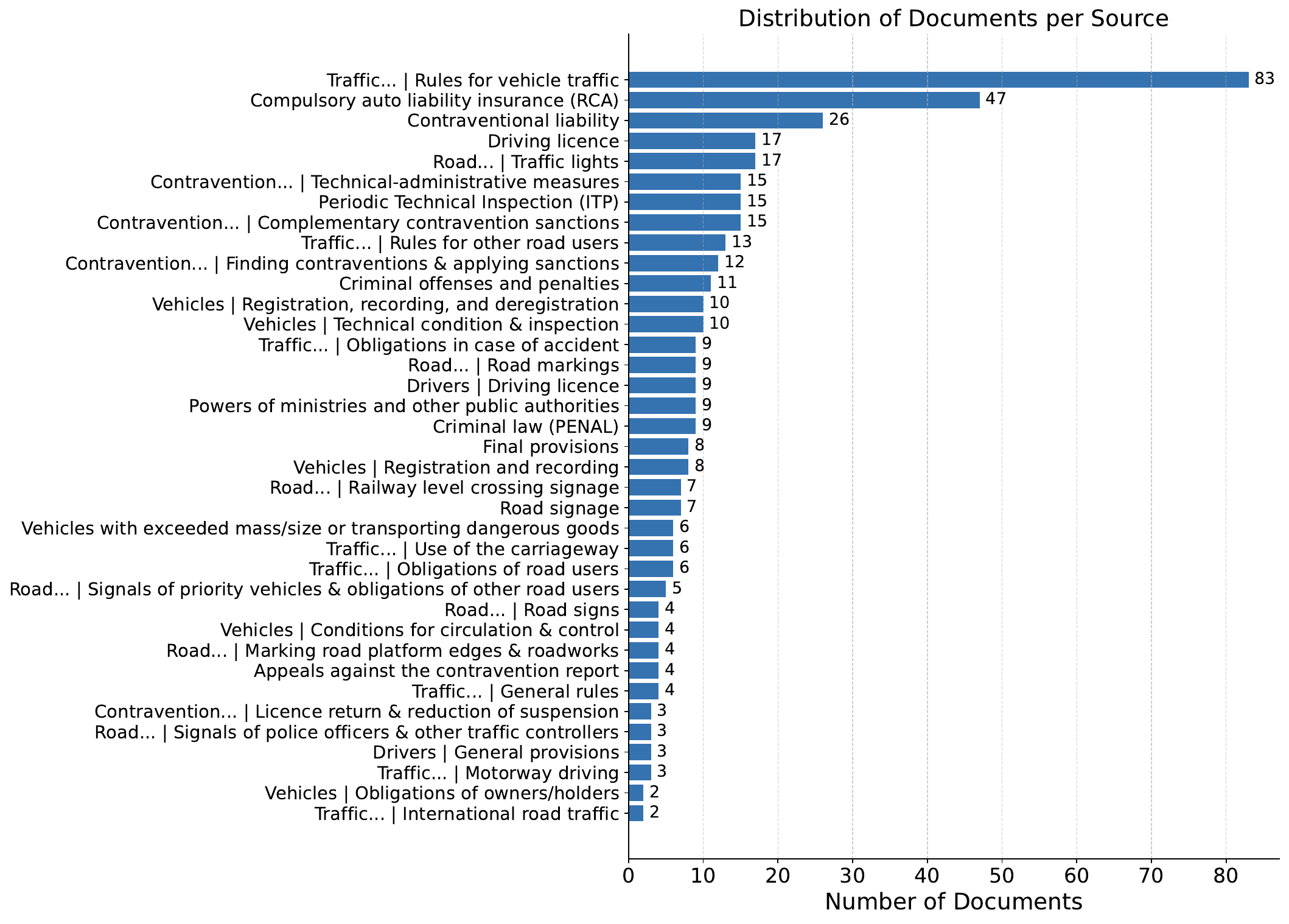}
\caption{Distribution of legal documents per topic.}
\label{fig:distribution_documents_per_source}
\end{figure*}

\subsection{RoD-QA Statistics}
\label{app:rod_qa_stats}

The QA dataset comprises 1,156 samples across text and image modalities, with and without legal reference annotations, as presented in \S\ref{sec:rod_qa_dataset}.

For the data annotated with legal references, we illustrate in Figure \ref{fig:distribution_legal_articles} the distribution per question. Most of the questions contain up to 10 references, which is also the main motivation to set $k=10$ documents retrieved during experiments. In Figure \ref{fig:distribution_indicators}, we present the distribution of the number of traffic signs illustrated in the image-based questions, where most of the questions show up to three traffic signs.

\subsection{Skewed Distribution of Questions per Categories}

We show the distribution of legal documents per topic in Figure \ref{fig:distribution_documents_per_source}. Compared to Figure \ref{fig:distribution_questions_per_category} from \S\ref{sec:rod_qa_dataset}, where certain categories have fewer samples, such as \textit{eco-driving}, \textit{highway driving}, or \textit{railroad crossing}, we want to find a similar mapping to the distribution of the laws. For example, the \textit{motorway driving} only has three legal articles associated with it. The \textit{railway level crossing signage} has seven articles. Also, maneuvers in \textit{rules regarding maneuvers} in Romanian legislation are a small set of actions (i.e., overtaking, stopping, stationary, parking, turning 180, and driving backward). They involve only 19 different articles across all the questions, whereas we have 27.

We have also checked the coverage of the RoD-TAL corpus (which includes all the existing articles in the useful law), and the 1,156 driving questions cover only 185 unique articles out of 443 (41\%). 

The official guidelines as to how the driving tests are created can be found in the MAI Order no. 268/2010 Article 10\footnote{\url{https://legislatie.just.ro/Public/DetaliiDocument/124490}}. Paragraph (4) explicitly describes the content of the theoretical questions. The first points refer to general driving rules (for example, (a) states about legal provisions on road traffic such as road signs, priority rules, and speed restrictions), while the last points refer to vehicle safety equipment (point (h)) and environmental protection (point (i)). The ordering may also illustrate the importance.

Given this information, we conclude the following:
\begin{itemize}
    \item The underrepresented categories lack many associated laws, resulting in non-repetitive questions.
    \item The driving tests reflect more real-world situations and, therefore, will include more questions from those situations.
\end{itemize}

\begin{table*}[!t]
    \centering
    \small
    \resizebox{\textwidth}{!}{
    \begin{tabular}{|l|c|c|l|}
        \hline
        \textbf{Model} & \textbf{Num. of Params.} & \textbf{Ctx. Size} & \textbf{Checkpoint} \\
        \hline
        mE5\textsubscript{small} & 118M & 512 & \href{https://huggingface.co/intfloat/multilingual-e5-small}{\texttt{multilingual-e5-small}} \\
        Passage Reranking Multilingual BERT & 168M & 512 & \href{https://huggingface.co/amberoad/bert-multilingual-passage-reranking-msmarco}{\texttt{bert-multilingual-passage-reranking-msmarco}} \\
        Jina Reranker v2 & 278M & 1024 & \href{https://huggingface.co/jinaai/jina-reranker-v2-base-multilingual}{\texttt{jina-reranker-v2-base-multilingual}} \\
        Mistral Small 3.1 & 24B & 128k & \href{https://huggingface.co/mistralai/Mistral-Small-3.1-24B-Instruct-2503}{\texttt{Mistral-Small-3.1-24B-Instruct-2503}} \\
        Gemma 3 27B Instruct  & 27B & 128k & \href{https://huggingface.co/google/gemma-3-27b-it}{gemma-3-27b-it} \\
        GPT-4o mini & undisclosed & 128k & gpt-4o-mini-2024-07-18 \\
        o4-mini (medium) & undisclosed & 200k & o4-mini-2025-04-16 \\
        \hline
    \end{tabular}}
    \caption{Model checkpoints used during experiments.}
    \label{tab:checkpoints}
\end{table*}

A particular trait of our dataset is that it contains only the official driving test questions, as stated in the FAQ Q1\footnote{\url{https://www.scoalarutiera.ro/intrebari-frecvente.html}} of \textit{Scoala Rutiera}. Therefore, our question distributions are skewed due to (1) having only official questions and (2) having a focus on more daily situations rather than complete coverage of the law.

\section{Experimental Setup}
\label{sec:experimental_setup}


\subsection{Model Checkpoints}
\label{sec:model_checkpoints}

Table \ref{tab:checkpoints} presents the models and checkpoints used during the experiments. We indicate the size as the number of parameters, the context size in terms of tokens, and the checkpoint on HuggingFace\footnote{\url{https://huggingface.co/}} or the OpenAI platform\footnote{\url{https://platform.openai.com}}.

\subsection{Information Retrieval Setting}
\label{sec:ir-details}

\textbf{Embedding model.} We initially employ the multilingual model mE5\textsubscript{small} \cite{wang2024multilingual} to generate dense text embeddings for the legal corpus and user queries. This model supports Romanian and offers a sufficiently large context window to embed the relatively long articles typical of traffic law. It is also ranked among the top models on the MTEB leaderboard \cite{muennighoff2022mteb} for retrieval tasks while supporting Romanian.
We embed each article by concatenating the title metadata and content. We truncate the content if it exceeds the maximum context of 512, avoiding splitting it into smaller chunks. During QA setups, when we use RAG, even if the retrieval step is performed on the first 512 tokens, we still include the entire document in the generation step. This decision does not affect the end result.

\textbf{Query formatting.} To improve retrieval performance, we experimented with various input formulations, including concatenating the question with its answer options, which led to better embedding-based similarity scores. 

\textbf{Reranking.} Further improvements were sought by incorporating multilingual reranking models: Jina Reranker v2\footnote{\href{https://jina.ai/news/jina-reranker-v2-for-agentic-rag-ultra-fast-multilingual-function-calling-and-code-search/}{Jina Reranker V2}} and Passage Reranking Multilingual BERT \cite{nogueira2020passagererankingbert}.

\textbf{Query rewriting.} We also explore query rewriting using LLMs, specifically GPT-4o mini via API \cite{openai2024gpt4ocard}, attempting to rephrase questions in a way that aligns with the embedding model's representation space.

\textbf{Fine-tuning retriever}. Following poor results with the previous techniques, we hypothesize that domain and language mismatches are a core bottleneck, particularly in the specialized, underrepresented Romanian legal language.
To address this, we fine-tune the mE5\textsubscript{small} model on our dataset for 10 epochs, 80\%-20\% train-test split, using the InfoNCE loss \cite{oord2018representation}, consistent with the model's original training regime. We constructed a training set of approximately 6,960 samples, comprising positive pairs (i.e., questions with their correct legal references), and 
hard negatives (i.e., derived from top candidates retrieved by the base model, but judged incorrect -- positive aware hard-negative mining, 5 each).
 We also test reranking on top of the fine-tuned retriever. This fine-tuning process was designed to fill the knowledge gap in the retriever. We believe that if certain articles are repealed in the future, this will not affect retrieval, as they will simply be removed from the retrieval corpus. The retriever learns how to align driving questions and legal articles, not overfitting much, as shown in the results between the Split 1 train \& test.

The fine-tuned model can be found on HuggingFace\footnote{\EmbeddingURL}.

\textbf{Data augmentation.} We further experiment with data augmentation via LLM-based synthesis. A few-shot prompt was created (two examples of documents with associated questions and answers), and 1,000 sets of 2–6 legal documents were sampled. GPT-4o mini is used to generate 5 QA pairs per set. However, the LLM typically uses only 1–2 references per question, likely due to contextual incompatibility. After removing duplicates and entries with a similarity score over 0.98, we obtain 2,259 valid pairs, totaling 14,055 training samples. For this augmented dataset, we apply the same contrastive-based fine-tuning regime. The goal is not to leak any of the dataset distribution or statistics to our retriever, and to validate based on the entire initial dataset.

\subsection{Hyperparameters}
\label{sec:hyperparams}

For fine-tuning the dense retriever model, we largely follow the original work's settings, with several modifications specific to our setup. The model is trained using the Sentence Transformers library \cite{reimers-gurevych-2019-sentence}, with the following hyperparameters:

\begin{itemize}
\item \textbf{Number of epochs}: We train the first version of the model for 10 epochs, and the second one is early stopped after 1 epoch due to poor performance.
\item \textbf{Batch size}: Both the training and evaluation batch sizes are set to 64.
\item \textbf{Learning rate schedule}: We use a warmup ratio of 0.1.
\item \textbf{Mixed precision}: Training is performed with FP16 precision enabled, while BF16 is disabled.
\item \textbf{Batch sampler}: We use \texttt{BatchSamplers.NO\_DUPLICATES} to avoid duplicate samples within a batch, which is beneficial for in-batch negative sampling losses.
\item \textbf{Loss function}: The MultipleNegativesRankingLoss (InfoNCE) is used, with mE5\textsubscript{small} as the base model.
\item \textbf{Evaluation and saving}: The model is evaluated and checkpoints are saved every 100 steps, with only the best two checkpoints retained.
\item \textbf{Early stopping}: The best model is loaded at the end of training based on the evaluation metric (\texttt{eval\_cosine\_recall@10}), with \texttt{greater\_is\_better=True}.
\item \textbf{Random seed}: To ensure reproducibility, all relevant random seeds (\texttt{torch}, \texttt{numpy}, \texttt{random}, and \texttt{transformers}) are set to 42 and deterministic training options are enabled.
\end{itemize}

\subsection{LLM Prompts}
\label{sec:appendix-prompts}

To maintain readability and keep the paper focused on analysis and results, we make our LLM prompts, both in Romanian and English, publicly available in the GitHub repository\footnote{\GitHubURL}. In the following, we provide several prompts translated into English.

\textbf{QA Setup.}
We use the following prompt with the variables for each question pair, using an LLM such as OpenAI's o4-mini (medium) via the API or Gemma 3 27B Instruct via VLLM, with temperature=0 and seed=25.

\begin{promptbox}{English Version: Enhanced QA prompt}
You are a traffic police officer. You only speak Romanian.\\
You need to solve a multiple-choice question from a driving test. The question may have one or more correct answers. You will use only Romanian laws.

Think logically, but do not extrapolate beyond the information provided. Judge only the described moment; do not assume other situations.

Thinking rules:

1. Read the question and answer choices very carefully.

2. Strictly identify which provisions of Romanian traffic legislation apply to the given situation.

3. If the answer seems "safer" but is contrary to the legislation, follow the law, not instinct.

4. Select ONLY the answers that are completely correct according to the letter of the law --- do not guess, do not add missing information.

5. If one correct answer is better than another marked as correct, includes more specific situations or exceptions, then only that one should be chosen.

6. Clearly argue why you chose each correct answer. If there are multiple correct answers, explain each choice separately.

7. Pay attention to small details that can change the meaning of the question or answers (some questions are trick questions).

At the end, the last part of your answer must be the correct letter or letters.\\
For example, your answer should end with:

"Correct answer: A"

or

"Correct answer: A,B"

This is the question:

\{question\}

These are the answer choices:

\{answers\}

These are the relevant laws, but not all may be relevant:

\{documents\}

===================
\end{promptbox}

\textbf{VIR Setup.}
Similarly to the previous setup, we use OpenAI's o4-mini (medium) due to its vision capabilities, with the same hyperparameters.

\begin{promptbox}{English Version: VIR by Image + QA + Caption}
You are a traffic police officer. You only speak Romanian.\\
You receive a multiple-choice question from a driving test, which also has an attached image. You need to select the necessary information from the image, so as to improve the original question and facilitate the search for relevant legal articles.

Include relevant information about the situation, road signs, and other elements specific to driving and the law.

Thinking rules:

1. Read the question and answer options very carefully.

2. Analyze the image and extract the most important information.

3. Pay attention to small details that may change the meaning of the question or the answers.

4. Explain what should be added to the question to help search for the correct information.

At the end, the last part of your answer must be the reformulated question.

"Final answer: [question]"

This is the question:

\{question\}

These are the answer options:

\{answers\}

This is the image caption:

\{caption\}

===================
\end{promptbox}

\textbf{VQA Setup.}
We also employ OpenAI's o4-mini (medium) using the same hyperparameters as in the prior settings.
 
\begin{promptbox}{English Version: Prompts for VQA with RAG}
You are a traffic police officer. You only speak Romanian.\\
You need to solve a multiple-choice question from a driving test. The question may have one or more correct answers. You will use only the laws from Romania.\\
You will receive a question along with an image; the question is closely related to the image.

Think logically, but do not extrapolate beyond the provided information. Judge only the described moment; do not assume other situations.

Thinking rules:

[...]

These are the relevant laws, but not all may be relevant:

\{documents\_laws\}

\{documents\_indicators\}

===================
\end{promptbox}

\subsection{Hardware Infrastructure and Computational Costs}

We train the open-source embedding model in Google Colab on a single NVIDIA A100 GPU and perform inference on the open-source LLMs using the \textit{vllm} package \cite{kwon2023efficient} on 4 NVIDIA A100 GPUs from the institutional cluster, thereby improving scalability. 

For the closed-source LLM experiments, the GPT-4o mini model is used with a seed set to 25 and a temperature of 0. The o4-mini model is used without the option to set the temperature or seed, for now. The total cost for LLM experiments using the OpenAI API\footnote{\url{https://platform.openai.com/}} is \$115.72.

\subsection{QA and VQA Tasks Evaluation}
\label{app:qa_vqa_task_eval}

In \S\ref{sec:evaluation_metrics}, we present that the exact match metric is used mainly to evaluate LLMs, similar to how a candidate is assessed during a driving test. The reason is that during the exam, candidates are not given partial scores for missing an option or selecting another one. This would also be critical to driving capabilities, as giving the wrong answer would mean that the candidate made a mistake and does not understand the law, which could lead to dangerous situations. Precision is paramount in this context.

However, in the context of LLMs, \textit{partially correct} answers are an important aspect of benchmarking models \cite{kalai2025languagemodelshallucinate}. Therefore, we also provide precision (P), recall (R), and F1 scores (F1) in Appendix \ref{app:additional_experiments}, which account for partially correct answers. In the following, we describe how we compute these metrics.

For each question, we compute:

\begin{itemize}
  \item $TP$ -- number of predicted answers that are actually correct
  \item $FP$ -- number of predicted answers that are not correct
  \item $FN$ -- number of correct answers that were missed
\end{itemize}

The metrics for each question are calculated as follows:

\begin{gather}
P_{question} = \frac{TP}{TP + FP} \\
R_{question} = \frac{TP}{TP + FN} \\
F1_{question} = 2 \cdot \frac{P_{question} \cdot R_{question}}{P_{question} + R_{question}}
\end{gather}

The overall metrics are then computed as the average over all questions, where $N$ is the total number of questions:

\begin{gather}
P = \frac{1}{N} \sum_{i=1}^{N} P_{question_i} \\
R = \frac{1}{N} \sum_{i=1}^{N} R_{question_i} \\
F1 = \frac{1}{N} \sum_{i=1}^{N} F1_{question_i}
\end{gather}

\begin{table*}[!t]
\centering
\begin{tabular}{|l|l|c c|c c|c c|}
\hline
\multirow{2}{*}{\textbf{Model}} & \multirow{2}{*}{\textbf{Split}} & \multicolumn{2}{c|}{\textbf{Answer A}} & \multicolumn{2}{c|}{\textbf{Answer B}} & \multicolumn{2}{c|}{\textbf{Answer C}} \\
 &  & \textbf{P (\%)} & \textbf{R (\%)} &\textbf{ P (\%)} & \textbf{R (\%)} &  \textbf{P (\%)} & \textbf{R (\%)} \\
\hline
Gemma 3  & 1\_test & 33.68 & 33.56 & 36.32 & 36.91 & 30.00 & 29.53 \\
Gemma 3 & 1\_train & 34.31 & 34.99 & 34.89 & 34.99 & 30.80 & 30.02 \\
Gemma 3 & 2 & 34.78 & 33.17 & 40.10 & 41.21 & 25.12 & 25.63 \\
Mistral  & 1\_test & 34.31 & 33.56 & 37.96 & 36.91 & 27.74 & 29.53 \\
Mistral  & 1\_train & 35.22 & 34.99 & 32.59 & 34.99 & 32.19 & 30.02 \\
Mistral  & 2 & 33.70 & 33.17 & 40.76 & 41.21 & 25.54 & 25.63 \\
o4-mini  & 1\_test & 33.33 & 33.56 & 36.00 & 36.91 & 30.67 & 29.53 \\
o4-mini  & 1\_train & 34.58 & 34.99 & 35.09 & 34.99 & 30.32 & 30.02 \\
o4-mini  & 2 & 33.49 & 33.17 & 41.04 & 41.21 & 25.47 & 25.63 \\
\hline
\end{tabular}
\caption{Predicted (P) vs. reference (R) distribution per model and split considering option letter frequency.}
\label{tab:picked_ref_distribution}
\end{table*}

To illustrate how these metrics work, we provide some examples below. Suppose that for a given question, the correct answers are $\{A, B\}$. Then, if:

\begin{itemize}
  \item prediction $= \{A\} \rightarrow TP = 1, FP = 0, FN = 1 \rightarrow precision = 1.0, recall = 0.5, F1 = 0.667$ (partially correct)
  \item prediction $= \{A, C\} \rightarrow TP = 1, FP = 1, FN = 1 \rightarrow P = 0.5, R = 0.5, F1 = 0.5$ (partial with a penalty)
  \item prediction $= \{A, B, C\} \rightarrow TP = 2, FP = 1, FN = 0 \rightarrow P = 0.667, R = 1.0, F1 = 0.8$ (complete but with an extra wrong choice)
  \item prediction $= \{A, B\} \rightarrow TP = 2, FP = 0, FN = 0 \rightarrow P = 1.0, R = 1.0, F1 = 1.0$ (complete and correct answer)
\end{itemize}

\section{Discussions}
\label{sec:discussions}

\subsection{Open-Source vs. Closed-Source Models}

Analyzing Table \ref{tab:full-qa-results} from \S\ref{sec:qa_results}, if we compare experiment triplets (1,8,13), (2,9,14), (3,10,15), (4,11,16), or (5,12,17), the maximum performance gap for EM is 30\% (experiments 3 and 10) and 65\% on split 2. On split 2, this is primarily a Mistral failure due to how we prompted the model. More specifically, split 2, compared to split 1, does not include legal references. However, our experiment still includes ``\texttt{Here are the relevant documents [BLANK]}'' (where \texttt{[BLANK]} represents that no document is passed). As a result, Mistral would start and complete the missing documents instead of answering the questions without them.

This issue is not seen in OpenAI's GPT-4o mini or Gemma 3 when they answer the questions, even when the missing documents are present. When comparing GPT-4o mini and Gemma 3 on the mentioned sets of experiments, their performance is relatively close, and in some cases Gemma 3 even outperforms GPT-4o mini. The real performance gap begins with OpenAI's o4-mini reasoning model, which indeed outperforms Gemma 3 (i.e., a non-reasoning model). However, this comparison would not be fair.

\subsection{Question Difficulty}

We discuss the difficulty of the question in our dataset from the perspective of LLM performance. We consider a question difficult in this setting if no model can correctly predict it.
For this, we take scenarios (6, 11, 12) in Table \ref{tab:full-qa-results}, and intersect the wrong questions from each model to find the ones that are always wrongly answered. 

There are a total of 46 questions, with a higher value count of 7 and 4 in \textit{general rules}, \textit{Sanctions and offenses}, \textit{right of way}, and \textit{defensive driving}. Other than that, there is no observable pattern in these questions that would reflect that they are more difficult.

\subsection{Answer Bias}

Based on Table \ref{tab:picked_ref_distribution}, the models show no bias regarding answer selection. The answer-picking distribution matches the reference answer distribution. We observe that the answer "C" typically occurs less frequently in the references as well.

\section{Hallucination Analysis}
\label{sec:hallucinations-analysis}

\subsection{Citations}

\begin{figure*}[!ht]
\centering
\includegraphics[width=\textwidth,trim=0 0 0 0.7cm,clip]{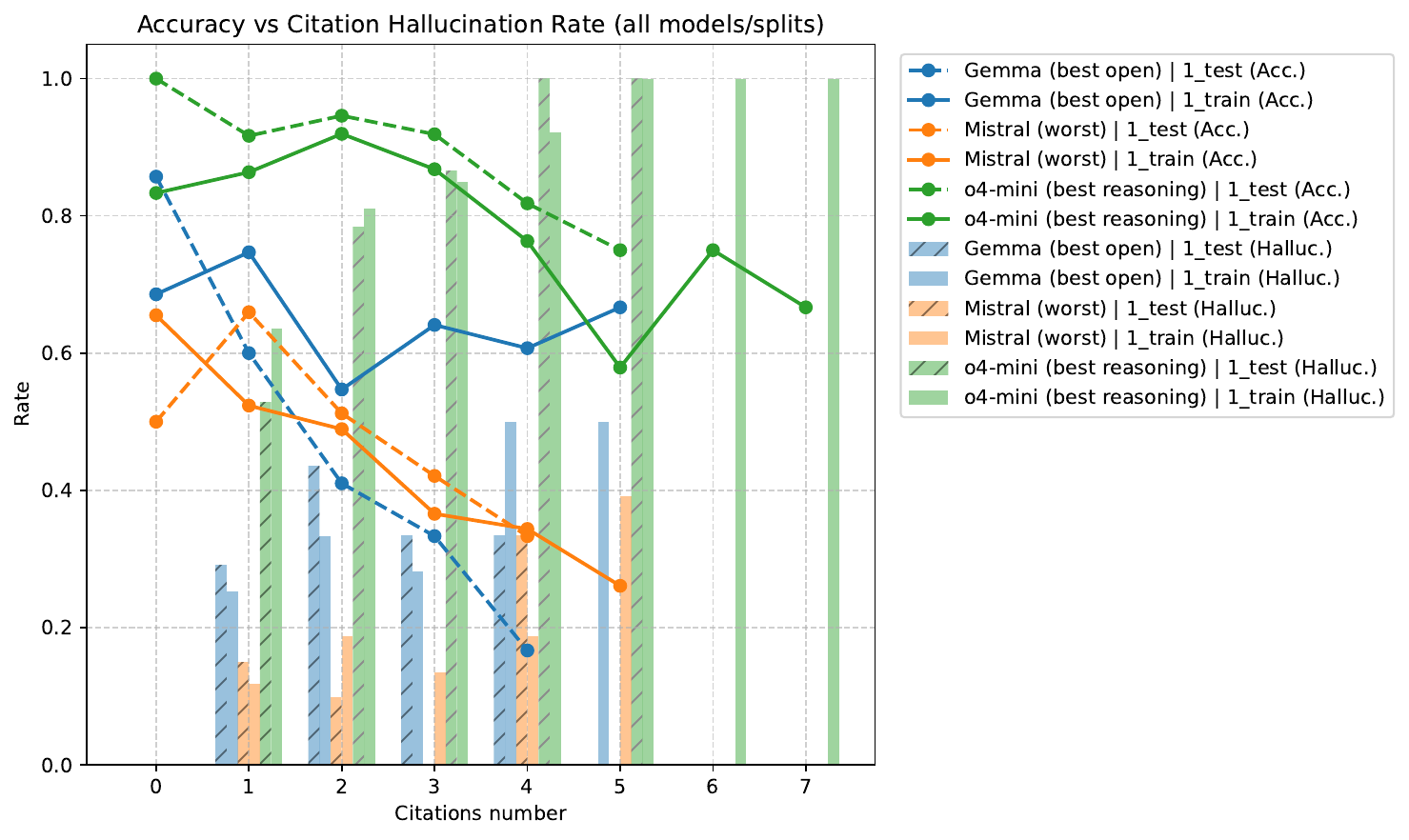}
\caption{Accuracy vs. citation hallucination rate on the number of unique citations present in output in the "RAG + BP" QA scenario.}
\label{fig:acc_vs_hallucination}
\end{figure*}

We investigate how citations count in the model output correlate with the performance. For this analysis, we chose the "RAG + BP" scenario, where each model performs well. 
We extracted the citations using GPT 5-mini\footnote{\url{https://openai.com/index/introducing-gpt-5/}}, a model different from the ones used during the experiments, because the model tends to output the citations in a different format than provided. We discuss this in the next section.
First, we counted the citations per document and aggregated the results. We computed the average accuracy per group and observed a decrease in performance as the models cite more unique documents; the findings are shown in Figure \ref{fig:acc_vs_hallucination}. After this, using the same grouping, we investigated how often these citations were not part of the documents we provided in the retrieval step. If any response included a citation not given initially, we marked it as hallucinated. We then aggregated the number of hallucinated responses per unique citation count, defining this as the \textit{citation hallucination rate}. Figure \ref{fig:acc_vs_hallucination} shows a clear trend among all models: the citation hallucination rate increases constantly with the number of unique citations. OpenAI's o4-mini has the most hallucination cases among all the models, even though it retains the best performance.

\subsection{Instruction Following}

\begin{table}
\centering
\begin{tabular}{|l|l|c|c|}
\hline
\textbf{Model} & \textbf{Split} & \textbf{Total} & \textbf{Rate (\%)} \\
\hline
Gemma 3 & 1\_test & 128 & 4.69 \\
Gemma 3 & 1\_train & 510 & 2.35 \\
Gemma 3 & 2 & 181 & 2.76 \\
Mistral  & 1\_test & 128 & 25.78 \\
Mistral  & 1\_train & 510 & 30.20 \\
Mistral  & 2 & 181 & 28.18 \\
o4-mini  & 1\_test & 128 & 2.34 \\
o4-mini  & 1\_train & 510 & 1.76 \\
o4-mini  & 2 & 181 & 0.55 \\
\hline
\end{tabular}
\caption{Instruction following in output formatting per model and split.}
\label{tab:instruction_following}
\end{table}

\begin{table*}[!ht]
\centering
\begin{tabular}{|l|c|c|c|c|c|}
\hline
\textbf{Criterion} & \textbf{Agreement (\%)} & \textbf{True (\%)} & \textbf{Majority True} (\%) & \textbf{False (\%)}  & \textbf{Majority False (\%)}
\\
\hline
C1 & 28.0 & 27.3 & 69.8 & 0.3 & 41.8 \\
C2 & 23.2 & 20.7 & 70.5 & 3.2 & 39.2 \\
C3 & 43.1 & 41.0 & 85.6 & 4.4 & 27.7 \\
C4 & 85.0 & 20.9 & 54.9 & 84.1 & 93.3 \\
C5 & 90.9 & 4.0 & 30.7 & 90.3 & 97.5 \\
\hline
\end{tabular}
\caption{LLM-as-a-Judge Agreement.}
\label{tab:llm-judge-agreement}
\end{table*}

We analyze the model's instruction-following consistency regarding the output format in Table \ref{tab:instruction_following}. We mark it as incorrect if there is anything else present in the output after the "final response:" section other than the letters A, B, and C. We observe that Gemma and o4-mini are consistent and follow instructions, while Mistral tends to have a higher incorrectness rate. Looking at outputs, for the 1\_train split, of the 154 incorrect formatting cases, 115 are due to the model providing the answers at the beginning, then the actual reasoning. In other examples, it adds even more explanations after giving the answers.

Another instruction-following evaluation criterion is that if the model cites articles using the given IDs, we consider it a success. Looking at the outputs, instead of using the given ID of the law, it would usually rephrase it and cite it in a more official manner (e.g., "According to \textit{article 4} from \textit{OUG 195/2002}" instead of "According to \textit{Regulation-4}" -- the way it is cited is the wrong way as well, there is first an OUG (Emergency Ordinance) and then it has Regulations on how to apply it, therefore citing the Regulations as the OUG is wrong in the first place).

\begin{table}
\centering
\begin{tabular}{|l|l|c|c|}
\hline
\textbf{Model} & \textbf{Split} & \textbf{Total} & \textbf{Rate (\%)} \\
\hline
Gemma 3  & 1\_test & 114 & 9.65 \\
Gemma 3 & 1\_train & 475 & 6.53 \\
Gemma 3 & 2 & 58 & 15.52 \\
Mistral  & 1\_test & 118 & 2.54 \\
Mistral  & 1\_train & 452 & 1.77 \\
Mistral  & 2 & 77 & 1.30 \\
o4-mini  & 1\_test & 125 & 11.20 \\
o4-mini  & 1\_train & 498 & 10.44 \\
o4-mini  & 2 & 80 & 8.75 \\
\hline
\end{tabular}
\caption{Wrong citation format per model and split.}
\label{tab:wrong_citation_format}
\end{table}

In Table \ref{tab:wrong_citation_format}, we analyze how many of the cases that contain citations do not contain occurrences of the correct pattern. For example, a model can cite "According to \textit{article 4} from \textit{OUG 195/2002} (\textit{Regulation-4})", so including the correct format along with the incorrect one will be counted as correct. Mistral is the model that best follows the citation format, while Gemma and o4-mini, on average, have 10\% of cases that do not follow it. Based on this, we assess that the model tends to follow the training data format rather than the regulatory data format.

\subsection{LLM-as-a-Judge for Hallucination Detection}

We use the same OpenAI o4-mini, Gemma 3 27B Instruct, and Mistral Small 3.1 Instruct to perform an LLM-as-a-Judge \cite{zheng2023judging} evaluation regarding hallucinations. For the samples, we used the results for scenarios (6, 11, 12) in Table \ref{tab:full-qa-results}, and we filtered for those that were answered incorrectly, leaving us with 789 samples to analyze.
We identify a few criteria that can be evaluated as True or False.

For this step, we use two methodologies. First, we examine the outputs to identify some common mistakes. We propose two initial criteria that stood out to us and will be denoted C4 and C5 in what follows.

A different methodology was to first use an LLM as a judge (i.e., Gemma 3) to identify (i.e., not classify) mistakes in the initial outputs, in the form of a few bullet points, yielding 3-4 mistakes per output. We then split the text based on the bullet points and newlines, embed them using mE5\textsubscript{small}, and then use KMeans \cite{lloyd1982least} to identify 15 clusters with random choice. We examine these clusters and choose three that seem more consistent.

Based on both methodologies, we get five final criteria which we aim to analyze:
\begin{itemize}
    \item C1: \textbf{Over-interpretation.} The model introduces legal rules, conditions, or consequences that are not explicitly stated in the normative texts.
    \item C2: \textbf{Negligence.} The model ignores relevant legal information, such as exceptions, special conditions, or complementary provisions.
    \item C3: \textbf{Erroneous interpretation.} The model misreads or misapplies existing legal rules without inventing new ones.
    \item C4: \textbf{Values.} The model applies a wrong legal threshold, numerical value, or standard, or uses the correct value in the wrong context.
    \item C5: \textbf{Recommendations} The model confuses the legal force of recommendations, treating them as strict obligations or vice versa.
\end{itemize}

Table \ref{tab:llm-judge-agreement} shows the agreement of the judges for each criterion. C3 is the best criterion we have, followed by C2 and C1. These have a strong majority agreement on a True basis.

C4 and C5 show strong agreement on the False basis, although C4 still shows decent agreement on True.

\begin{table}
\centering
\begin{tabular}{|l|c|c|c|}
\hline
\textbf{Criterion} & \textbf{G-O} & \textbf{M-G} & \textbf{M-O} \\
\hline
C1 & 3.85 & 6.18 & 22.09 \\
C2 & 2.97 & 12.78 & 17.62 \\
C3 & 15.54 & 19.24 & 22.86 \\
C4 & 49.99 & 59.38 & 49.46 \\
C5 & 20.97 & 42.42 & 30.81 \\
\hline
\end{tabular}
\caption{LLM-as-a-Judge Pairwise Correlations (M - Mistral, G - Gemma 3, O - o4-mini).}
\label{tab:llm-judge-corr}
\end{table}

\begin{table}
\centering
\begin{tabular}{|l|c|c|c|}
\hline
\textbf{Criterion} & \textbf{G-O} & \textbf{M-G} & \textbf{M-O} \\
\hline
C1 & 0.30 & 0.76 & 19.83 \\
C2 & 0.79 & 7.26 & 12.92 \\
C3 & 8.36 & 19.12 & 13.59 \\
C4 & 46.09 & 58.87 & 43.10 \\
C5 & 17.02 & 37.46 & 17.34 \\
\hline
\end{tabular}
\caption{LLM-as-a-Judge Pairwise Cohen's Kappa (M - Mistral, G - Gemma 3, O - o4-mini).}
\label{tab:llm-judge-kappa}
\end{table}

Looking at the Pairwise Correlations and Cohen's Kappa \cite{kohen1960coefficient} in Tables \ref{tab:llm-judge-corr} and \ref{tab:llm-judge-kappa}, there is a strong correlation on criteria C4 and C5 for all the judges, which also reflect on the False agreement previously. C3 has the next-highest correlation, while C1 and C2 have lower correlations.

\begin{figure*}[!h]
\centering
\includegraphics[width=0.9\textwidth,trim=0 0 0 1.2cm,clip]{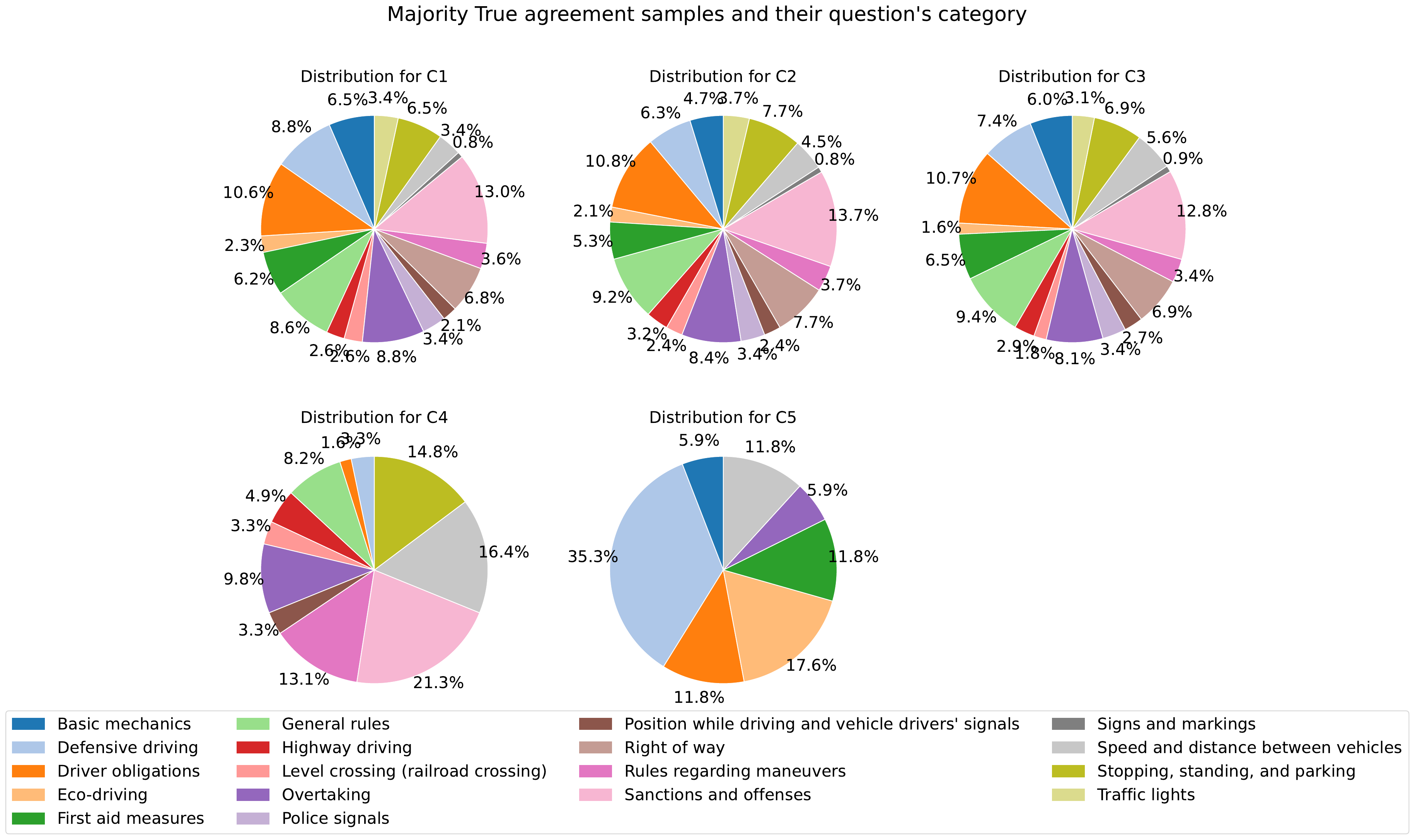}
\caption{Majority True agreement samples and their question's category.}
\label{fig:majority_true_categories}
\end{figure*}

\begin{figure*}[!h]
\centering
\includegraphics[width=0.9\textwidth,trim=0 0 0 1.2cm,clip]{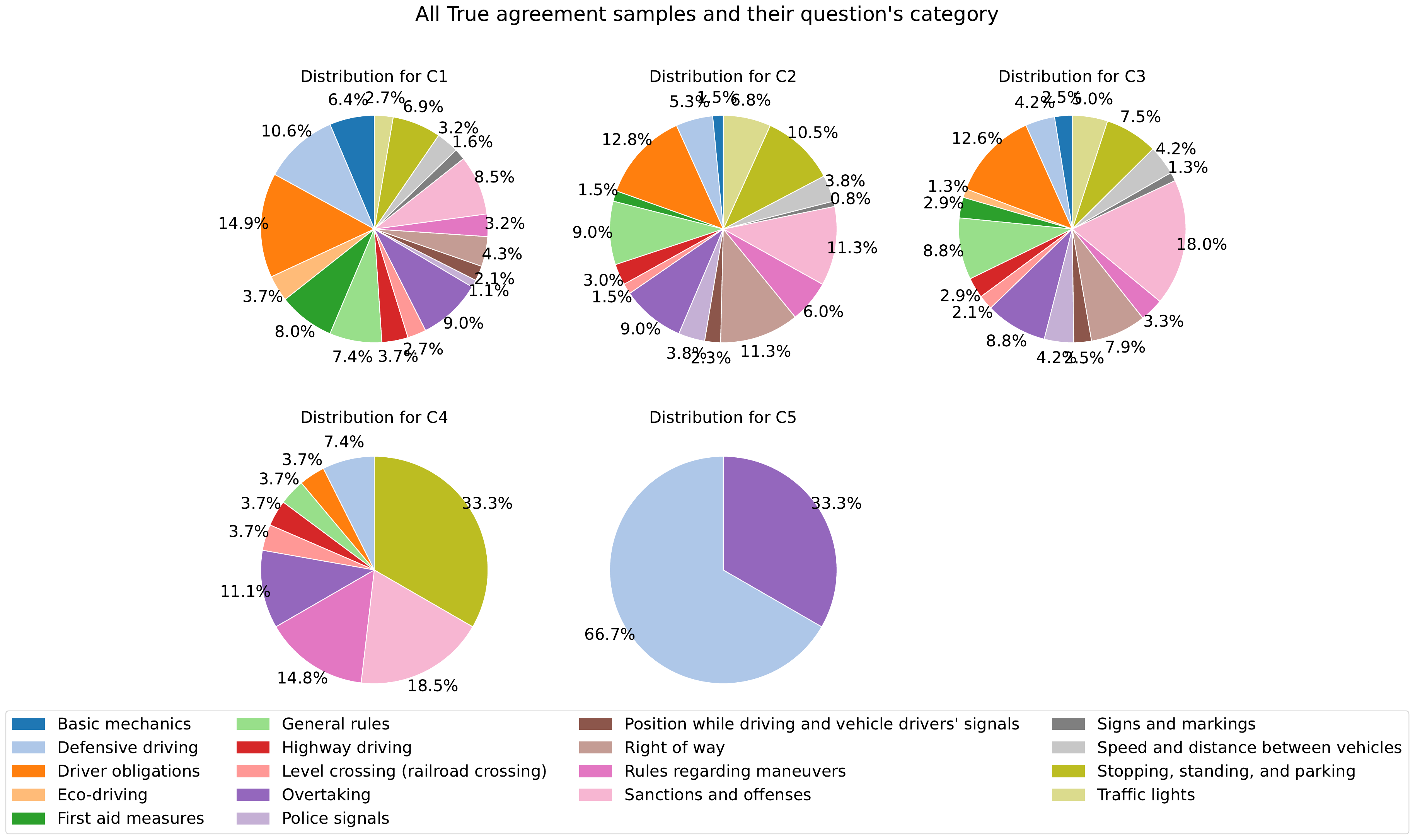}
\caption{All True agreement samples and their question's category.}
\label{fig:all_true_categories}
\end{figure*}

We observe the distribution in Figures \ref{fig:majority_true_categories} and \ref{fig:all_true_categories} for samples in which the LLMs agree on hallucinations by question category. For C1, C2, and C3, we see higher ratios for "\textit{Driver obligations}", "\textit{Sanctions and offenses}", "\textit{General rules}", and "\textit{Overtaking}". For C4, we see higher ratios for "\textit{Stopping, standing, and parking}", "\textit{Sanctions and offenses}", "\textit{Rules regarding maneuvers}", and "\textit{Overtaking}". For C5, we see the ratio is dominant for "\textit{Defensive driving}" and "\textit{Overtaking}", but this category also had the least agreement and samples. After the previous filtering, these samples were also marked as wrong during the QA evaluation.

We added a few examples in which all three models agreed on one of the criteria in Appendix~\ref{sec:appendix-qa}.

\subsection{Captioning Analysis}

We decided to inspect the generated captions based on the following criteria:
\begin{itemize}
    \item \textbf{Is the caption wrong (i.e., even partially)?} -- it contains a wrong description, added elements, or misses something very important.
    \item \textbf{Does the caption contain any legal interference?} -- the caption also contains legal obligations, such as "therefore the driver has to ...".
    \item \textbf{Is the legal interference wrong?} -- if the previous option is True, then is this interference correct or not.
\end{itemize}

We aim to analyze why VQA performance is lower when a caption is included in the prompt alongside the image.
We identify and perform analysis based on five categories:
\begin{itemize}
    \item (1) Correct caption, no legal interference, and therefore no wrong legal interference.
    \item (2) Correct caption, it has legal interference, but there is no wrong legal interference.
    \item (3) Correct caption, it has legal interference, and there is a wrong legal interference.
    \item (4) Wrong caption, no legal interference, and therefore no wrong legal interference.
    \item (5) Wrong caption, it has legal interference, and there is wrong legal interference.
\end{itemize}

\begin{table*}
\centering
\begin{tabular}{|l|l|c|c|c|}
\hline
\textbf{Category} & \textbf{Scenario} & \textbf{Total} & \textbf{Accuracy (\%)} & \textbf{Reasoning count} \\
\hline
(1)  & C & 41 & 80.4 & 1.80 \\
(1)  & C and I & 41 & 82.9 & 1.56\\
(1)  & I & 41 & 87.8 & 1.58\\
\hline
(2)  & C & 167 & 83.8 & 1.26 \\
(2)  & C and I & 167 & 83.8 & 1.29\\
(2)  & I & 167 & 86.8 & 1.70\\
\hline
(3)  & C & 33 & 63.6 & 1.48 \\
(3)  & C and I & 33 & 66.6 & 1.63\\
(3)  & I & 33 & 75.7 & 2.00 \\
\hline
(4)  & C & 30 & 46.6 & 2.30 \\
(4)  & C and I & 30 & 56.6 & 2.33 \\
(4)  & I & 30 & 73.3 & 2.26\\
\hline
(5)  & C & 116 & 50.8 & 1.91 \\
(5)  & C and I & 116 & 52.5 & 1.87 \\
(5)  & I & 116 & 64.6 & 2.13\\
\hline
\end{tabular}
\caption{Caption category accuracy and reasoning count per scenario (C - caption, I - image).}
\label{tab:caption_analysis}
\end{table*}

Based on Table \ref{tab:caption_analysis}, and on the categories (1) and (2), where the caption is correct and if it has legal interference, that is also correct, we observe that there is not a high difference between the experiments (3-7\%). In the other case, (3) where the caption is correct, but the legal interference is wrong, the performance varies by 9-12\%.
Going further, with categories (4) and (5), where the captions are descriptively incorrect and also may or may not contain wrong legal interference, the performance varies between 14-27\% when looking at the experiments with and without the added captions.
This means that by providing a caption, the model will pay less attention to the image and reason less about the question, resulting in weaker results.

We can also see in the same table that when the LLM is given the caption (e.g., comparing image and caption versus image only), the model will have fewer reasoning steps on average. This means that by providing a caption, the model will pay less attention to the image and reason less about the question, resulting in weaker results.

\clearpage
\onecolumn

\section{Additional Experimental Results}
\label{app:additional_experiments}

\begin{table*}[h!]
\centering
\small
\begin{tabular}{|l|c|c|c|c|}
\hline
\multirow{2}{*}{\centering \textbf{Method}} & \multirow{2}{*}{\centering \textbf{Metric}} & \multicolumn{2}{c|}{\textbf{Split 1}} & \textbf{Split 2} \\
\cline{3-5}
                                   & & \textbf{Train} & \textbf{Test} & \textbf{Test} \\
\hline
(1) GPT-4o mini + CoT + RAG & Precision & 77.0 & 74.7 & 85.5 \\
(2) GPT-4o mini + CoT w/o RAG  & Precision & 69.0 & 67.3 & 85.2 \\
(3) GPT-4o mini + CoT + Ideal RAG & Precision & 76.5 & 76.8 & 82.7 \\
(4) GPT-4o mini + CoT + RAG + better prompt & Precision & 80.9 & 85.8 & 87.7 \\
(5) GPT-4o mini + RAG + better prompt w/o CoT & Precision & 68.5 & 70.2 & 79.2 \\
(6) o4-mini + CoT + RAG + better prompt & Precision & \textbf{91.7} & \textbf{95.3} & \textbf{91.2} \\
(7) o4-mini + CoT + better prompt w/o RAG & Precision & 76.8 & 77.1 & 89.2 \\
\hdashline
(8) Mistral + CoT + RAG & Precision & 65.2 & 66.8 & 70.2 \\
(9) Mistral + CoT w/o RAG  & Precision & 80.3 & 78.6 & 91.0 \\
(10) Mistral + CoT + Ideal RAG & Precision & 46.9 & 51.7 & 18.5 \\
(11) Mistral + CoT + RAG + better prompt & Precision & 65.5 & 69.5 & 66.6 \\
(12) Mistral + RAG + better prompt w/o CoT & Precision & 88.6 & 86.8 & 93.4 \\
\hdashline
(13) Gemma 3 + CoT + RAG & Precision & 76.0 & 68.5 & 83.9 \\
(14) Gemma 3 + CoT w/o RAG  & Precision & 63.0 & 58.4 & 75.5  \\
(15) Gemma 3 + CoT + Ideal RAG & Precision & 75.3 & 71.5 & 84.6 \\
(16) Gemma 3 + CoT + RAG + better prompt & Precision & 80.0 & 70.8 & 87.5  \\
(17) Gemma 3 + RAG + better prompt w/o CoT & Precision & 73.3 & 67.6 & 84.3 \\
\hline
(1) GPT-4o mini + CoT + RAG & Recall & 92.8 & 92.7 & 96.9 \\
(2) GPT-4o mini + CoT w/o RAG  & Recall & 87.0 & 90.2 & 97.3 \\
(3) GPT-4o mini + CoT + Ideal RAG & Recall & 91.7 & 94.9 & 94.6 \\
(4) GPT-4o mini + CoT + RAG + better prompt & Recall & 90.7 & 92.8 & 94.6 \\
(5) GPT-4o mini + RAG + better prompt w/o CoT & Recall & 93.6 & 92.6 & 95.5 \\
(6) o4-mini + CoT + RAG + better prompt & Recall & 93.0 & 97.0 & 95.1 \\
(7) o4-mini + CoT + better prompt w/o RAG & Recall & 82.7 & 84.9 & 92.4 \\
\hdashline
(8) Mistral + CoT + RAG & Recall & \textbf{97.6} & 97.5 & \textbf{98.6} \\
(9) Mistral + CoT w/o RAG  & Recall & 96.7 & 95.7 & \textbf{98.6} \\
(10) Mistral + CoT + Ideal RAG & Recall & 94.4 & 95.0 & 84.9 \\
(11) Mistral + CoT + RAG + better prompt & Recall & 97.1 & \textbf{97.9} & 97.2 \\
(12) Mistral + RAG + better prompt w/o CoT & Recall & 97.3 & 97.2 & 97.9 \\
\hdashline
(13) Gemma 3 + CoT + RAG & Recall & 89.4 & 87.1 & 93.9 \\
(14) Gemma 3 + CoT w/o RAG  & Recall & 76.0 & 79.3 & 84.1 \\
(15) Gemma 3 + CoT + Ideal RAG & Recall & 88.8 & 88.9 & 94.6 \\
(16) Gemma 3 + CoT + RAG + better prompt & Recall & 90.5 & 87.2 & 90.7 \\
(17) Gemma 3 + RAG + better prompt w/o CoT & Recall & 88.8 & 88.3 & 92.0 \\
\hline
(1) GPT-4o mini + CoT + RAG & F1-score & 81.9 & 80.1 & 85.5 \\
(2) GPT-4o mini + CoT w/o RAG  & F1-score & 74.1 & 74.1 & 89.1 \\
(3) GPT-4o mini + CoT + Ideal RAG & F1-score & 81.2 & 83.7 & 86.4 \\
(4) GPT-4o mini + CoT + RAG + better prompt & F1-score & 83.6 & 87.4 & 89.8 \\
(5) GPT-4o mini + RAG + better prompt w/o CoT & F1-score & 76.4 & 76.6 & 84.1 \\
(6) o4-mini + CoT + RAG + better prompt & F1-score & \textbf{91.7} & \textbf{95.7} & 92.2 \\
(7) o4-mini + CoT + better prompt w/o RAG & F1-score & 78.1 & 78.7 & 89.9 \\
\hdashline
(8) Mistral + CoT + RAG & F1-score & 69.1 & 70.9 & 73.6 \\
(9) Mistral + CoT w/o RAG  & F1-score & 84.8 & 83.3 & 93.8 \\
(10) Mistral + CoT + Ideal RAG & F1-score & 50.8 & 55.2 & 23.4 \\
(11) Mistral + CoT + RAG + better prompt & F1-score & 68.2 & 72.4 & 69.6 \\
(12) Mistral + RAG + better prompt w/o CoT & F1-score & 92.0 & 90.7 & \textbf{94.5} \\
\hdashline
(13) Gemma 3 + CoT + RAG & F1-score & 79.7 & 73.8 & 86.7 \\
(14) Gemma 3 + CoT w/o RAG  & F1-score & 66.5 & 64.2 & 77.9 \\
(15) Gemma 3 + CoT + Ideal RAG & F1-score & 79.2 & 76.3 & 87.5 \\
(16) Gemma 3 + CoT + RAG + better prompt & F1-score & 82.9 & 75.1 & 88.2 \\
(17) Gemma 3 + RAG + better prompt w/o CoT & F1-score & 77.8 & 73.3 & 86.3 \\
\hline
\end{tabular}
\caption{Precision, recall, and F1-score on the IR/QA RAG pipeline.}
\label{tab:full-qa-results-v0}
\end{table*}

\begin{table}[h!]
\centering
\small
\begin{tabular}{|l|c|c|c|}
\hline
\textbf{Method} & \textbf{Metric} & \textbf{Split 3} & \textbf{Split 4} \\
\hline
(1) o4-mini + Caption + QA + CoT w/o RAG & Precision & 65.8 & 72.5 \\
(2) o4-mini + Image + QA + CoT w/o RAG & Precision & 72.4 & 76.8 \\
(3) o4-mini + Image  + Caption + QA + CoT w/o RAG & Precision & 66.3 & 74.6 \\
(4) o4-mini + Caption + QA + CoT + Ideal RAG & Precision & 70.7 & 80.3 \\
(5) o4-mini + Image + QA + CoT + Ideal RAG  & Precision & 78.6 & 79.6 \\
(6) o4-mini + Image  + Caption + QA + Ideal RAG & Precision & 73.1 & 78.9 \\
(7) o4-mini + Caption + QA + CoT + RAG & Precision & 68.5 & 76.1 \\
(8) o4-mini + Image + QA + CoT + RAG  & Precision & \textbf{76.3} & \textbf{90.1} \\
(9) o4-mini + Image  + Caption + QA + RAG & Precision & 69.8 & 78.2 \\
\hline
(1) o4-mini + Caption + QA + CoT w/o RAG & Recall & 67.4 & 73.2 \\
(2) o4-mini + Image + QA + CoT w/o RAG & Recall & 73.4 & 78.9 \\
(3) o4-mini + Image  + Caption + QA + CoT w/o RAG & Recall & 67.7 & 74.6 \\
(4) o4-mini + Caption + QA + CoT + Ideal RAG & Recall & 71.5 & 81.7 \\
(5) o4-mini + Image + QA + CoT + Ideal  RAG  & Recall & \textbf{79.4} & 80.3 \\
(6) o4-mini + Image  + Caption + QA + RAG & Recall & 74.4 & 78.9 \\
(7) o4-mini + Caption + QA + CoT + RAG & Recall & 69.6 & 76.1 \\
(8) o4-mini + Image + QA + CoT + RAG  & Recall & 76.9 & \textbf{90.1} \\
(9) o4-mini + Image  + Caption + QA + RAG & Recall & 70.3 & 78.9 \\
\hline
(1) o4-mini + Caption + QA + CoT w/o RAG & F1-score & 66.4 & 72.8 \\
(2) o4-mini + Image + QA + CoT w/o RAG & F1-score & 72.7 & 77.5 \\
(3) o4-mini + Image  + Caption + QA + CoT w/o RAG & F1-score & 66.8 & 74.6 \\
(4) o4-mini + Caption + QA + CoT + Idea RAG & F1-score & 71.0 & 80.8 \\
(5) o4-mini + Image + QA + CoT + Ideal RAG  & F1-score & \textbf{78.9} & 79.8 \\
(6) o4-mini + Image  + Caption + QA + Ideal RAG & F1-score & 73.5 & 78.9 \\
(7) o4-mini + Caption + QA + CoT + RAG & F1-score & 68.9 & 76.1 \\
(8) o4-mini + Image + QA + CoT + RAG  & F1-score & 76.5 & \textbf{90.1} \\
(9) o4-mini + Image  + Caption + QA + RAG & F1-score & 69.9 & 78.4 \\
\hline
\end{tabular}
\caption{Precision, recall, and F1-score on the VIR/VQA RAG pipeline using strategy (5) for laws and strategy (2*) for traffic signs from Table \ref{tab:full-vir-results-indicators}.}
\label{tab:full-vqa-results-v0}
\end{table}

\newpage
\section{Question and Answer Samples}
\label{sec:appendix-qa}

This section highlights key examples of observed failure modes in strategy (1) (base LLM output) and their successful mitigation by strategy (4). Each example demonstrates a specific challenge in legal question-answering for LLMs and shows how a refined approach (i.e., strategy (4)) can address it. Check the comments for every question for our observations.

\textbf{Difficult Questions:}
These examples illustrate the difficulty of strategy (1) in discerning subtle nuances or misleading elements in questions or options, which can lead to incorrect or incomplete answers. Strategy (4) consistently provides more precise and legally accurate responses.

We make these sample prompts publicly available in this repository\footnote{\GitHubURL}.

\textbf{LLM as a judge outputs:} 

\begin{itemize}[leftmargin=*,label=\textbullet]
\small

\item \textbf{Gemma 3 / 0692408a}:
\begin{itemize}[noitemsep,topsep=0pt]
    \item \textbf{Question:} What should you consider when intending to overtake both cyclists and groups of pedestrians moving on the roadway?
    \item \textbf{Options:}
    \begin{itemize}
        \item A. You may overtake only after pedestrians and cyclists signal you to do so.
        \item B (True). You must comply with the legal provisions.
        \item C. Cyclists and pedestrians may only be overtaken if the lateral distance is 3 meters.
    \end{itemize}
    \item \textbf{Model answer:} Correct answers: B, C. \\
    B is correct because overtaking cyclists and pedestrians must comply with the general legal provisions regarding overtaking (OUG-46). \\
    C is considered correct because, although not explicitly stated for pedestrians, the lateral distance of 3 meters is mandatory when overtaking cyclists according to the Traffic Regulations. \\
    A is incorrect because it is not necessary for pedestrians or cyclists to signal before being overtaken; it is the driver's responsibility to ensure safe overtaking.
    \item \textbf{Hallucination criteria:} Over-interpretation.
    \item \textbf{Gemma judge:} The answer claims that a fixed lateral distance of 3 m is mandatory when overtaking cyclists, based on the Traffic Regulations. However, the regulations do not set a fixed value; they only mention maintaining a “sufficient lateral distance” (Regulation 118 c), with additional rules depending on speed (under/over 50 km/h) regarding crossing the lane marking.
    \item \textbf{Mistral judge:} The person generalized the 3 m lateral distance rule (applicable to cyclists) and applied it to pedestrians, although Regulation 118 c) does not specify such a distance for pedestrians.
    \item \textbf{o4-mini judge:} The candidate incorrectly stated that the lateral distance when overtaking cyclists is 3 m, while Regulation 118 c) only requires a “sufficient lateral distance” without specifying a fixed number.
\end{itemize}

\item \textbf{o4-mini / cef83a0f}:
\begin{itemize}[noitemsep,topsep=0pt]
    \item \textbf{Question:} You are required to use dipped headlights during the daytime:
    \item \textbf{Options:}
    \begin{itemize}
        \item A (True). On all categories of public roads.
        \item B. Only on county, communal, and local roads.
        \item C. Only on motorways, expressways, and European national roads.
    \end{itemize}
    \item \textbf{Model answer:} The model reasoned that Article 32(2) of OUG 44/2002 establishes that the obligation to use dipped headlights unconditionally during the day exists only on motorways, expressways, and European national roads. Elsewhere, the use of dipped headlights depends on lighting conditions (Regulation 114/2002). The model selected option C as correct.
    \item \textbf{Hallucination criteria:} Negligence.
    \item \textbf{Gemma judge:} The answer misinterprets Article 32(2) of OUG 44/2002, assuming it establishes the only obligation regarding daytime dipped headlights, whereas it applies only to motor vehicles on motorways, expressways, and European national roads.
    \item \textbf{Mistral judge:} The person neglected Article 32(3) of OUG 44/2002, which requires motorcycles and mopeds to use dipped headlights on all public roads, regardless of road category.
    \item \textbf{o4-mini judge:} Article 32(3) of OUG 44/2002 was not considered; it mandates that motorcycles and mopeds must use dipped headlights on all public roads.
\end{itemize}

\item \textbf{o4-mini / ee8de379}:
\begin{itemize}[noitemsep,topsep=0pt]
    \item \textbf{Question:} What does a continuous longitudinal line separating traffic directions mean?
    \item \textbf{Options:}
    \begin{itemize}
        \item A. It guides drivers in low-visibility conditions.
        \item B (True). It prohibits crossing it.
        \item C. It prohibits overtaking on the section where it is applied.
    \end{itemize}
    \item \textbf{Model answer:} The model argued that: \\
    -- Article 19(2) of Regulation 77 states that a continuous line prohibits crossing. \\
    -- Article 103(i) of Regulation 120 prohibits overtaking when crossing the continuous line. \\
    Therefore, it concluded both B and C are correct.
    \item \textbf{Hallucination criteria:} Erroneous interpretation.
    \item \textbf{Gemma judge:} The person incorrectly equated “prohibiting crossing” with “prohibiting overtaking” entirely, without considering overtaking that does not cross the line.
    \item \textbf{Mistral judge:} The person interpreted the continuous line as an absolute ban on overtaking, ignoring Regulation 120, which allows overtaking if the line is not crossed.
    \item \textbf{o4-mini judge:} The candidate misapplied Article 120(i), interpreting the continuous line as prohibiting any overtaking, although the text only forbids overtaking when the line is crossed.
\end{itemize}

\item \textbf{Gemma 3 / 2d7567f7}:
\begin{itemize}[noitemsep,topsep=0pt]
    \item \textbf{Question:} U-turns are prohibited:
    \item \textbf{Options:}
    \begin{itemize}
        \item A (True). On pedestrian crossings.
        \item B. Less than 50 m from pedestrian crossings.
        \item C. At intersections, including roundabouts.
    \end{itemize}
    \item \textbf{Model answer:} According to the Traffic Regulations, U-turns are prohibited:
    \begin{itemize}
        \item on pedestrian crossings (letter e),
        \item in intersections, including roundabouts (letter b).
    \end{itemize}
    Less than 50 m from pedestrian crossings applies to stopping, not U-turns. \\
    Correct answers: A, C.
    \item \textbf{Hallucination criteria:} Values
    \item \textbf{Gemma judge:} The answer confused the 50 m stopping distance rule (Regulation 142 b) with U-turn prohibitions.
    \item \textbf{Mistral judge:} Regulation 142(e) prohibits stopping less than 25 m before and after pedestrian crossings, but no evidence supports a 50 m U-turn prohibition near crossings.
    \item \textbf{o4-mini judge:} The answer confused the 50 m stopping rule with U-turn rules; 50 m applies to stopping, not U-turns.
\end{itemize}

\item \textbf{Mistral / 3d4b79a3}:
\begin{itemize}[noitemsep,topsep=0pt]
    \item \textbf{Question:} What should you consider when approaching a school and noticing that students are leaving the building?
    \item \textbf{Options:}
    \begin{itemize}
        \item A. The law provides no obligations.
        \item B (True). Students may be inattentive and suddenly appear in front of the car.
        \item C. Students will only cross at pedestrian crossings.
    \end{itemize}
    \item \textbf{Model answer:} The model answered B and C, reasoning that:
    \begin{itemize}
        \item Students, especially younger ones, can be inattentive and may suddenly step onto the road.
        \item By law, pedestrians must cross at designated pedestrian crossings; however, drivers should not rely exclusively on this rule.
        \item It stated that the law does not impose specific obligations but recommended increased caution and reduced speed.
    \end{itemize}
    \item \textbf{Hallucination criteria:} Recommendations.
    \item \textbf{Gemma judge:} The answer described reducing speed and paying attention as merely “recommended”, downplaying the legal obligation under Regulation 123(h).
    \item \textbf{Mistral judge:} The person treated a legal obligation (Regulation 123 h) to reduce speed in school zones) as just a recommendation.
    \item \textbf{o4-mini judge:} Reducing speed in school zones was presented as a recommendation, but Regulation 123(g) makes it a legal obligation.
\end{itemize}

\end{itemize}

\section{Detailed Error Analysis}
\label{sec:appendix-error-analysis}

\subsection{Information Retrieval}

\begin{figure*}[htbp]
\centering
\includegraphics[width=\textwidth,trim=0 1cm 0 1cm,clip]{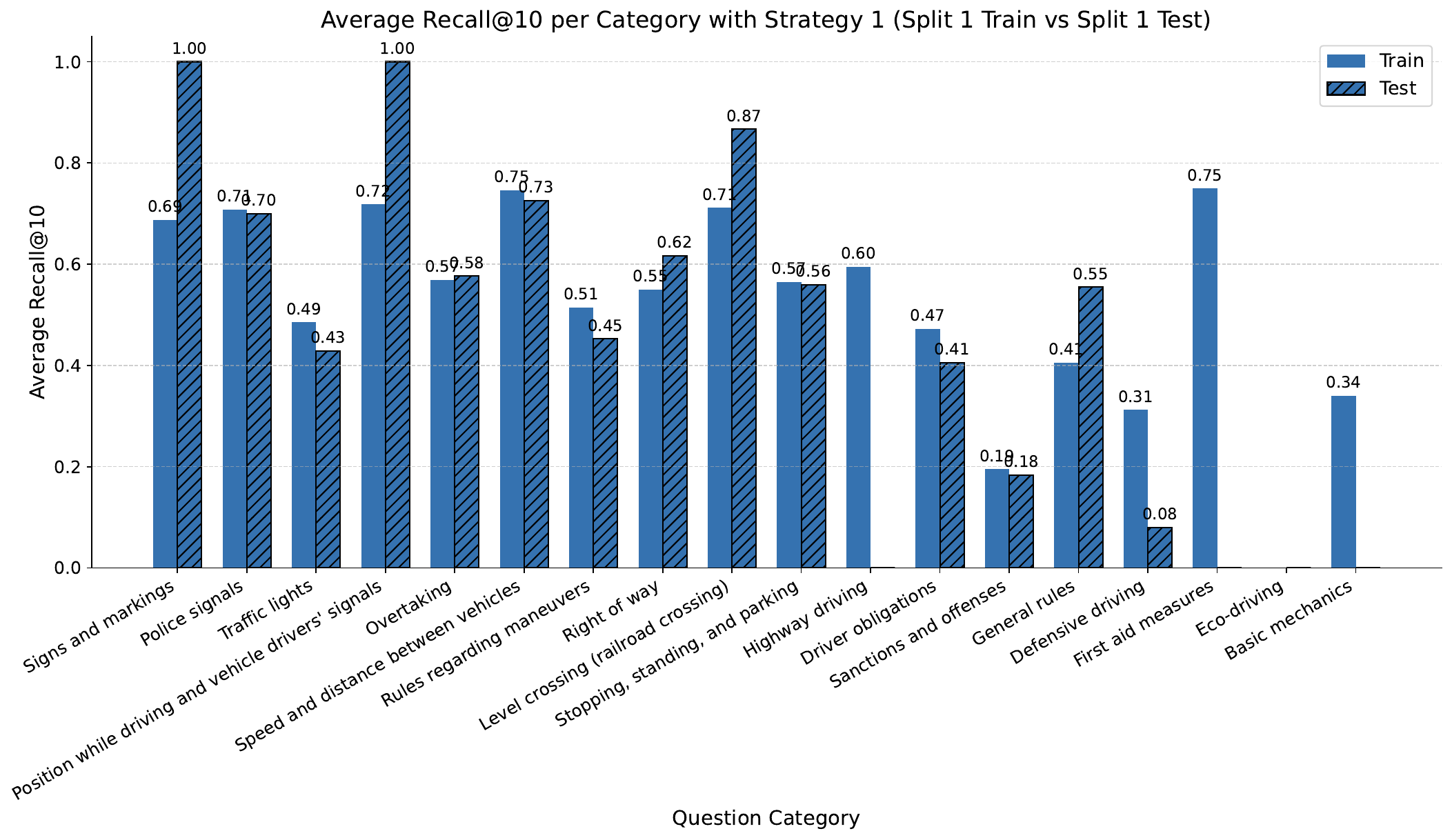}
\caption{Average Recall@10 per category for IR task with strategy (1) (split 1 train vs. split 1 test).}
\label{fig:ir_recall_per_category_train_test_st1}
\end{figure*}

\begin{figure*}[htbp]
\centering
\includegraphics[width=\textwidth,trim=0 1cm 0 1cm,clip]{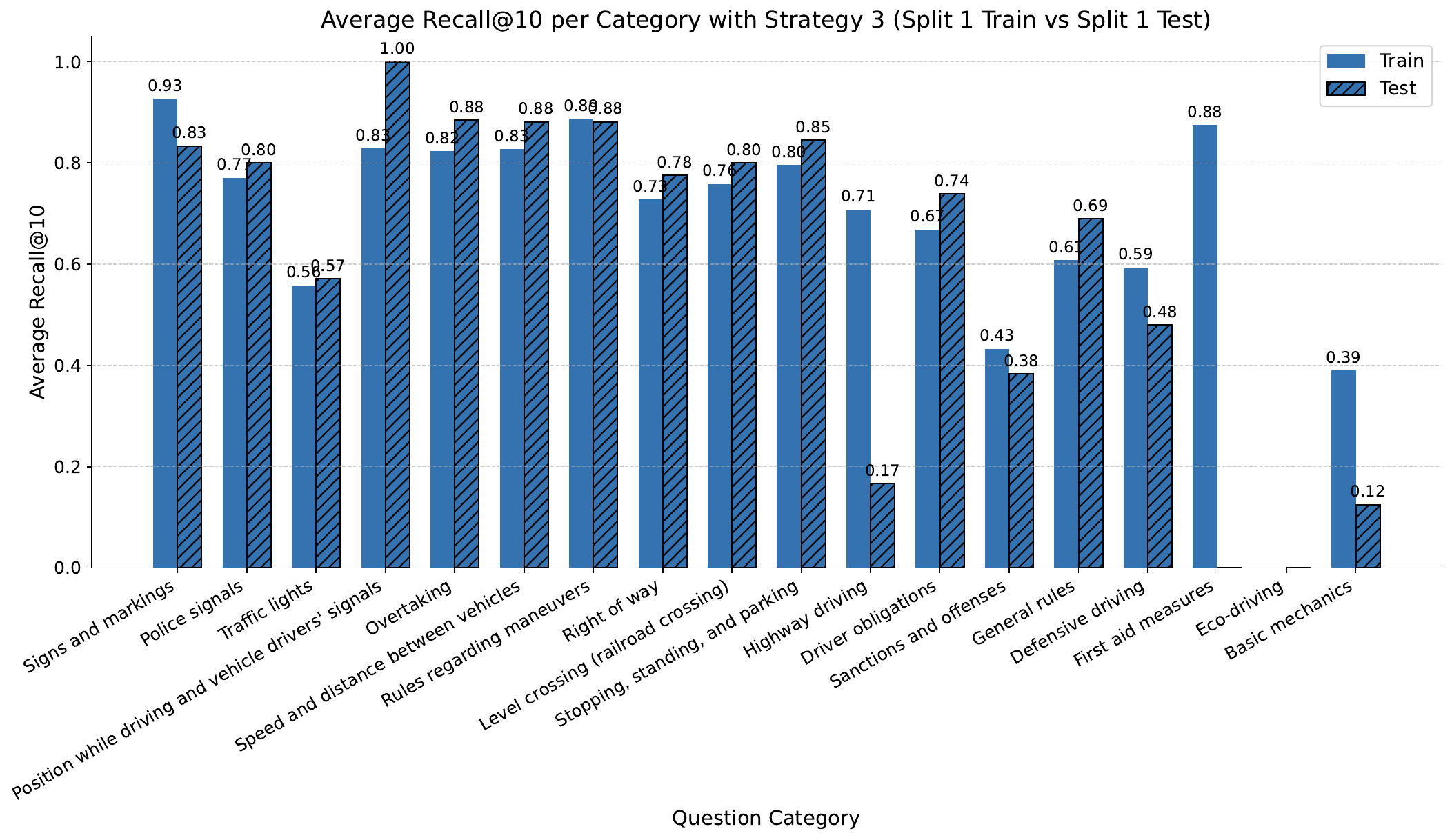}
\caption{Average Recall@10 per category for IR task with strategy (3) (split 1 train vs. split 1 test).}
\label{fig:ir_recall_per_category_train_test_st3}
\end{figure*}

Looking more in depth, if we compare strategies (1) and (3) in detail per category (see Figures \ref{fig:ir_recall_per_category_train_test_st1} and \ref{fig:ir_recall_per_category_train_test_st3}), we notice the strengths of the first strategy. In some categories, recall increased on the training split, but the score decreased on the test split, since these experiments do not involve training. This could also be due to their small size, and splitting it into so many categories further decreases the number of samples per category.

\begin{figure*}[htbp]
\centering
\includegraphics[width=\textwidth,trim=0 1cm 0 1cm,clip]{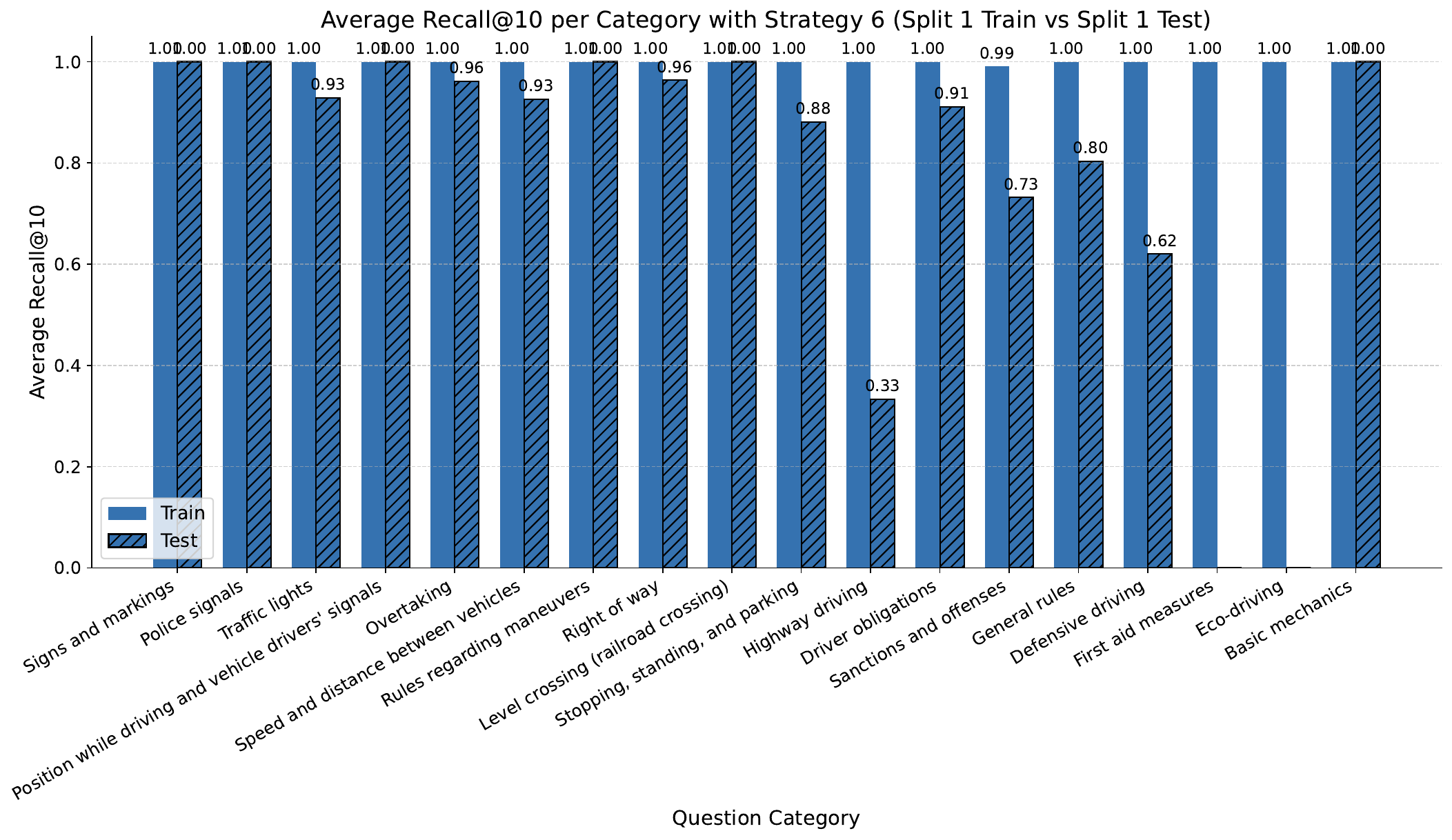}
\caption{Average Recall@10 per category for IR task with strategy (6) (split 1 train vs. split 1 test).}
\label{fig:ir_recall_per_category_train_test_st6}
\end{figure*}

\begin{figure*}[htbp]
\centering
\includegraphics[width=\textwidth,trim=0 1.3cm 0 1.3cm,clip]{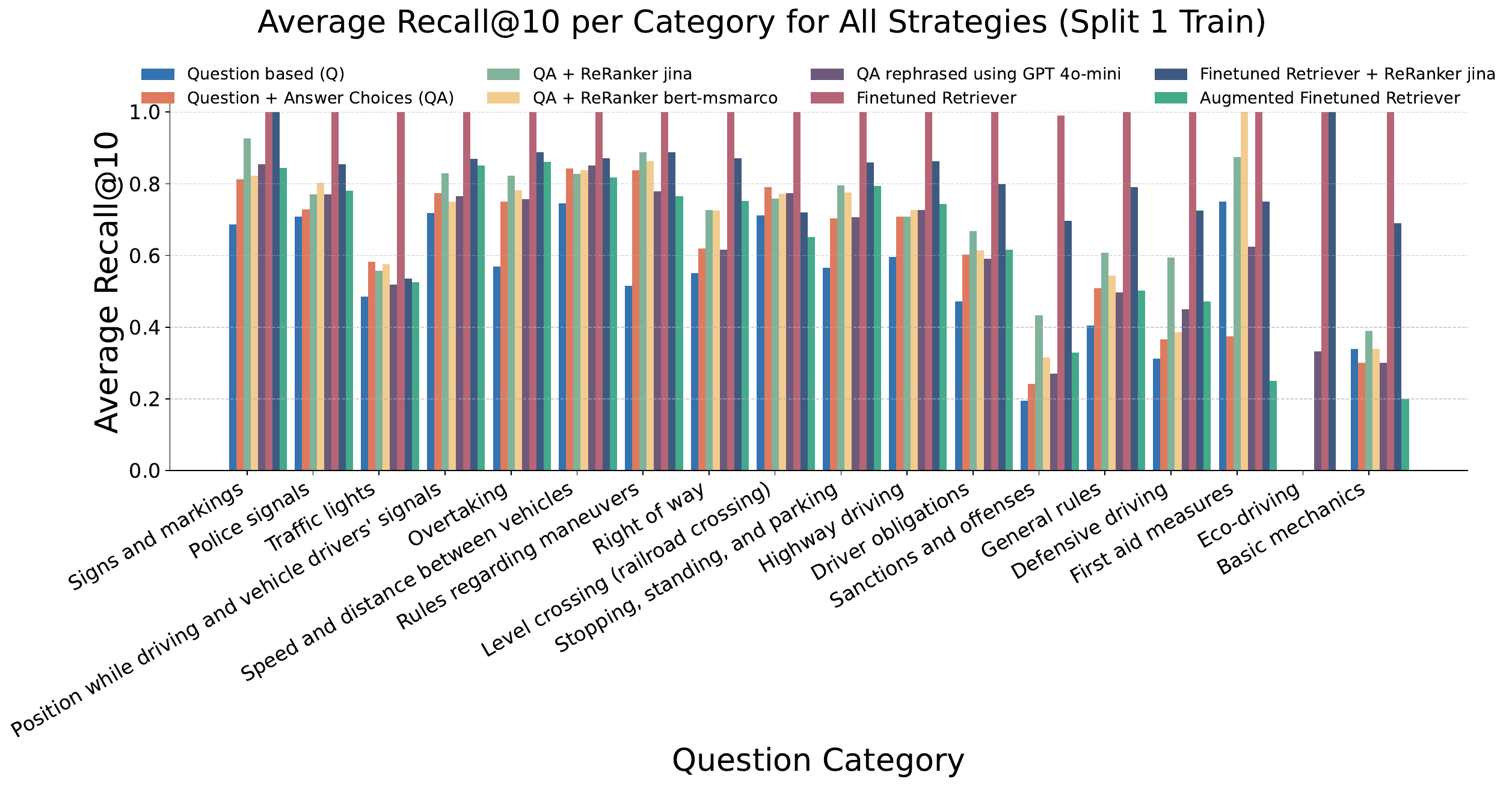}
\caption{Average Recall@10 per category for IR task with all strategies on split 1 train.}
\label{fig:ir_recall_per_category_all_strategies}
\end{figure*}

\begin{figure*}[htbp]
\centering
\includegraphics[width=\textwidth,trim=0 1.3cm 0 1.3cm,clip]{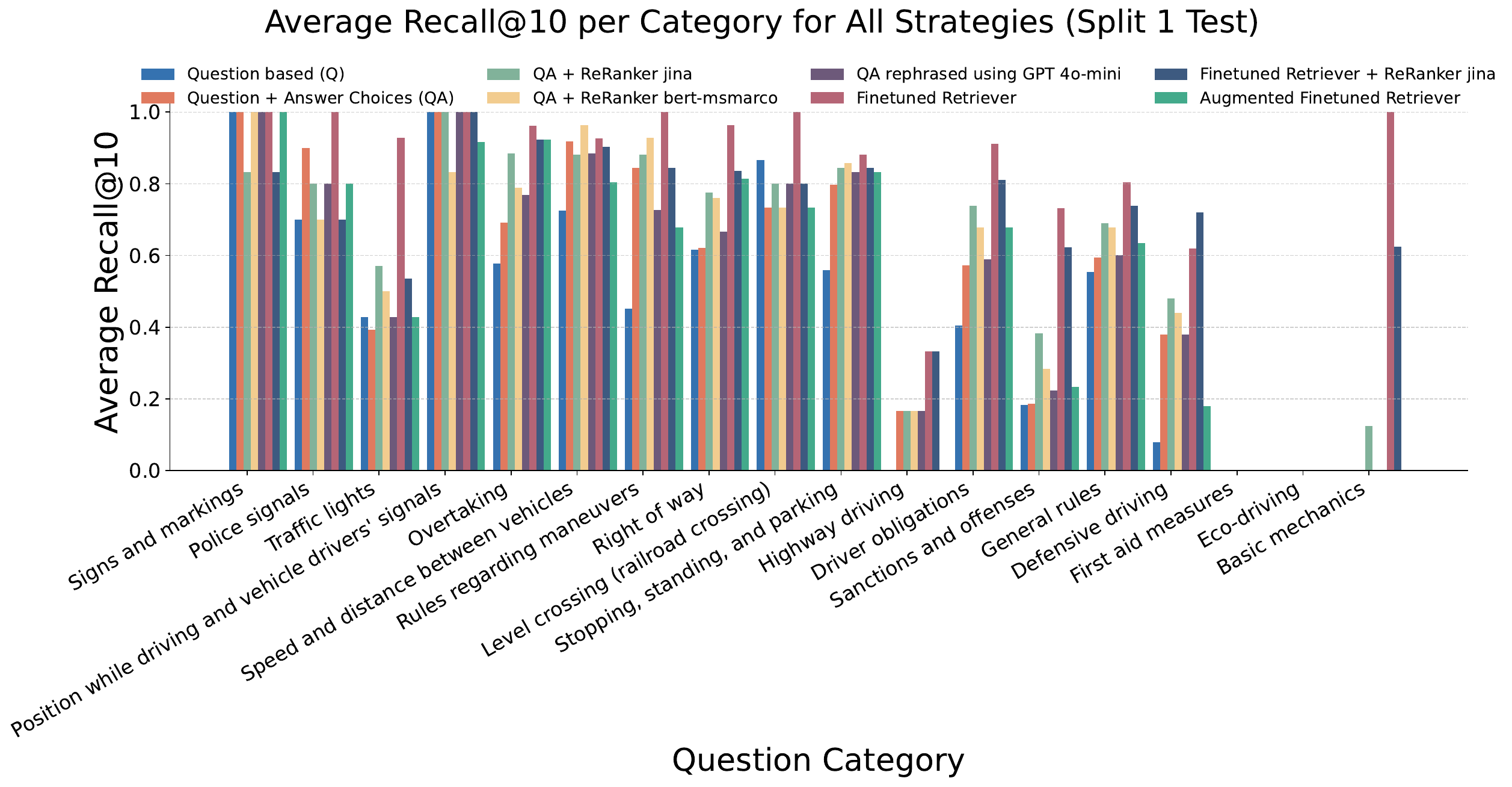}
\caption{Average Recall@10 per category for IR task with all strategies on split 1 test.}
\label{fig:ir_recall_per_category_all_strategies_test}
\end{figure*}

On (6) from Figure \ref{fig:ir_recall_per_category_train_test_st6}, where we employ fine-tuning, we see an overall improvement in all categories. 
We observe that categories such as \textit{Highway Driving} and \textit{Defensive Driving} underperform on the test set because they have a smaller share of the dataset, making it harder to learn relevant features. Similarly, the test set score is also lower in \textit{Sanctions and Offenses} and \textit{General Rules}. A fully comparative analysis between all the strategies can be seen in Figures \ref{fig:ir_recall_per_category_all_strategies} and \ref{fig:ir_recall_per_category_all_strategies_test}.

\clearpage 

\subsection{Question Answering}

\begin{figure*}[htbp]
\begin{subfigure}{\columnwidth}
\centering
\includegraphics[width=\textwidth,trim=0 1.3cm 0 1.3cm,clip]{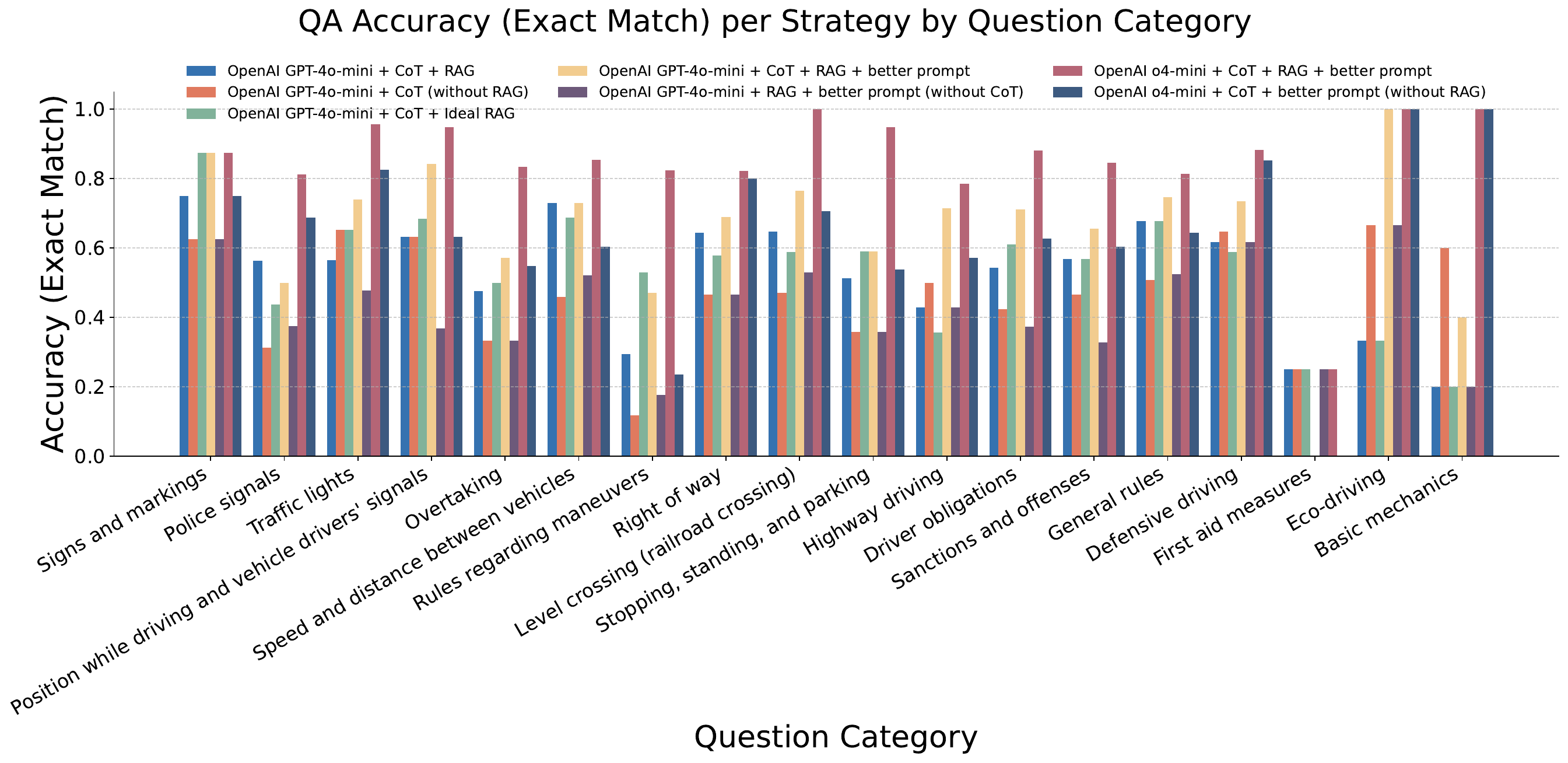}
\caption{Split 1 train.}
\label{fig:qa_accuracy_exact_match_per_category_strategies_1_train}
\end{subfigure}
\hfill
\begin{subfigure}{\columnwidth}
\centering
\includegraphics[width=\textwidth,trim=0 1.3cm 0 1.3cm,clip]{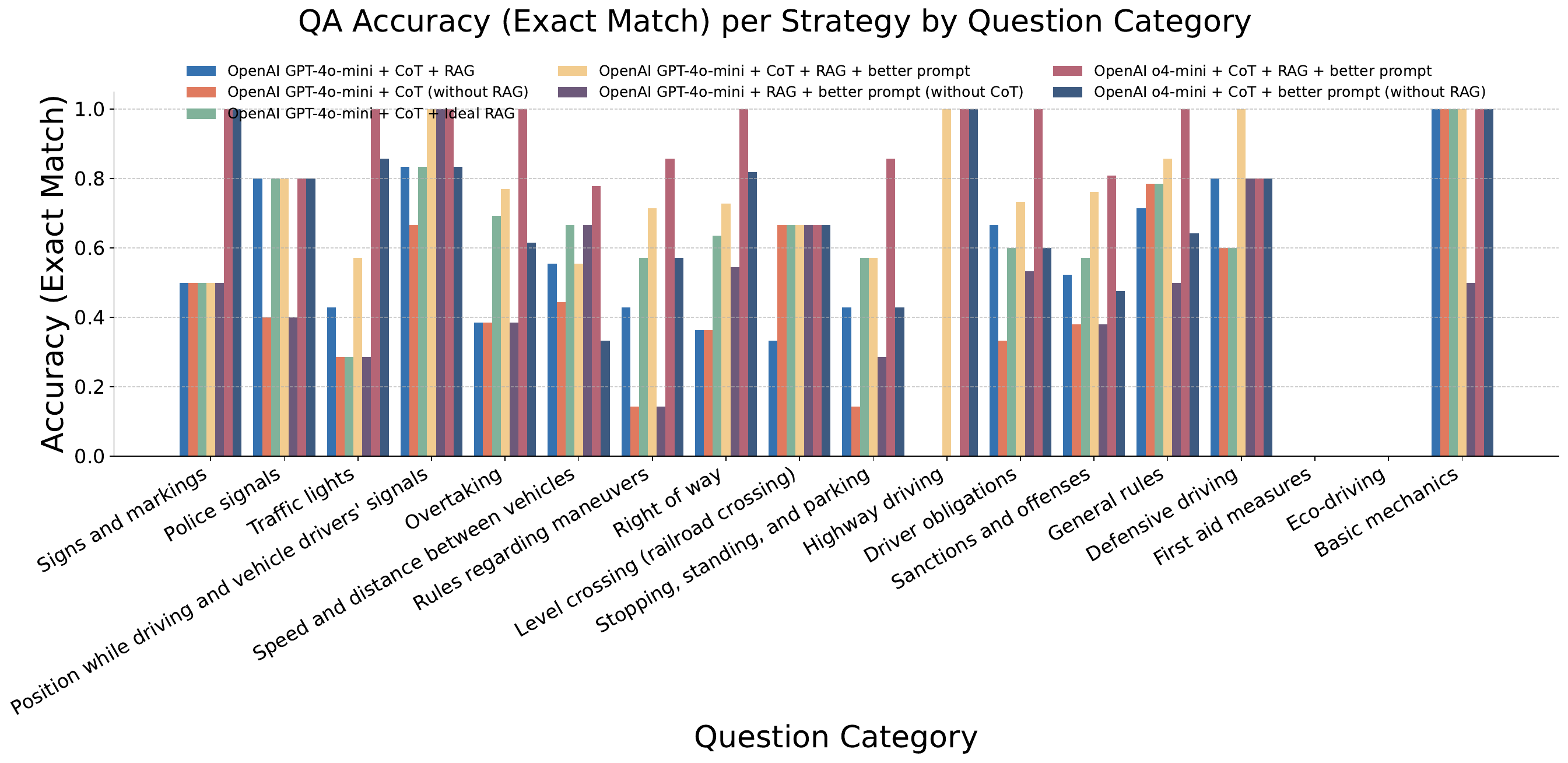}
\caption{Split 1 test.}
\label{fig:qa_accuracy_exact_match_per_category_strategies_1_test}
\end{subfigure}
\hfill
\begin{subfigure}{\columnwidth}
\centering
\includegraphics[width=\textwidth,trim=0 1.3cm 0 1.3cm,clip]{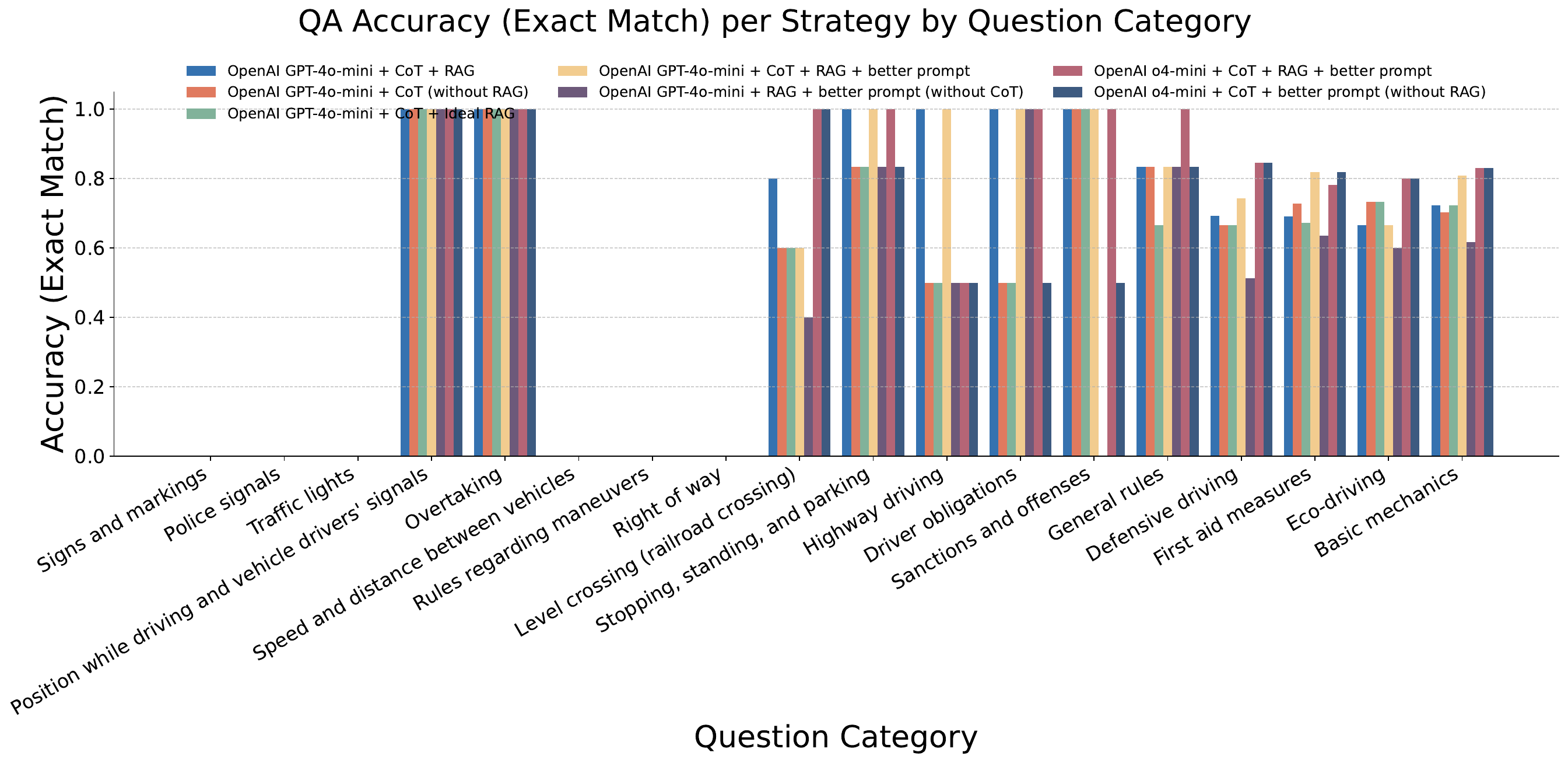}
\caption{Split 2.}
\label{fig:qa_accuracy_exact_match_per_category_strategies_2}
\end{subfigure}
\caption{QA exact match on o4-mini model for QA per strategy and category.}
\label{fig:qa_accuracy_exact_match_per_category}
\end{figure*}

Looking at performance per category in Figure \ref{fig:qa_accuracy_exact_match_per_category}, in most categories the RAG ablation has a significant impact or, in some cases, similar performance. However, in "\textit{eco driving}" and "\textit{basic mechanics}", categories that do not require legal grounding but rather general knowledge, the RAG is downgrading performance (due to context bloat). But there are not many entries in these categories in the split 1 train and test sets. In split 2, with more entries, we see more similar results across the strategies. We can also see that the ideal RAG has little impact compared to our proposed solution, suggesting that better retrieval does not lead to better QA in our case.
The reasoning models perform better than the normal ones, and the RAG continues to have an impact across most categories.

\begin{figure*}[htbp]
\begin{subfigure}{\columnwidth}
\centering
\includegraphics[width=\textwidth,trim=0 1.3cm 0 1.3cm,clip]{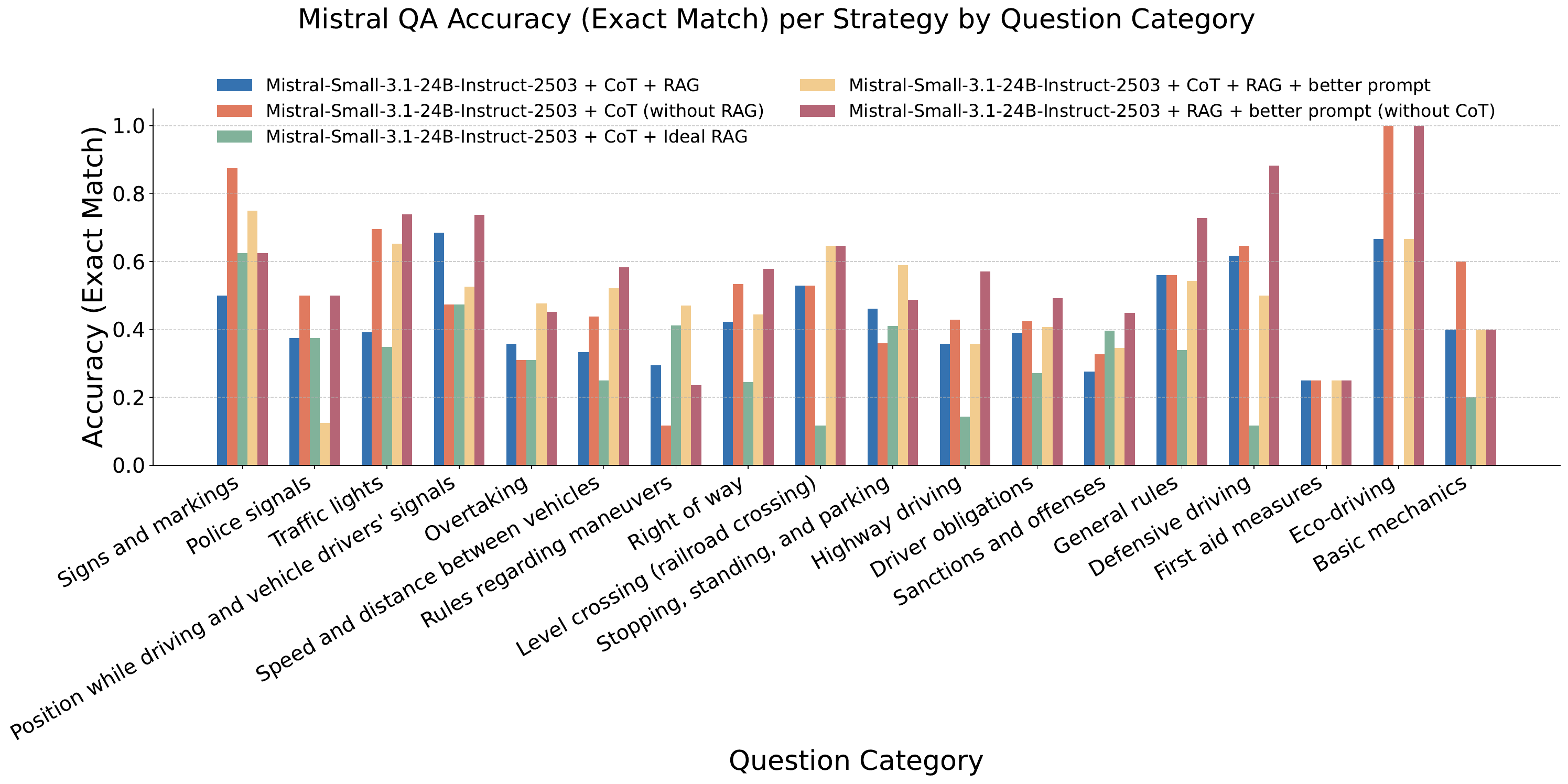}
\caption{Split 1 train.}
\label{fig:mistral_qa_accuracy_exact_match_per_category_strategies_1_train}
\end{subfigure}
\hfill
\begin{subfigure}{\columnwidth}
\centering
\includegraphics[width=\textwidth,trim=0 1.3cm 0 1.3cm,clip]{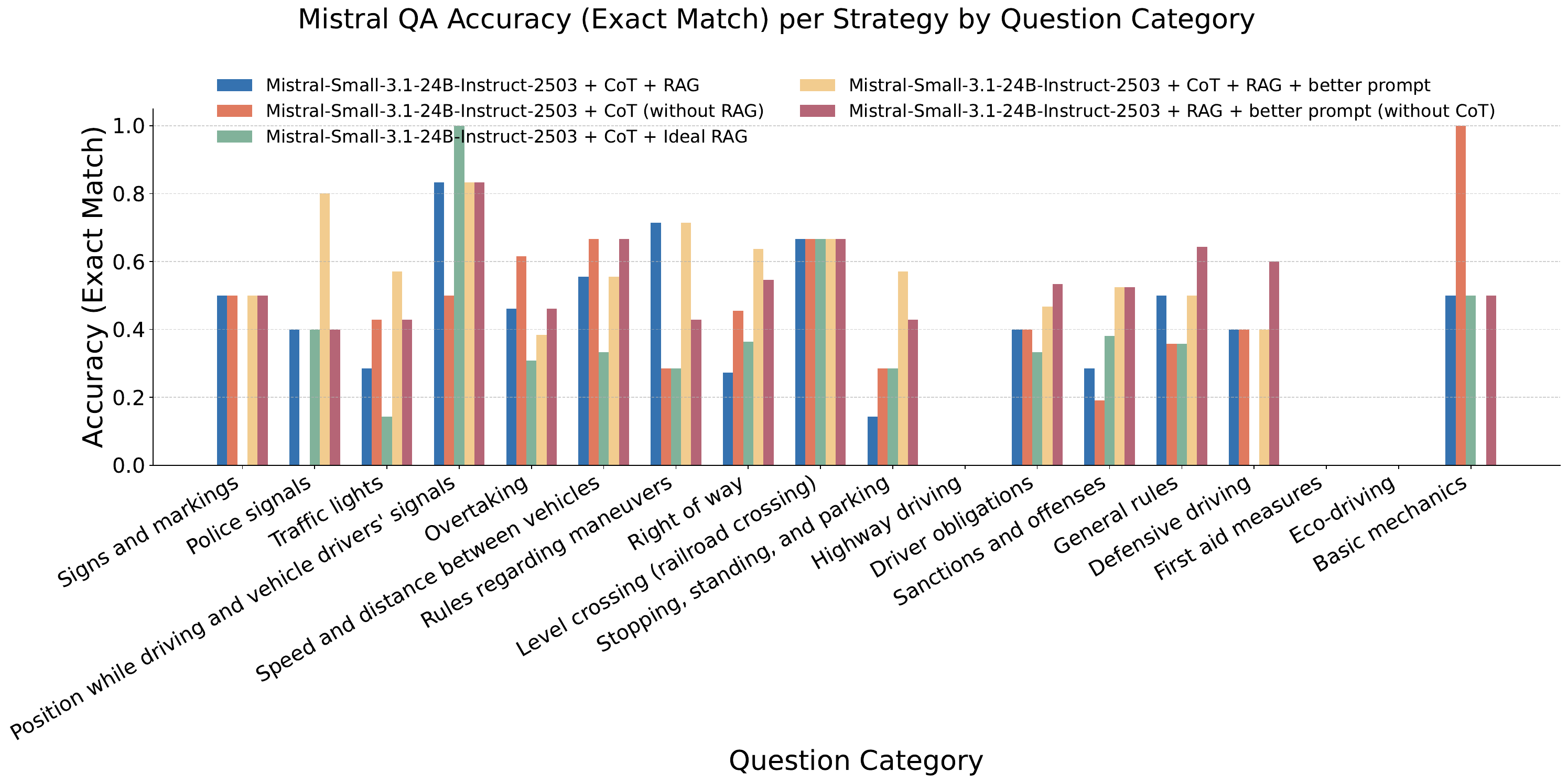}
\caption{Split 1 test.}
\label{fig:mistral_qa_accuracy_exact_match_per_category_strategies_1_test}
\end{subfigure}
\hfill
\begin{subfigure}{\columnwidth}
\centering
\includegraphics[width=\textwidth,trim=0 1.3cm 0 1.3cm,clip]{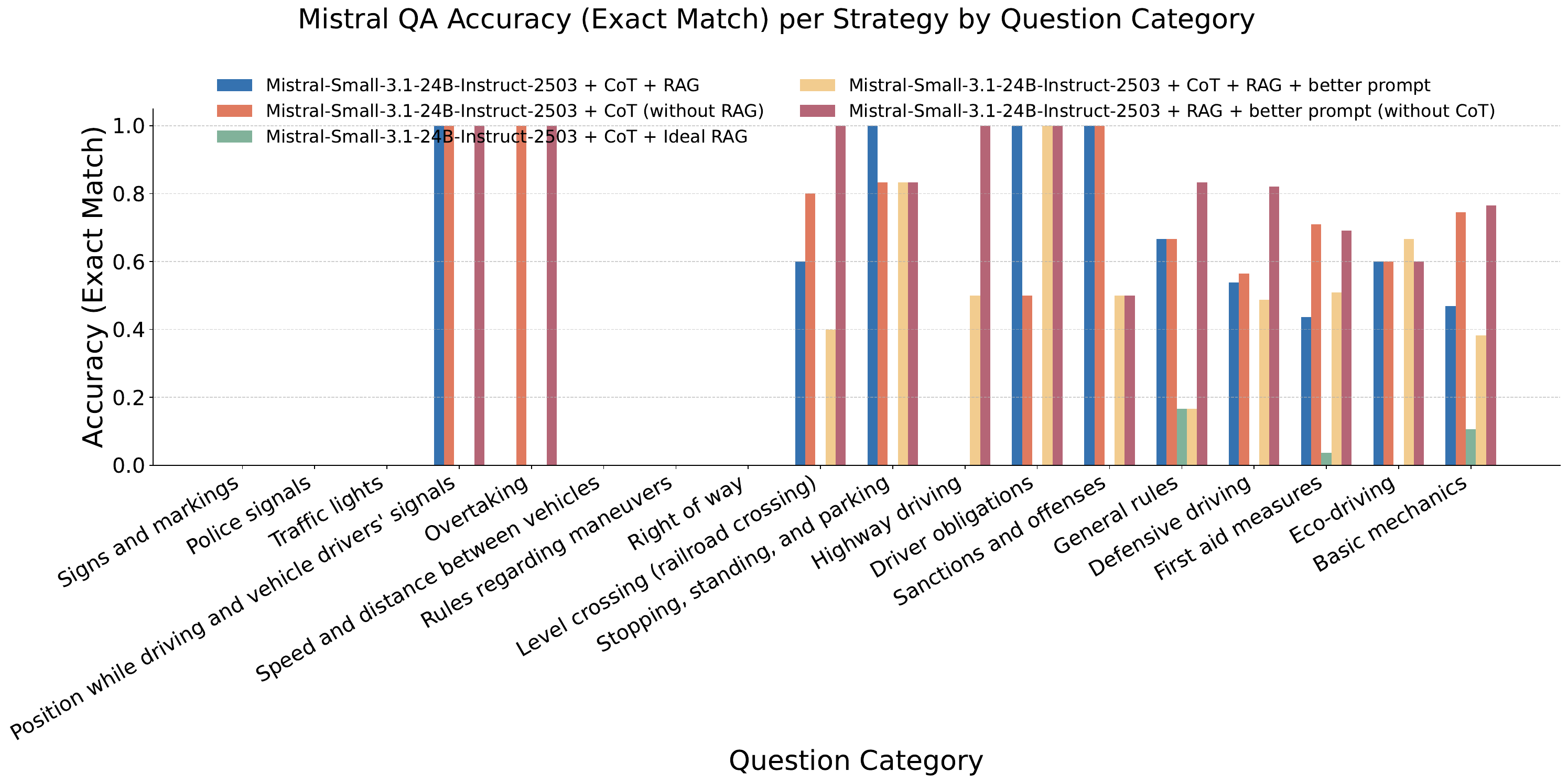}
\caption{Split 2.}
\label{fig:mistral_qa_accuracy_exact_match_per_category_strategies_2}
\end{subfigure}
\caption{QA exact match on Mistral model for QA per strategy and category.}
\label{fig:mistral_qa_accuracy_exact_match_per_category}
\end{figure*}

For the Mistral model, Figure \ref{fig:mistral_qa_accuracy_exact_match_per_category} shows that not employing RAG helps in most categories. This is likely a flaw in the Mistral model, as its outputs also tend to return in the wrong format or fail to follow instructions, instead continuing to write article paragraphs rather than responding to the query.

\begin{figure*}[htbp]
\begin{subfigure}{\columnwidth}
\centering
\includegraphics[width=\textwidth,trim=0 1.3cm 0 1.3cm,clip]{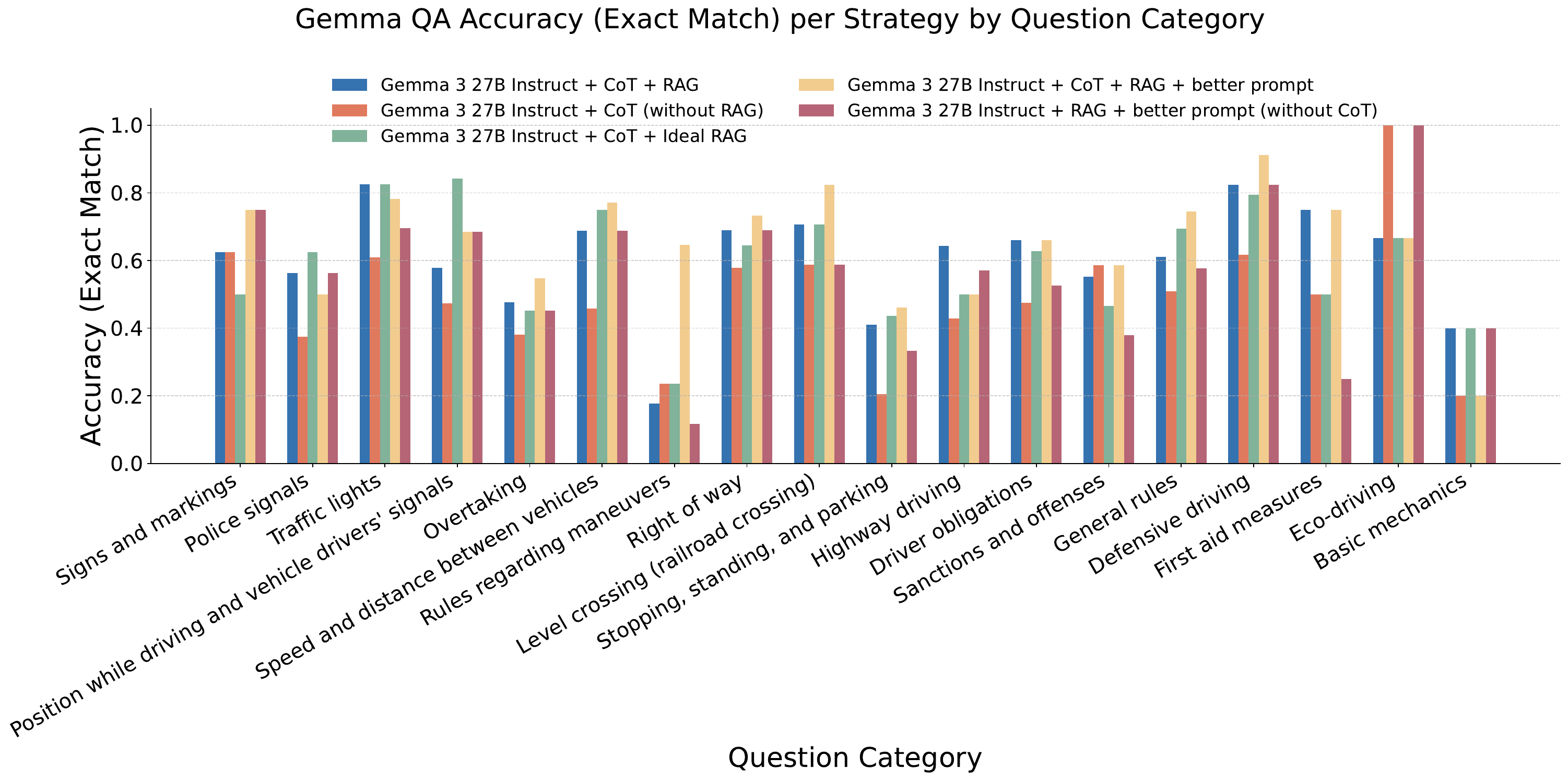}
\caption{Split 1 Train.}
\label{fig:gemma_qa_accuracy_exact_match_per_category_strategies_1_trai}
\end{subfigure}
\hfill
\begin{subfigure}{\columnwidth}
\centering
\includegraphics[width=\textwidth,trim=0 1.3cm 0 1.3cm,clip]{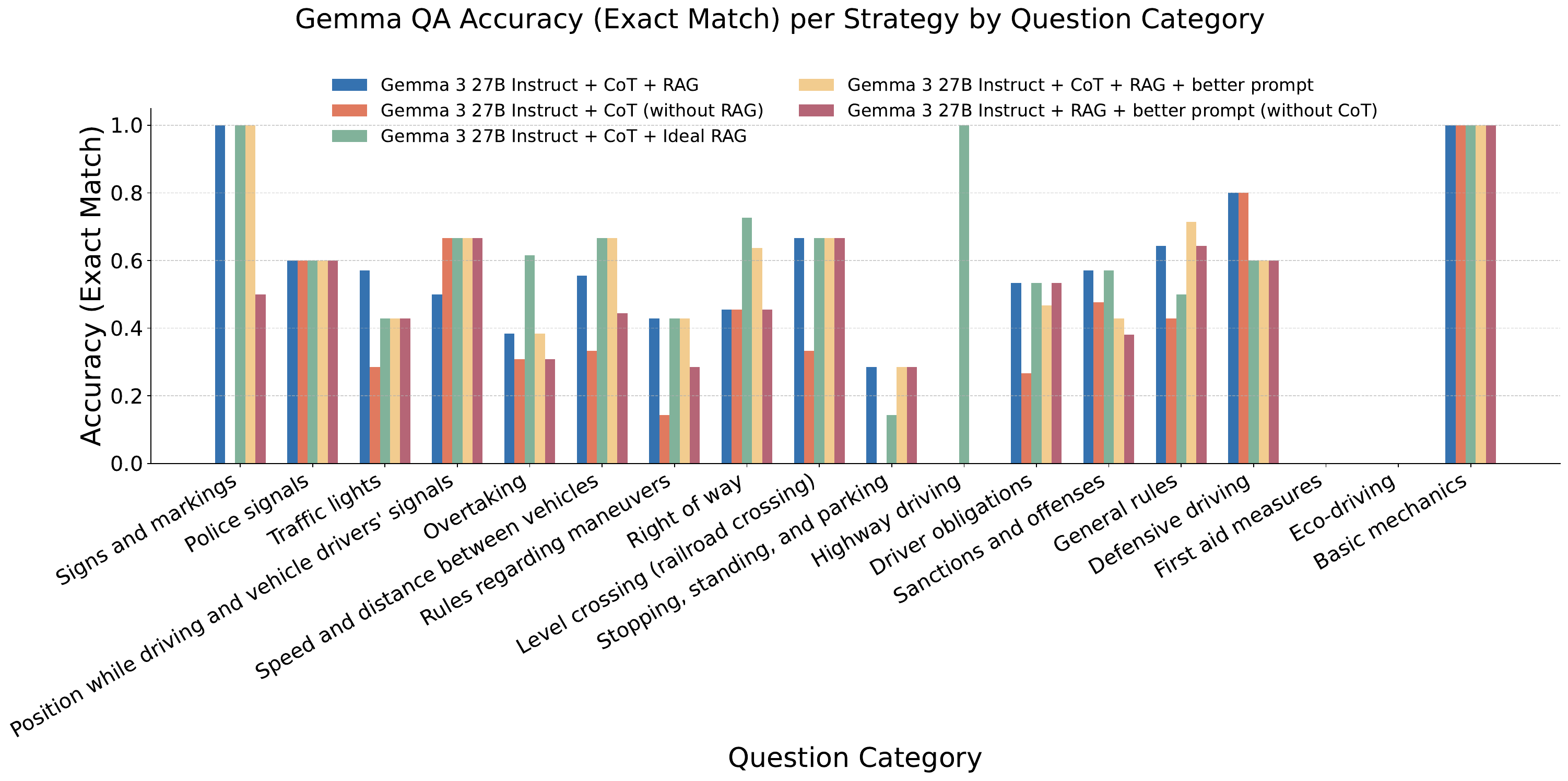}
\caption{Split 1 Test.}
\label{fig:gemma_qa_accuracy_exact_match_per_category_strategies_1_test}
\end{subfigure}
\hfill
\begin{subfigure}{\columnwidth}
\centering
\includegraphics[width=\textwidth,trim=0 1.3cm 0 1.3cm,clip]{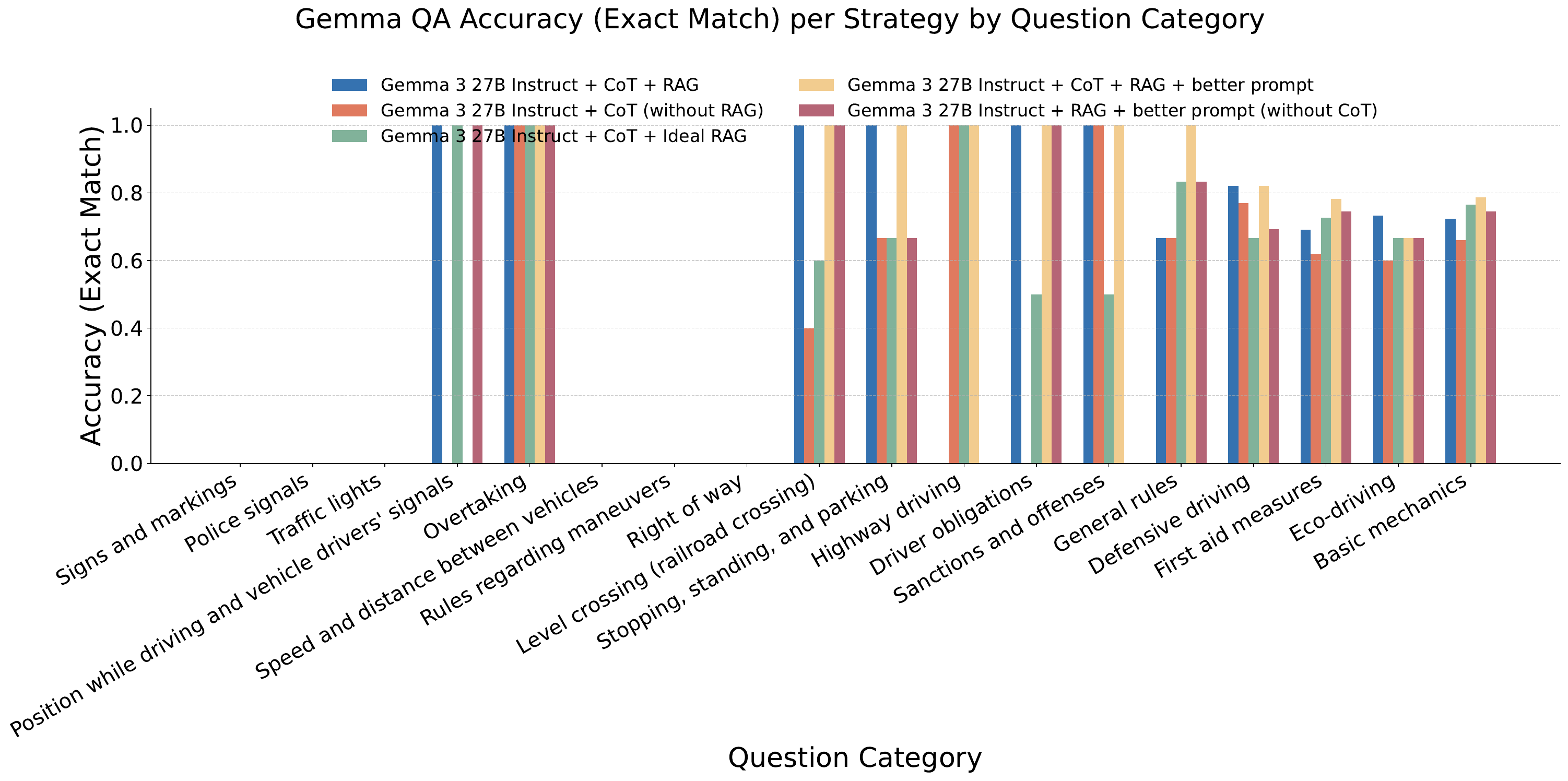}
\caption{Split 2.}
\label{fig:gemma_qa_accuracy_exact_match_per_category_strategies_2}
\end{subfigure}
\caption{QA exact match on Gemma 3 for QA per strategy and category.}
\label{fig:gemma_qa_accuracy_exact_match_per_category}
\end{figure*}

The Gemma 3 model (Figure \ref{fig:gemma_qa_accuracy_exact_match_per_category}) has similar performance to GPT-4o mini.

\begin{figure*}[htbp]
\begin{subfigure}{\columnwidth}
\centering
\includegraphics[width=\textwidth,trim=0 1.1cm 0 2cm,clip]{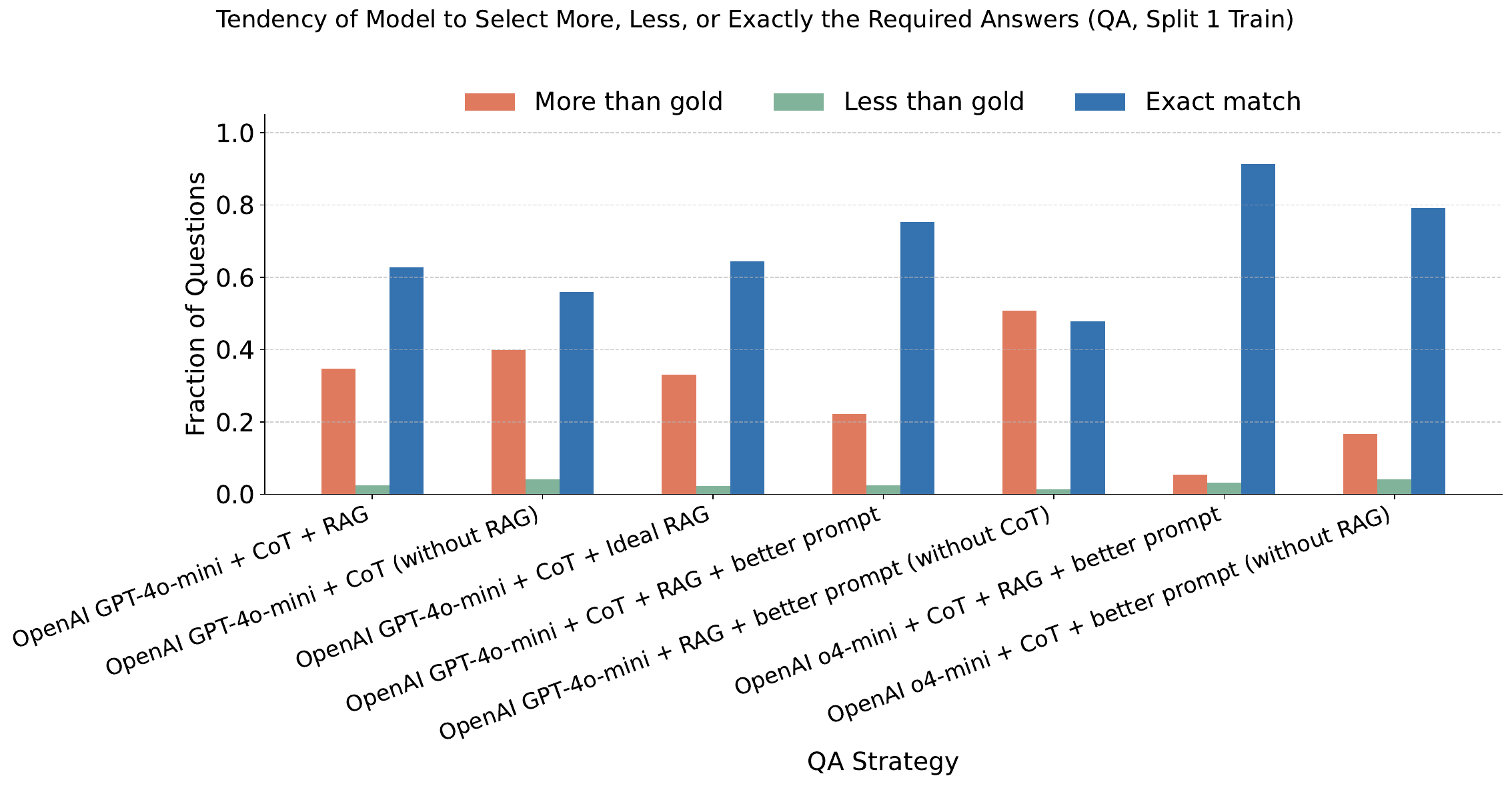}
\caption{Split 1 train.}
\label{fig:qa_tendency_over_under_exact_1_train}
\end{subfigure}
\hfill
\begin{subfigure}{\columnwidth}
\centering
\includegraphics[width=\textwidth,trim=0 1.1cm 0 2cm,clip]{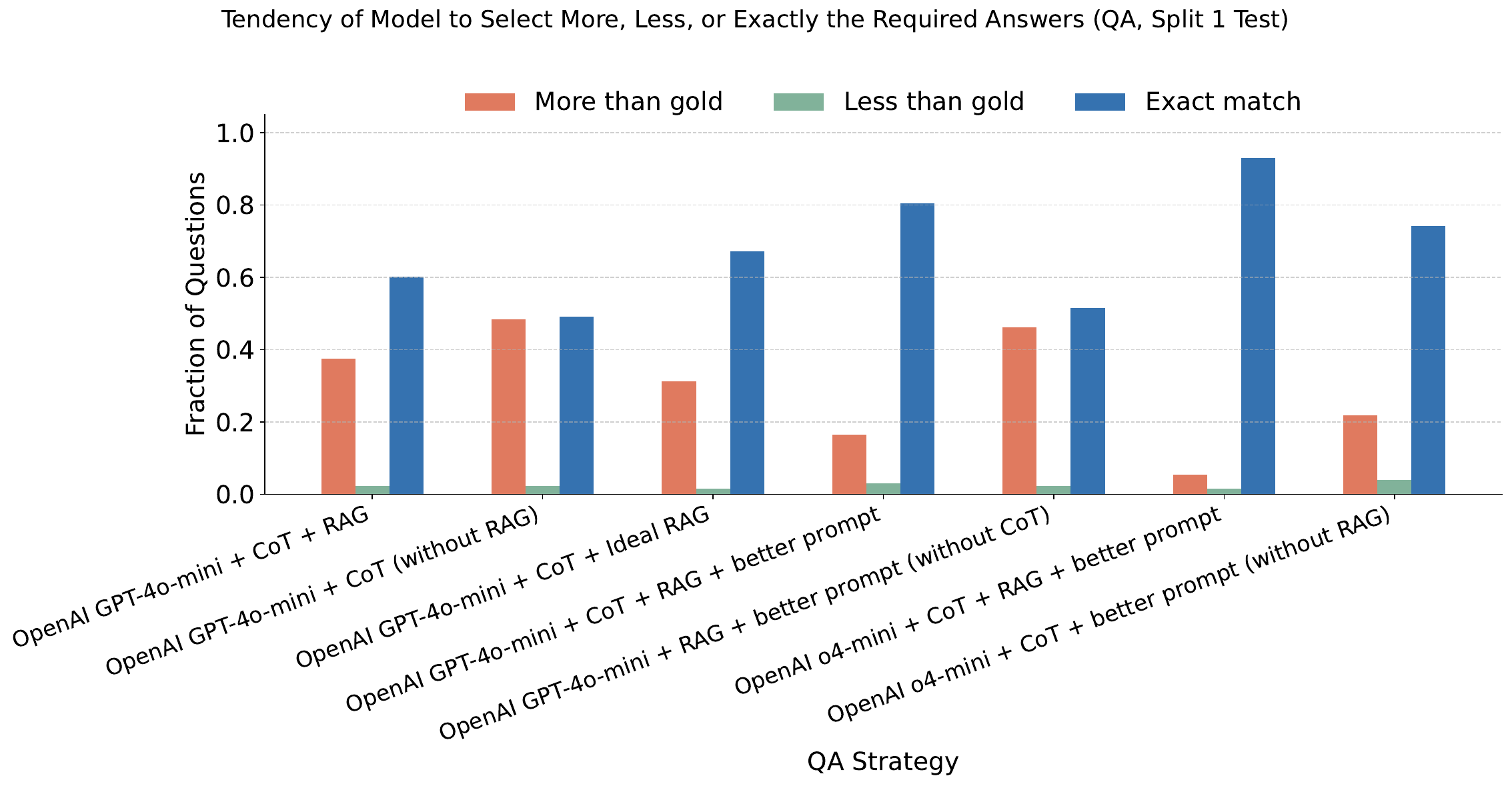}
\caption{Split 1 test.}
\label{fig:qa_tendency_over_under_exact_1_test}
\end{subfigure}
\hfill
\begin{subfigure}{\columnwidth}
\centering
\includegraphics[width=\textwidth,trim=0 1.1cm 0 2cm,clip]{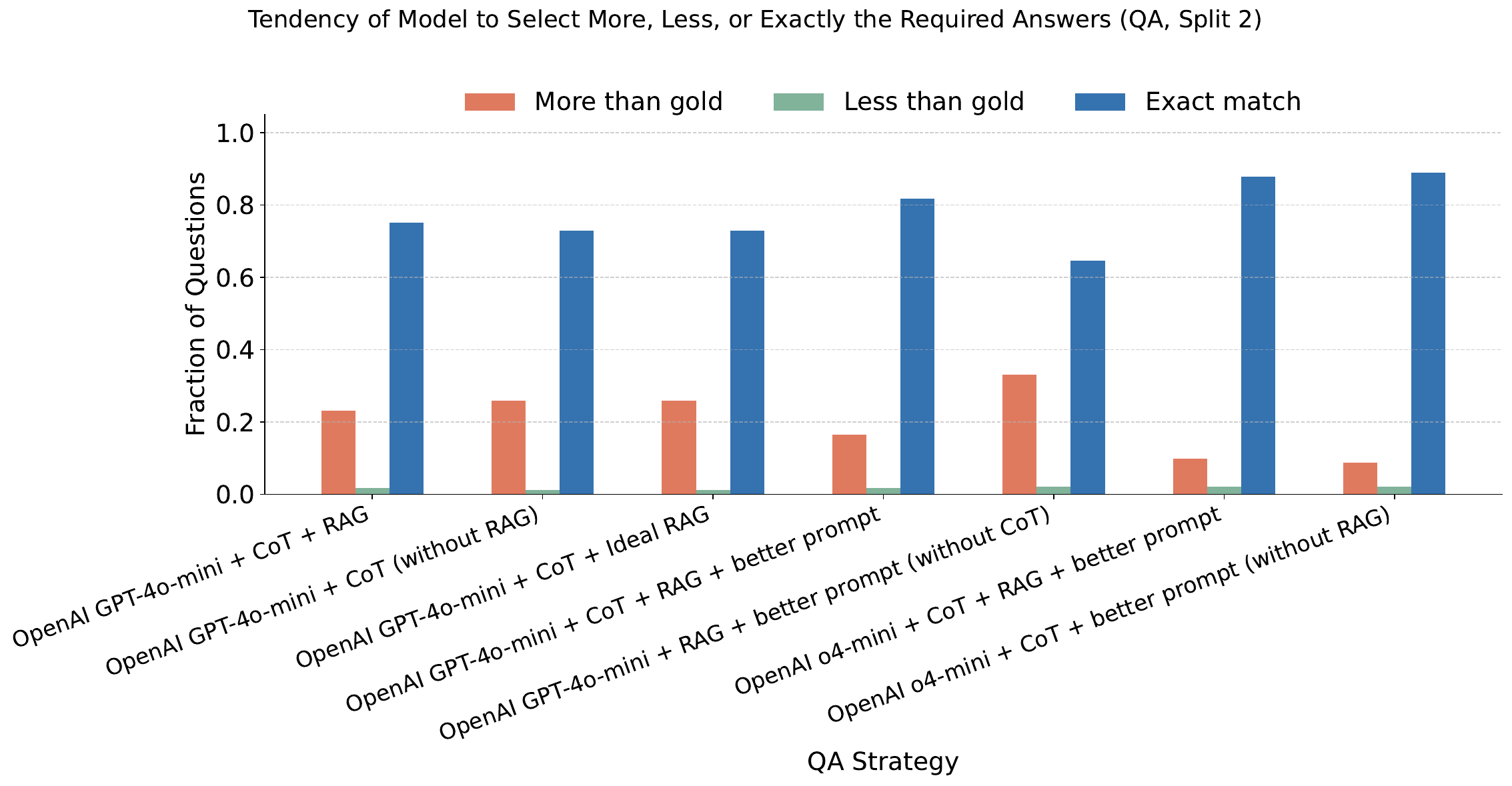}
\caption{Split 2.}
\label{fig:qa_tendency_over_under_exact_2}
\end{subfigure}
\caption{Tendency of o4-mini to select more, less, or exactly the required answers on the QA task.}
\label{fig:qa_tendency_over_under_exact}
\end{figure*}

If we look at the number of selected answers, in Figure \ref{fig:qa_tendency_over_under_exact}, most of the time the models select the right number of answers or more than the number of answers, but in very few cases, fewer than the number of answers. Similar to the recall vs precision comparison, these experiments validate that the model mostly chooses the right answers but sometimes picks more than it should when it makes errors.

\begin{figure*}[htbp]
\begin{subfigure}{\columnwidth}
\centering
\includegraphics[width=0.9\textwidth,trim=0 1.1cm 0 2cm,clip]{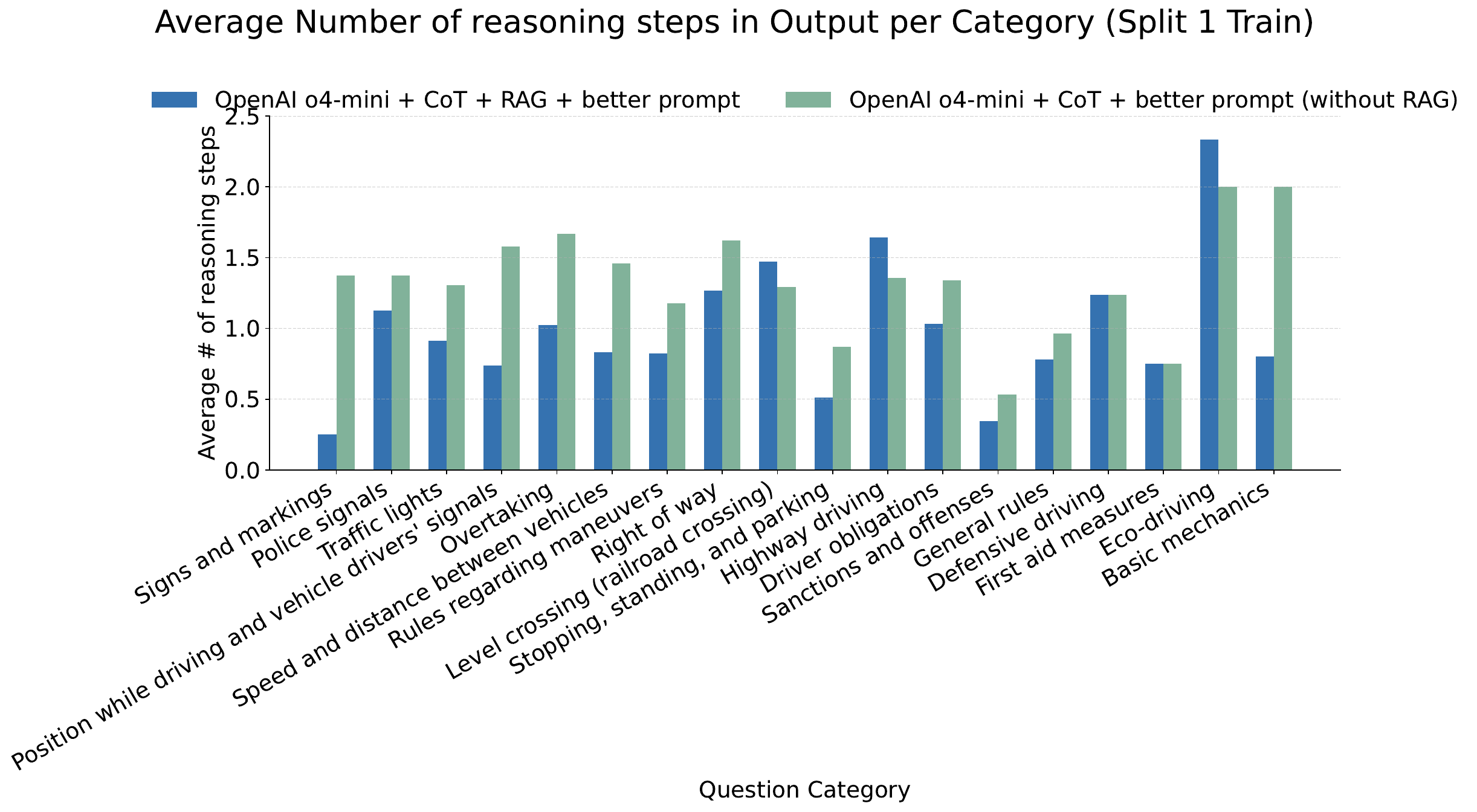}
\caption{Split 1 train.}
\label{fig:qa_reasoning_avg_occurrences_per_category_1_train}
\end{subfigure}
\hfill
\begin{subfigure}{\columnwidth}
\centering
\includegraphics[width=0.9\textwidth,trim=0 1.1cm 0 2cm,clip]{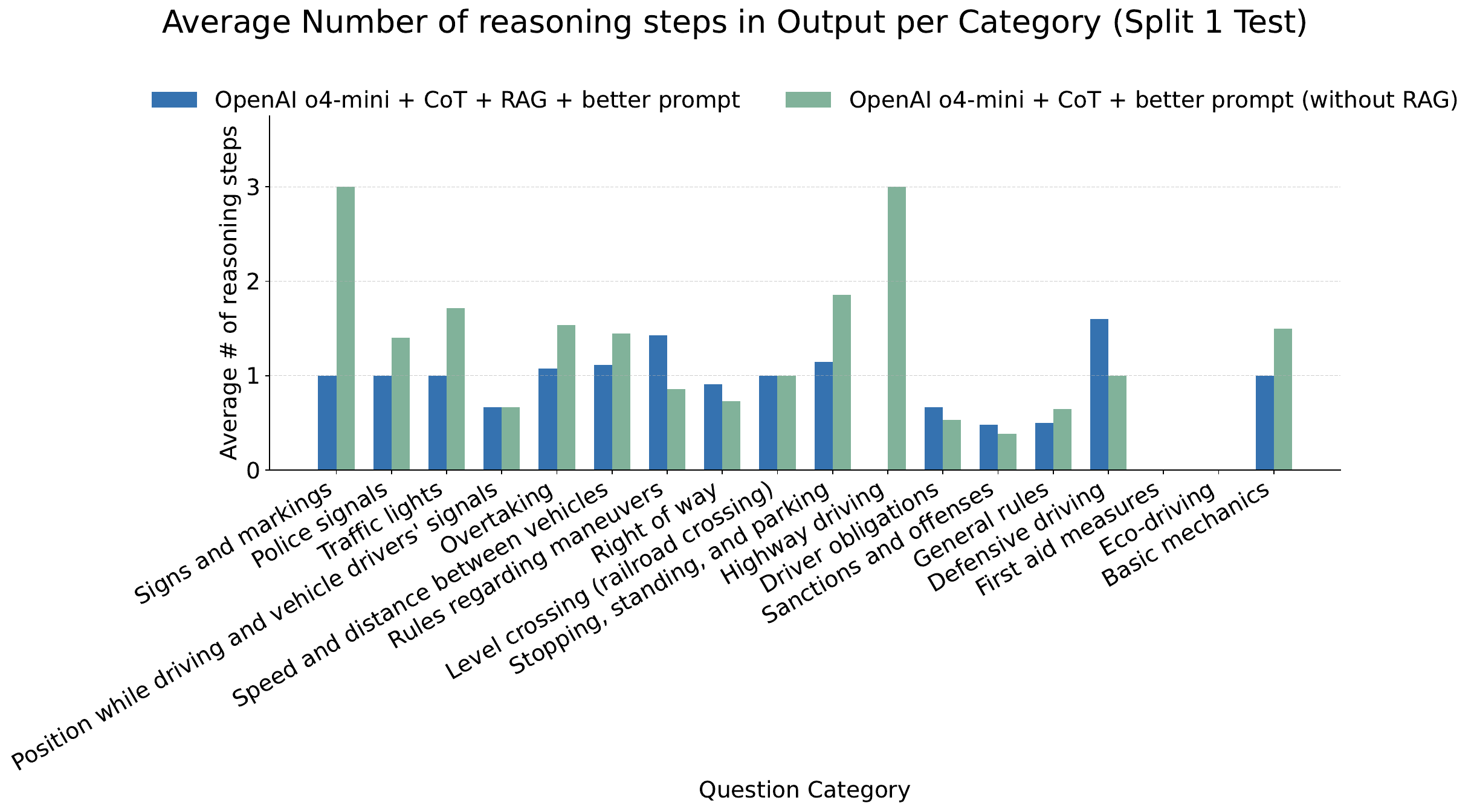}
\caption{Split 1 test.}
\label{fig:qa_reasoning_avg_occurrences_per_category_1_test}
\end{subfigure}
\hfill
\begin{subfigure}{\columnwidth}
\centering
\includegraphics[width=0.9\textwidth,trim=0 1.1cm 0 2cm,clip]{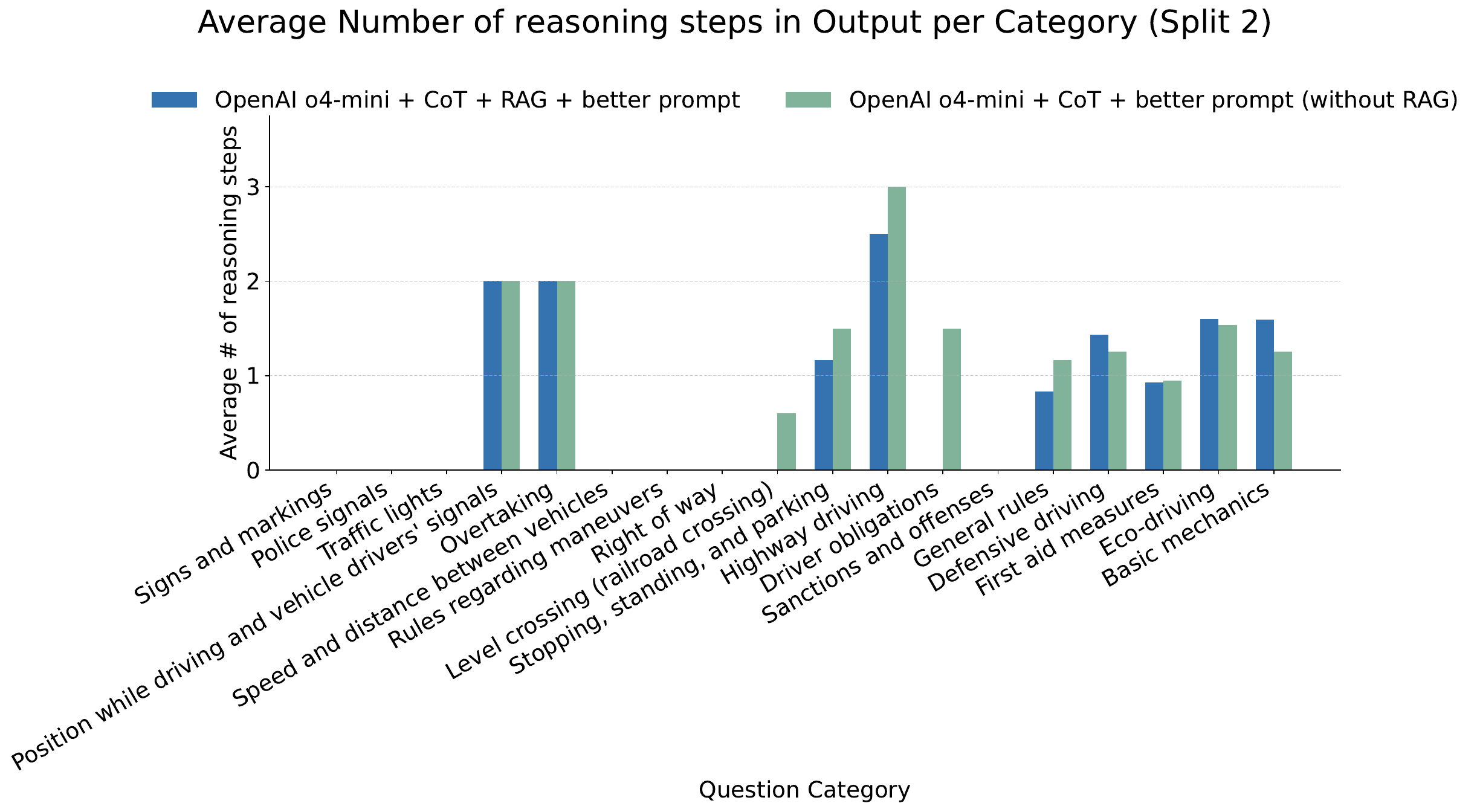}
\caption{Split 2.}
\label{fig:qa_reasoning_avg_occurrences_per_category_2}
\end{subfigure}
\caption{Average number of reasoning steps in output per question category on the QA task.}
\label{fig:qa_reasoning_avg_occurrences_per_category}
\end{figure*}

If we look at the number of reasoning steps, in Figure \ref{fig:qa_reasoning_avg_occurrences_per_category}, when we do not include RAG, the model needs more steps to arrive at a final answer. This seems like normal behavior because in the RAG case, it gets missing information rather than reasoning about it. It needs most of the reasoning steps in eco-driving or first-aid, where legal documents do not really help.

\clearpage

\subsection{Visual Information Retrieval}

\begin{figure*}[htbp]
\begin{subfigure}{\columnwidth}
\centering
\includegraphics[width=0.7\textwidth,trim=0 1.3cm 0 1.31cm,clip]{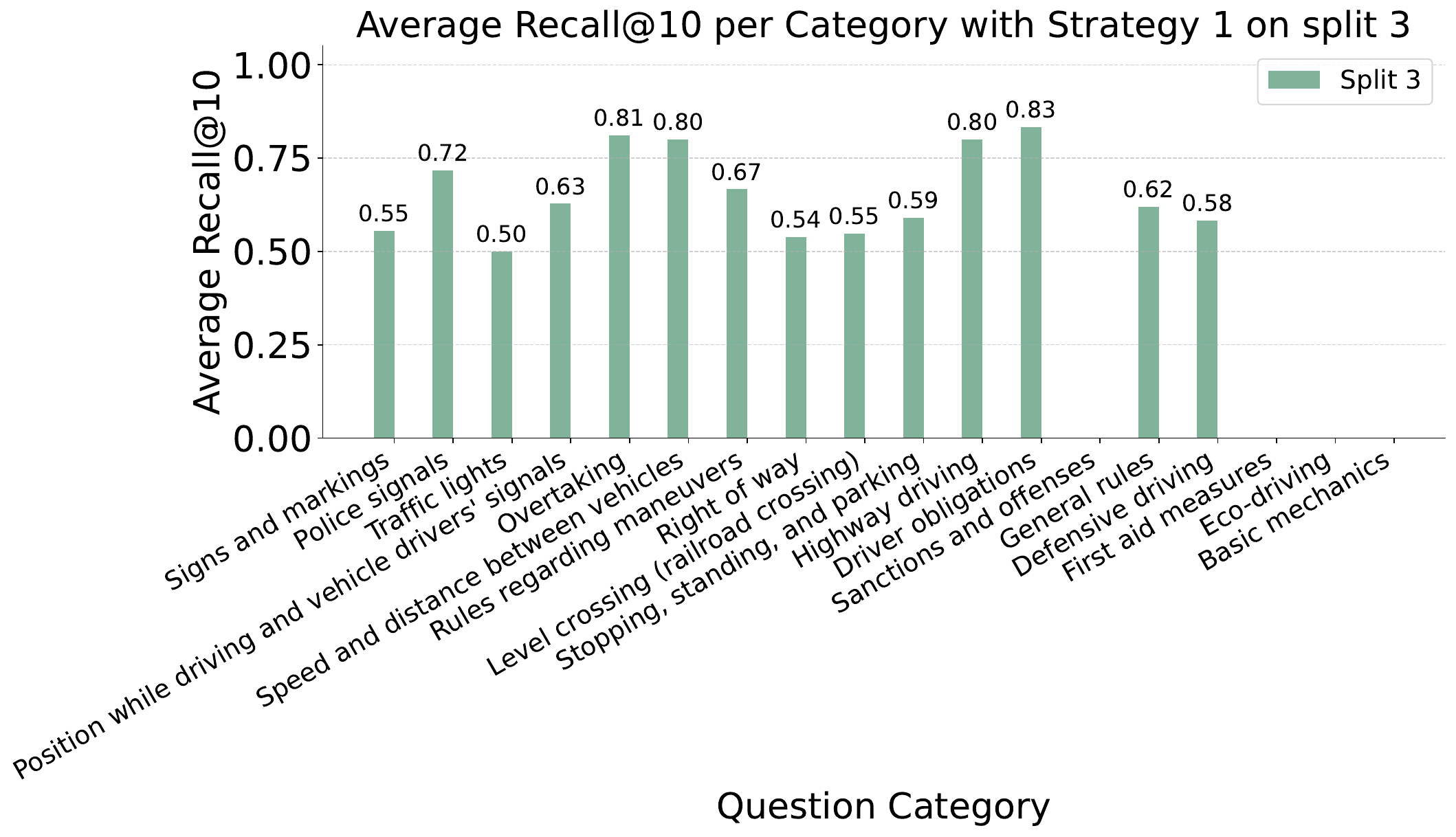}
\caption{Strategy (1).}
\label{fig:ir_recall_per_category_train_test_st7}
\end{subfigure}
\hfill
\begin{subfigure}{\columnwidth}
\centering
\includegraphics[width=0.7\textwidth,trim=0 1.3cm 0 1.31cm,clip]{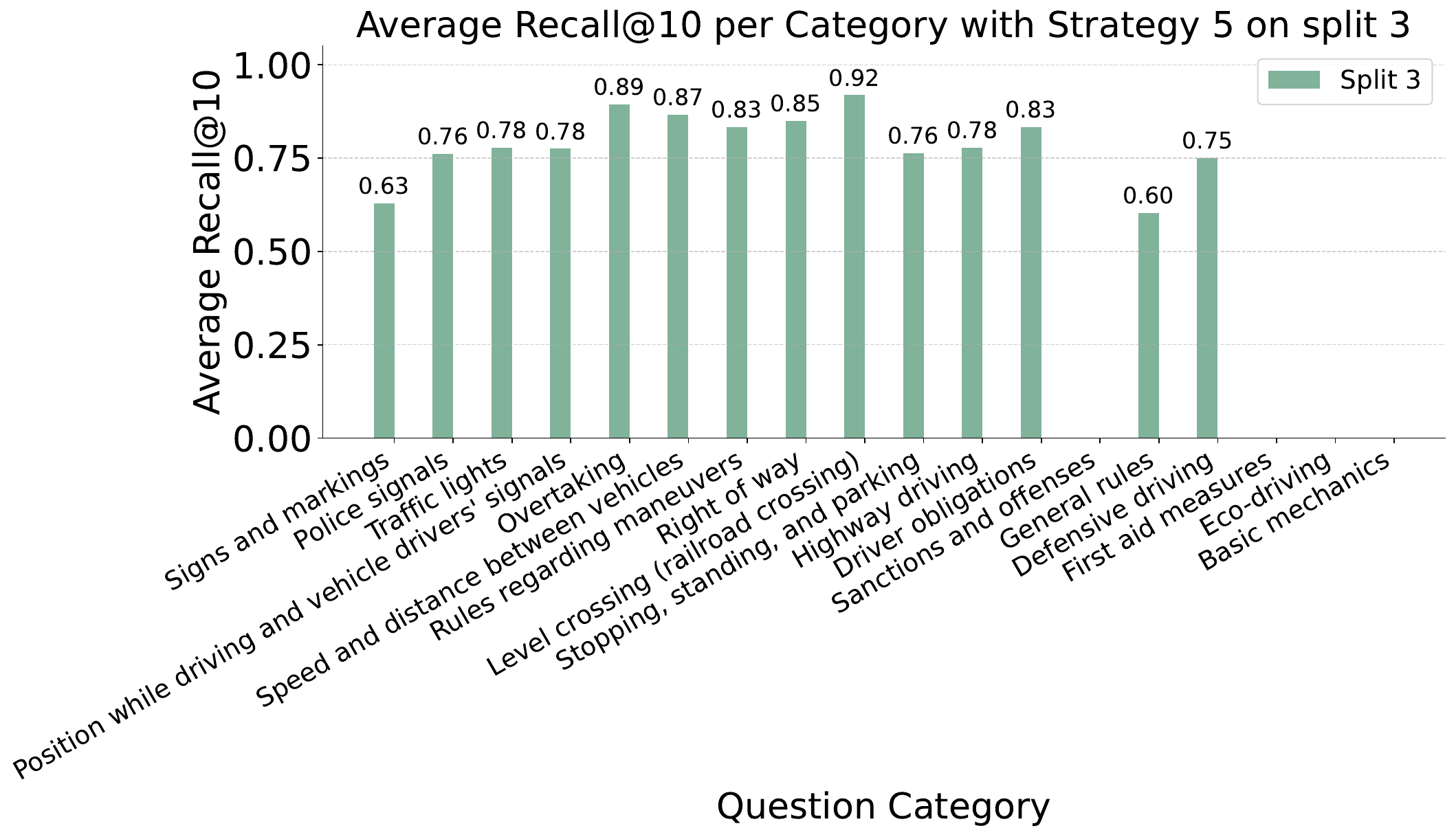}
\caption{Strategy (5).}
\label{fig:ir_recall_per_category_train_test_st9}
\end{subfigure}
\caption{Average Recall@10 per category for VIR task with strategy (1) on split 3.}
\label{fig:ir_recall_per_category_train_test}
\end{figure*}

\begin{figure*}[!hb]
\centering
\includegraphics[width=\textwidth,trim=0 1.3cm 0 1.8cm,clip]{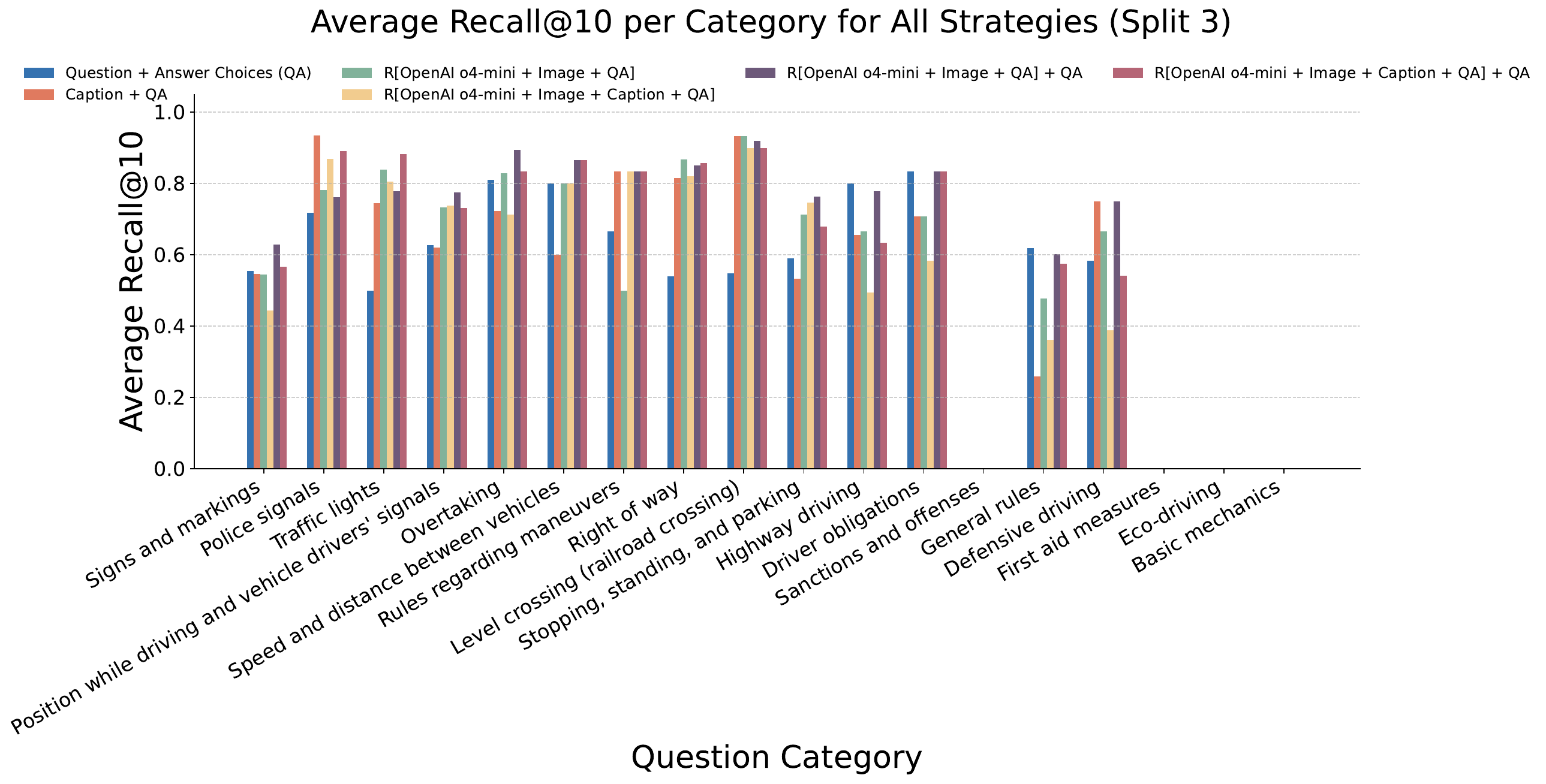}
\caption{Average Recall@10 per category for VIR task with all strategies on split 3.}
\label{fig:vir_recall_per_category_all_strategies_split_3}
\end{figure*}

Per category analysis of the retrieved laws, on split 3 using strategies (1) and (5), in Figure \ref{fig:ir_recall_per_category_train_test}, it performs worse on \textit{signs and markings}, \textit{general rules}, and \textit{defense driving}. The first one is more traffic-sign-intensive, suggesting that the model focuses more on the signs than on the laws related to the question. Figure \ref{fig:vir_recall_per_category_all_strategies_split_3} shows the average tendency on all strategies, with a similar tendency.

\clearpage
\subsection{Visual Question Answering}

\begin{figure*}[htbp]
\begin{subfigure}{\columnwidth}
\centering
\includegraphics[width=\textwidth,trim=0 1.3cm 0 1.8cm,clip]{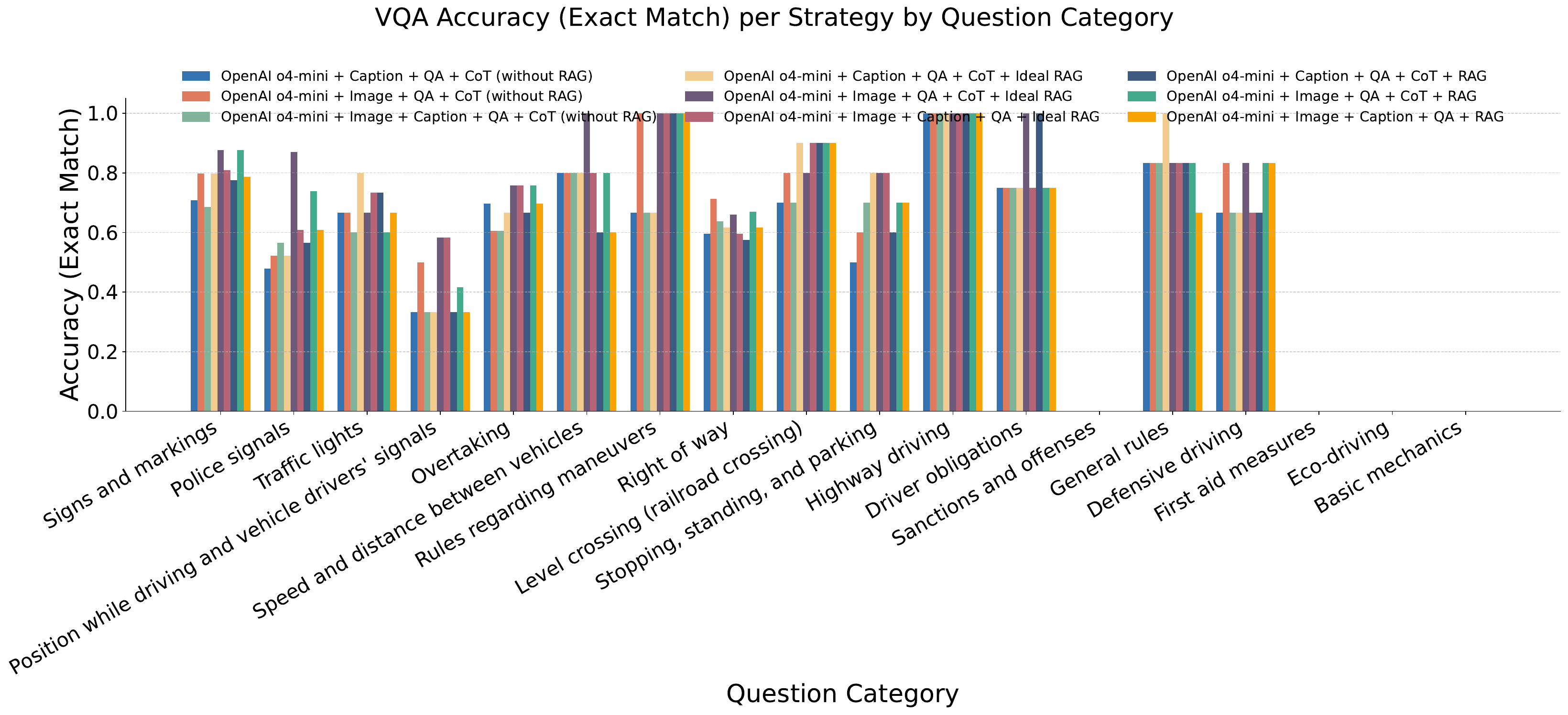}
\caption{Split 3.}
\label{fig:vqa_accuracy_exact_match_per_category_strategies_sp3}
\end{subfigure}
\hfill
\begin{subfigure}{\columnwidth}
\centering
\includegraphics[width=\textwidth,trim=0 1.3cm 0 1.8cm,clip]{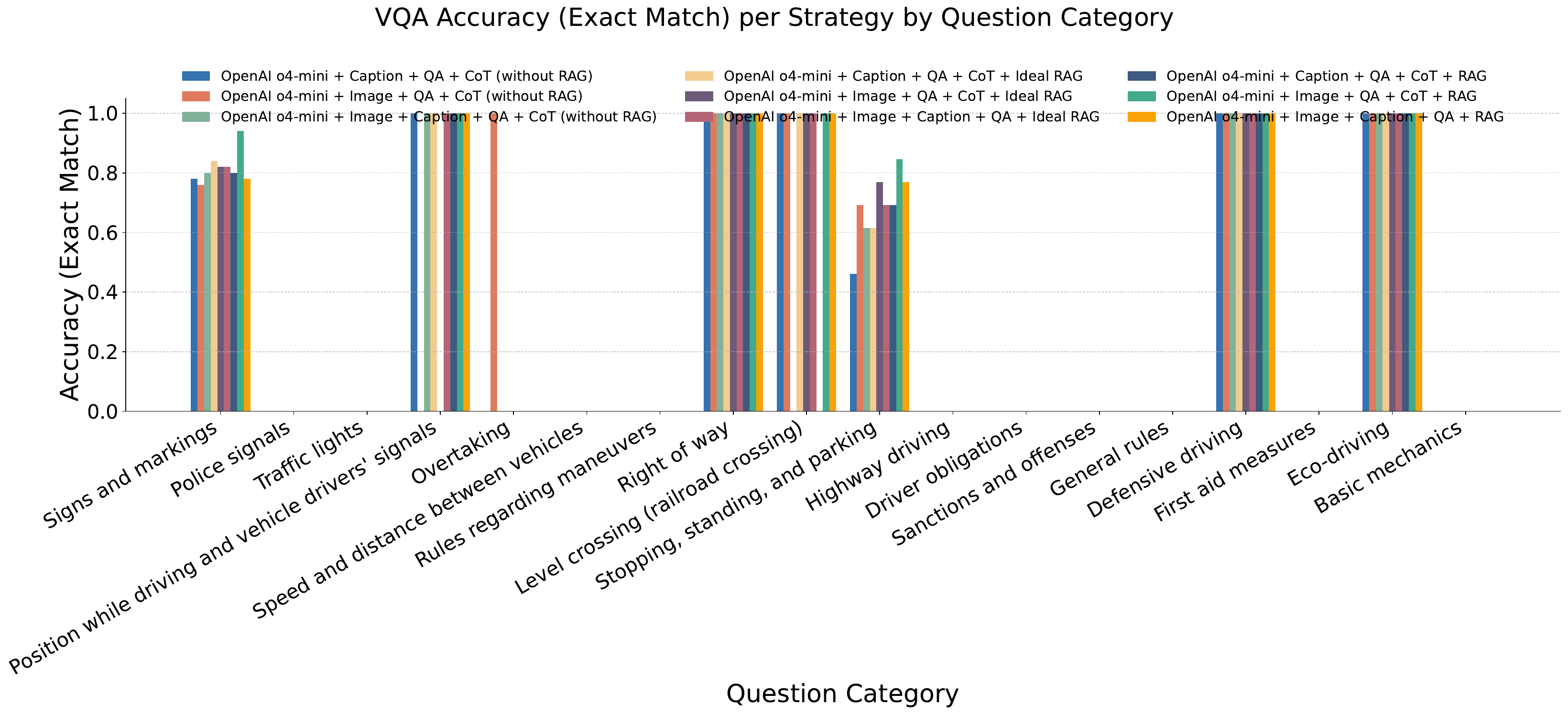}
\caption{Split 4.}
\label{fig:vqa_accuracy_exact_match_per_category_strategies_sp4}
\end{subfigure}
\caption{Exact Match score for VQA per strategy and question category.}
\label{fig:vqa_accuracy_exact_match_per_category}
\end{figure*}

\begin{figure*}[htbp]
\begin{subfigure}{\columnwidth}
\centering
\includegraphics[width=\textwidth,trim=0 0.8cm 0 1.8cm,clip]{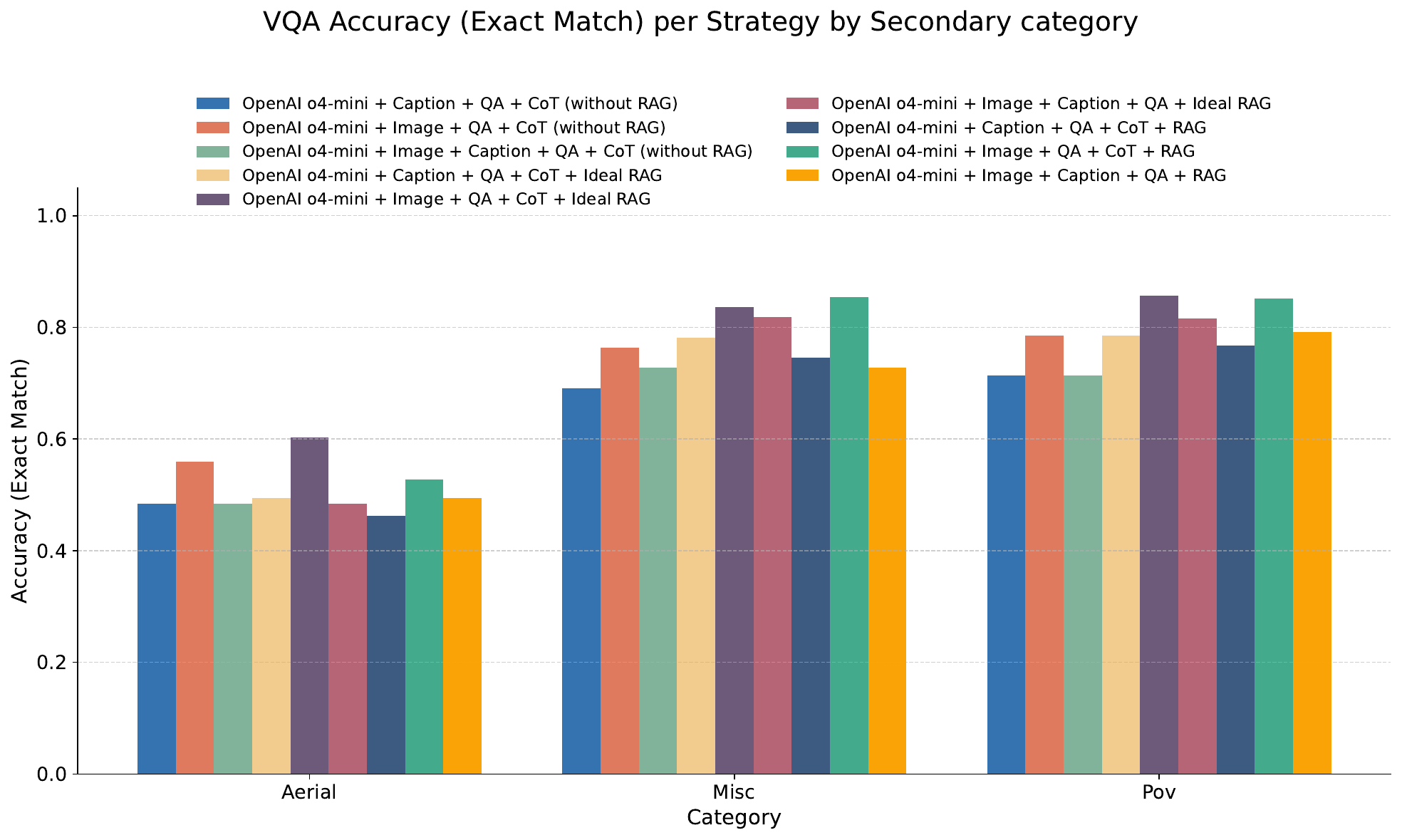}
\caption{Split 3.}
\label{fig:vqa_accuracy_exact_match_per_category_strategies_sp3_secondary}
\end{subfigure}
\hfill
\begin{subfigure}{\columnwidth}
\centering
\includegraphics[width=\textwidth,trim=0 0.8cm 0 1.8cm,clip]{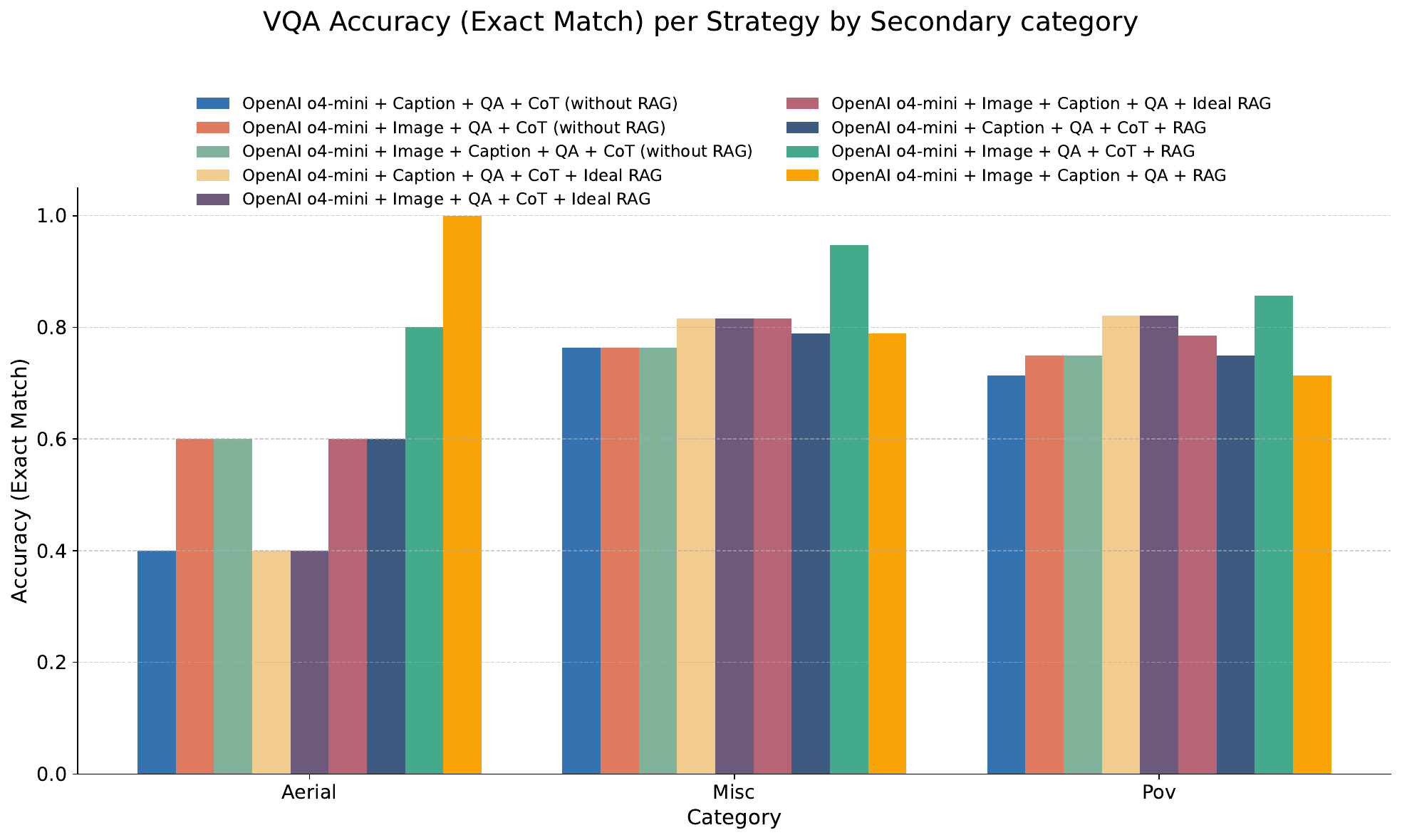}
\caption{Split 4.}
\label{fig:vqa_accuracy_exact_match_per_category_strategies_sp4_secondary}
\end{subfigure}
\caption{Exact Match score for VQA per strategy and secondary category.}
\label{fig:vqa_accuracy_exact_match_per_category_secondary}
\end{figure*}

\begin{figure*}[htbp]
\begin{subfigure}{\columnwidth}
\centering
\includegraphics[width=\textwidth,trim=0 1.2cm 0 1.8cm,clip]{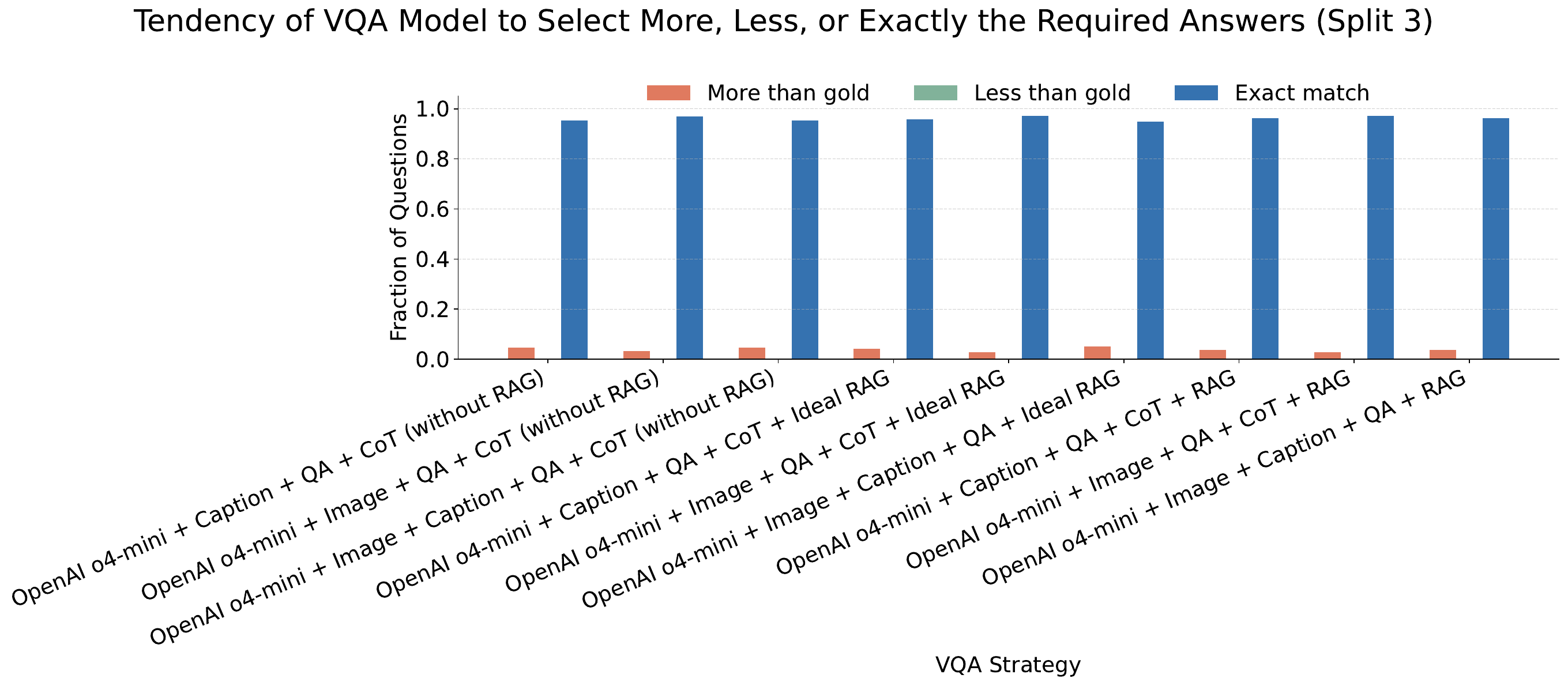}
\caption{Split 3.}
\label{fig:vqa_tendency_over_under_exact_3}
\end{subfigure}
\hfill
\begin{subfigure}{\columnwidth}
\centering
\includegraphics[width=\textwidth,trim=0 1.2cm 0 1.8cm,clip]{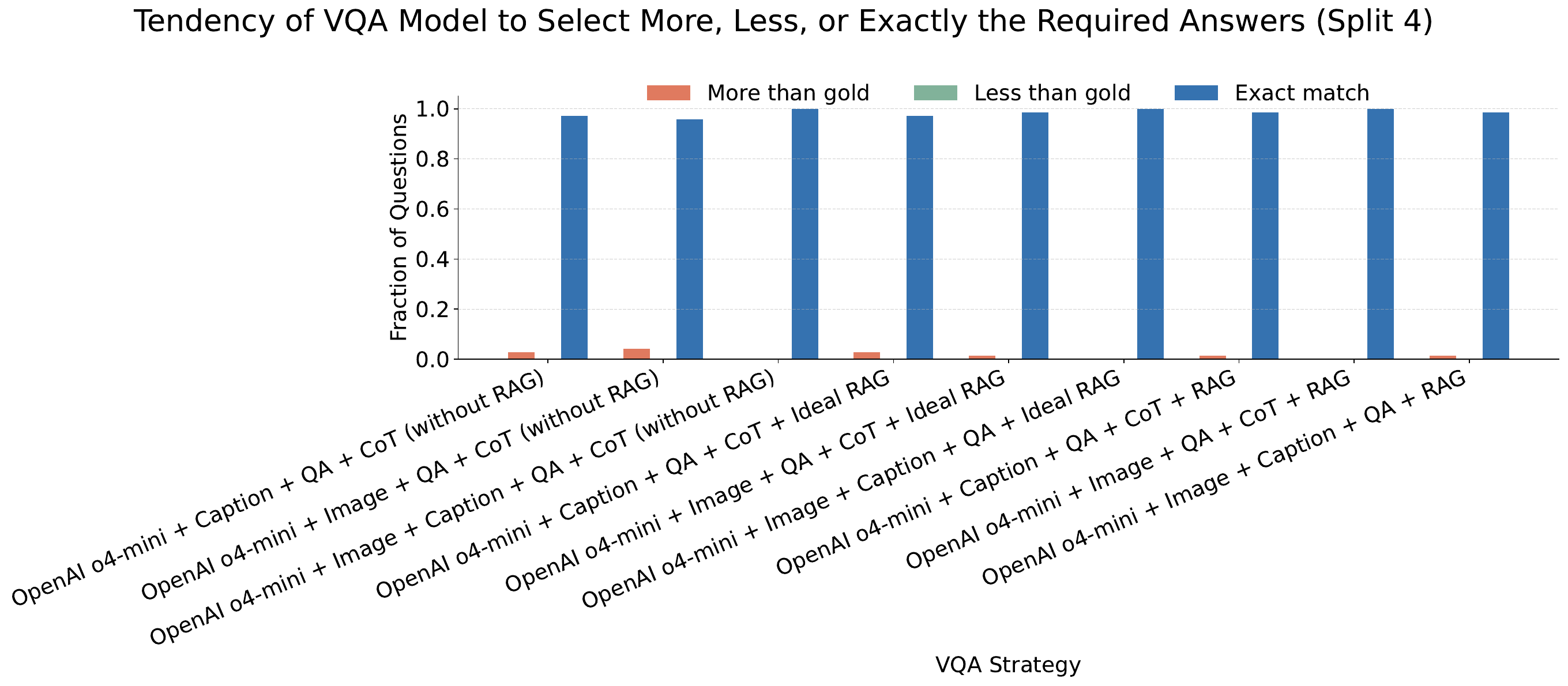}
\caption{Split 4.}
\label{fig:vqa_tendency_over_under_exact_4}
\end{subfigure}
\caption{Tendency of o4-mini to select more, less, or exactly the required answers on the VQA task.}
\label{fig:vqa_tendency_over_under_exact}
\end{figure*}

\begin{figure*}[htbp]
\begin{subfigure}{\columnwidth}
\centering
\includegraphics[width=\textwidth,trim=0 1.2cm 0 1.8cm,clip]{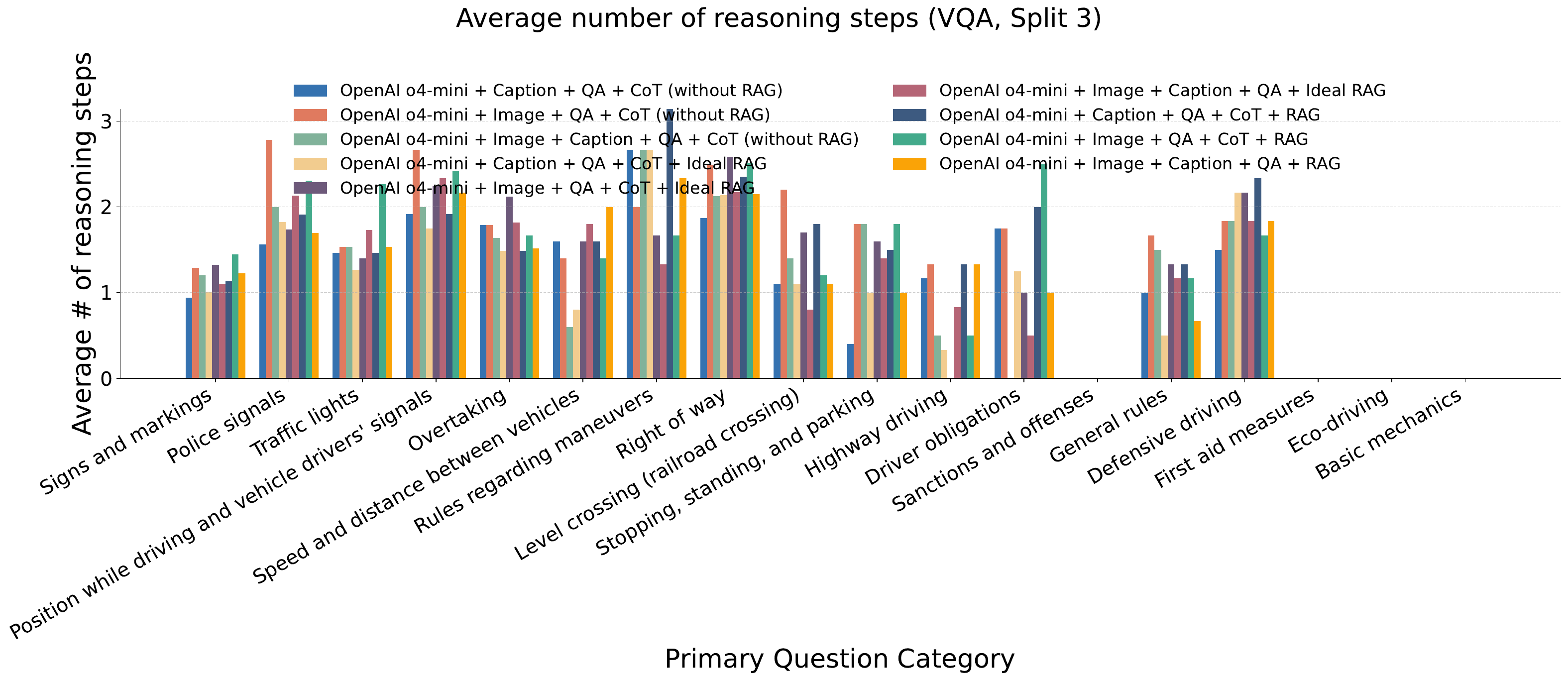}
\caption{Split 3.}
\label{fig:vqa_reasoning_avg_occurrences_per_question_category_strategies_sp3}
\end{subfigure}
\hfill
\begin{subfigure}{\columnwidth}
\centering
\includegraphics[width=\textwidth,trim=0 1.2cm 0 1.8cm,clip]{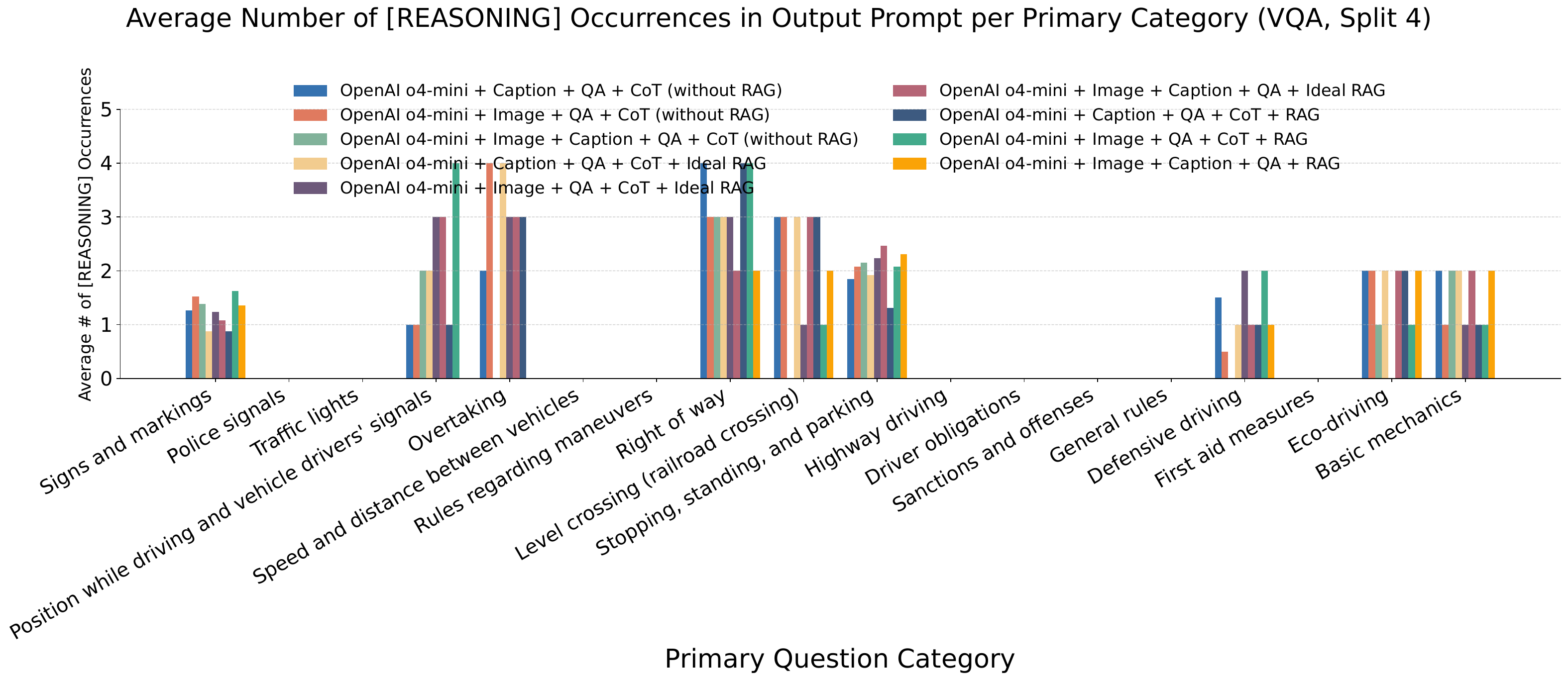}
\caption{Split 4.}
\label{fig:vqa_reasoning_avg_occurrences_per_question_category_strategies_sp4}
\end{subfigure}
\caption{Average number of reasoning steps in output per primary category on the VQA task.}
\label{fig:vqa_reasoning_avg_occurrences_per_question_category_strategie}
\end{figure*}

\begin{figure*}[htbp]
\begin{subfigure}{\columnwidth}
\centering
\includegraphics[width=\textwidth,trim=0 0.8cm 0 1.8cm,clip]{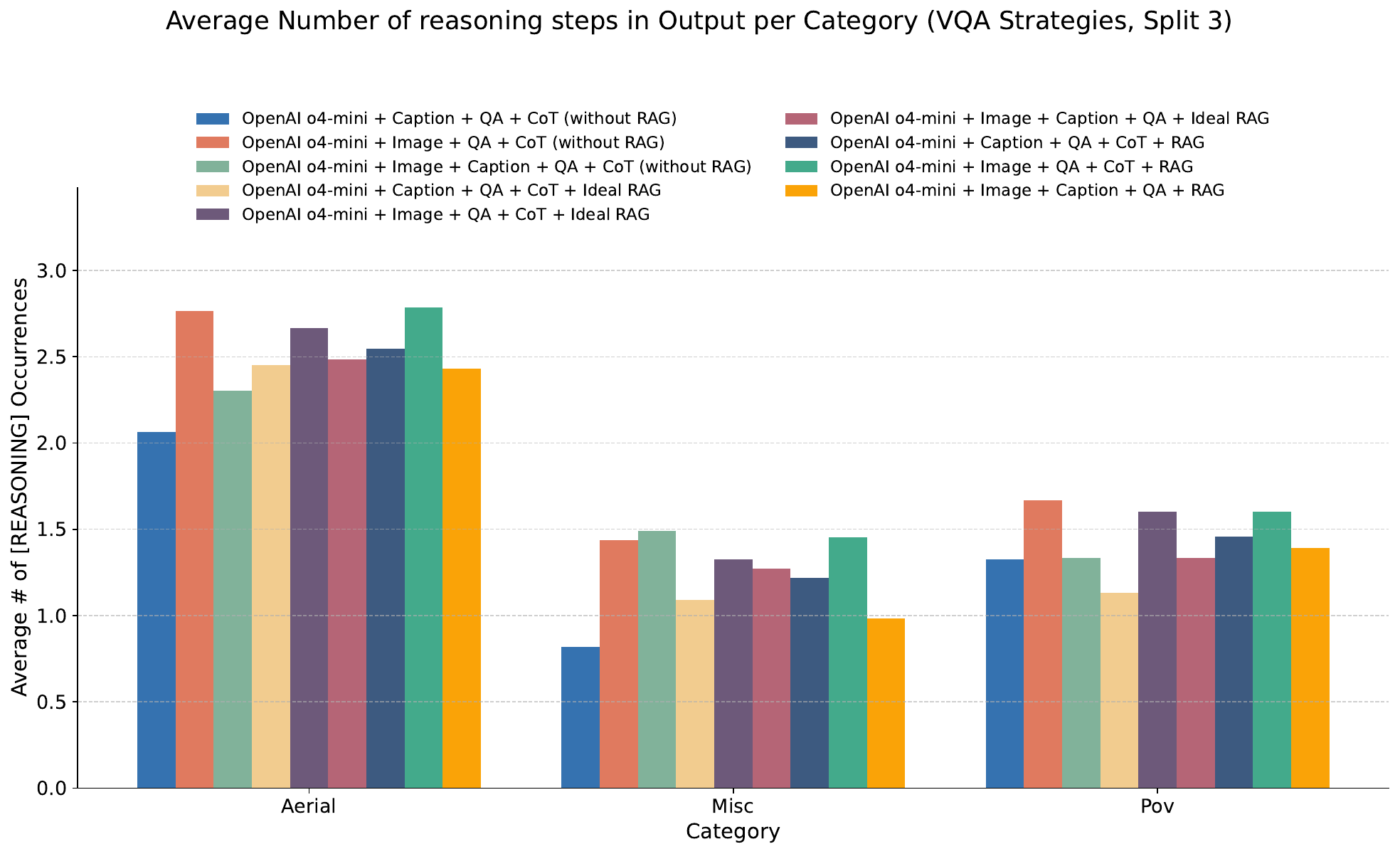}
\caption{Split 3.}
\label{fig:vqa_reasoning_avg_occurrences_per_category_strategies_secondary_sp3}
\end{subfigure}
\hfill
\begin{subfigure}{\columnwidth}
\centering
\includegraphics[width=\textwidth,trim=0 0.8cm 0 1.8cm,clip]{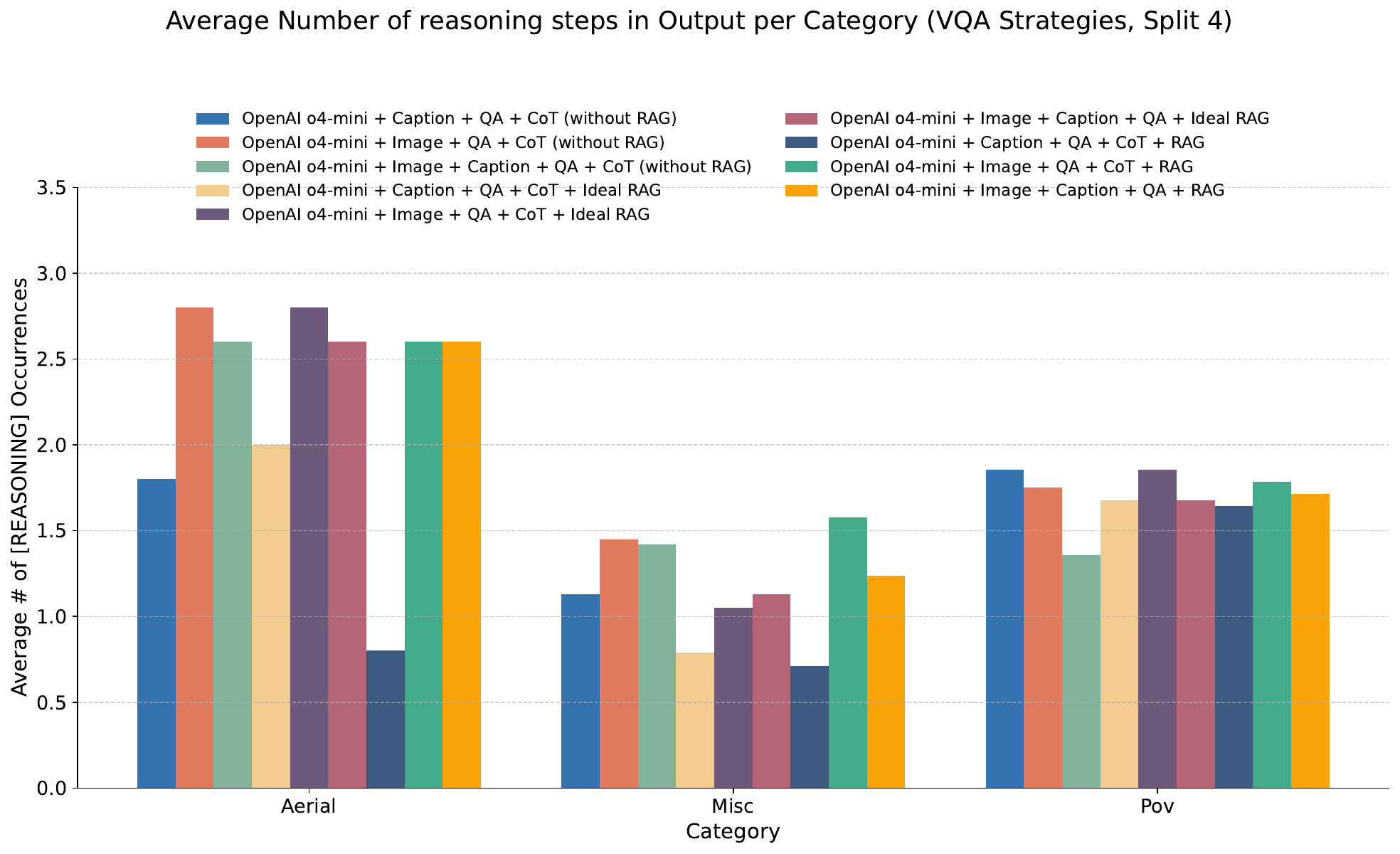}
\caption{Split 4.}
\label{fig:vqa_reasoning_avg_occurrences_per_category_strategies_secondary_sp4}
\end{subfigure}
\caption{Average number of reasoning steps in output per secondary category on the VQA task.}
\label{fig:vqa_reasoning_avg_occurrences_per_question_category}
\end{figure*}

Looking at Figure \ref{fig:vqa_accuracy_exact_match_per_category}, we see that the model has lower performance in \textit{position while driving as vehicle drivers' signals} and \textit{stopping, standing, and parking}. If we look at secondary categories, in Figure \ref{fig:vqa_accuracy_exact_match_per_category_secondary}, we notice a worse performance in the \textit{aerial} category and similar performance in the other two. Similarly, this is the category where, in Figure \ref{fig:vqa_reasoning_avg_occurrences_per_question_category_strategie}, the models tend to use more reasoning steps than the other.
If we look at the number of selected answers in Figure \ref{fig:vqa_tendency_over_under_exact}, the model tends not to select more than enough answers (even if incorrect).
We show the comparison of the reasoning steps in the primary category in Figure \ref{fig:vqa_reasoning_avg_occurrences_per_question_category}.

\end{document}